\documentclass[sn-mathphys-num,iicol]{sn-jnl}

\usepackage{graphicx}%
\usepackage{multirow}%
\usepackage{amsmath,amssymb,amsfonts}%
\usepackage{amsthm}%
\usepackage{mathrsfs}%
\usepackage[title]{appendix}%
\usepackage[table]{xcolor}%
\usepackage{textcomp}%
\usepackage{manyfoot}%
\usepackage{booktabs}%
\usepackage{algorithm}%
\usepackage{algorithmicx}%
\usepackage{algpseudocode}%
\usepackage{listings}%

\usepackage{tabularx}
\definecolor{TableGray}{gray}{0.9}
\usepackage{subcaption}
\usepackage{tikz}
\usepackage{pgfplots}
\pgfplotsset{compat=1.18}
\usetikzlibrary{mindmap, calc, patterns, shapes.misc, patterns.meta}
\usepackage{threeparttable}
\usepackage{xurl}

\begin{document}

\title[Article Title]{Co-designing Zoomorphic Robot Concepts for Animal Welfare Education}


\author*[1,2]{\fnm{Isobel} \sur{Voysey}}\email{i.a.voysey@sms.ed.ac.uk}

\author[2,1]{\fnm{Lynne} \sur{Baillie}}\email{l.baillie@hw.ac.uk}

\author[3]{\fnm{Joanne} \sur{Williams}}\email{jo.williams@ed.ac.uk}

\author[1,2]{\fnm{Michael} \sur{Herrmann}}\email{michael.herrmann@ed.ac.uk}

\affil*[1]{\orgdiv{School of Informatics}, \orgname{University of Edinburgh}, \orgaddress{\city{Edinburgh}, \country{UK}}}

\affil[2]{\orgdiv{School of Mathematics and Computer Science}, \orgname{Heriot-Watt University}, \orgaddress{\city{Edinburgh}, \country{UK}}}

\affil[3]{\orgdiv{School of Health in Social Science}, \orgname{University of Edinburgh}, \orgaddress{\city{Edinburgh}, \country{UK}}}


\abstract{Animal welfare education could greatly benefit from customized robots to help children learn about animals and their behavior, and thereby promote positive, safe child-animal interactions. To this end, we ran Participatory Design workshops with animal welfare educators and children to identify key requirements for zoomorphic robots from their perspectives. Our findings encompass a zoomorphic robot's appearance, behavior, and features, as well as concepts for a narrative surrounding the robot. Through comparing and contrasting the two groups, we find the importance of: negative reactions to undesirable behavior from children; using the facial features and tail to provide cues signaling an animal's internal state; and a natural, furry appearance and texture. We also contribute some novel activities for Participatory Design with children, including branching storyboards inspired by thematic apperception tests and interactive narratives, and reflect on some of the key design challenges of achieving consensus between the groups, despite much overlap in their design concepts.}


\maketitle

\section{Introduction}

Around 70\% of children in the UK grow up with pets as part of their families~\cite{marsa2016sociodemographics}. Positive interactions with animals, especially pets, can have benefits for children’s mental health and their cognitive, social, and emotional development~\cite{marsa2016short, purewal2017companionanimals}. On the other hand, negative interactions, like witnessing or engaging in animal abuse, are associated with a range of psychological risk factors and maladaptive behaviors~\cite{ladny2020traumatizedwitnesses, ferreira2023understanding, wauthier2022understanding}. With this in mind, organizations like the Scottish Society for the Prevention of Cruelty to Animals (Scottish SPCA) run animal welfare education (AWE) programs to educate and inspire children to treat animals with kindness and respect~\cite{sspca2019about}.

It is not possible to use live animals in these programs due to stress to the animal, hygiene, bite risks, etc.~\cite{glenk2017therapydog,muldoon2021delphi1establishing}, so they have to simulate interactions with animals using videos, imaginative exercises, or a stand-in, like a stuffed toy~\cite{muldoon2021delphi2challenges}.
However, the Scottish SPCA are expanding their tools and embracing technology, namely zoomorphic robots.

Any tool for use in AWE needs to be informative and engaging; 85\% of animal welfare (AW) educators agreed running fun, interactive sessions that ensure active learner participation and engagement is one of the most critical components for intervention success~\cite{muldoon2021delphi1establishing}. A robot has many benefits as an educational tool compared to traditional pen-and-paper tools, stuffed toys, or web-based activities because it allows children to physically interact with it and can provide autonomous behaviors~\cite{belpaeme2018education}. The robot can be combined with an interactive narrative to form the educational intervention. Interactive narratives are a form of storytelling where the storyline is not fixed in advance and can change in response to actions and decisions by the user~\cite{riedl2013interactive}. The goal of such narratives is to give the user a sense of affectance---that their actions have an impact on the direction or outcome of the story---which fits well with a phrase used by an AW educator, that “actions have consequences”.

Design input from educators can help facilitate desired learning outcomes~\cite{muldoon2021delphi1establishing, steel2022teacherreadingdog, elloumi2022exploring}.
Furthermore, we recognize the importance of involving children in research that affects them~\cite{morgan2002focus, alves2021children}.
To facilitate input from both groups, we conducted a Participatory Design (PD) workshop with 11 AW educators and another with 24 8--11-year-old children.
We discussed the participants' perspectives on AWE, interactive narratives, and robotics before gathering feedback on two existing zoomorphic robots. We then guided them to design a pet robot for AWE and describe stories around it. We qualitatively analyzed the data, comparing and contrasting the two groups, to provide insight into the design of a zoomorphic robot and narrative for AWE.

This paper provides two key contributions. The first concerns design insights for the appearance, behavior, and features of zoomorphic robots and the narratives that could surround them from the perspective of children and AW experts. This work is the first to seek the perspectives of either AW educators or children in the design of zoomorphic robots.
Furthermore, few works have focused on stakeholders' design requirements for zoomorphic robots: \citet{collins2023skinzoomorphic} studied ten participants' preferences for a zoomorphic robot for adults with depression. This is despite some findings suggesting a preference for zoomorphic robots over humanoid robots in certain domains~\cite{bradwell2019companion, zhang2022paincodesign, lohse2007appearance, marchetti2022petfloorwashing}. 
The second contribution concerns the PD methodology and tools for doing this with primary-school-aged children and educators. 
This work continues in the vein of \citet{newbutt2022codesigning} and \citet{elloumi2022exploring} by involving both children and educators to develop robots for education. It expands on these works by using two PD workshops with similar structure and activities to allow for direct comparison between the two groups. 

\section{Background}


\subsection{Animal Welfare Education}
\label{sec:bkg_awe}

Any tool for AWE needs to be informative and engaging; 85\% of AW educators agreed running fun, interactive sessions that ensure active learner participation and engagement is one of the most critical components for intervention success~\cite{muldoon2021delphi1establishing}.
Excluding animals, robots are one of the richest tools for AWE; they can facilitate displays of animal behavior and responsive, embodied interactions, highlighting to children how their actions affect the situation.
For example, the Scottish SPCA has used a robotic kit that small groups of 8--11-year-old children build and code to prompt them to think about similarities between robots' and animals' senses and decision-making processes~\cite{sspca2018robowunderkind}.
However, despite high engagement from the children, educators have expressed some challenges with keeping the groups focused on animal minds because the robots do not realistically simulate animals, in form or behavior.

Previous work has involved close collaboration between researchers and AW charities to reach consensus on target areas~\cite{muldoon2021delphi1establishing}, produce practical guides for professionals~\cite{williams2023animal,muldoon2021awetoolkit}, and to design and evaluate AWE interventions~\cite{williams2022rabbit, hawkins2020seriousgame,wauthier2023preliminary}.
Through these collaborations, researchers can be confident the interventions designed meet the needs of AW educators and will successfully effect change in knowledge and attitudes~\cite{hawkins2020seriousgame, williams2022rabbit}. Researchers have also involved schoolteachers in the design process to ensure interventions are acceptable to the education community~\cite{steel2022teacherreadingdog}, but, so far, children have not been directly involved in the design process.

\subsection{Robotics in Education}

Robots can fill a variety of roles in educational interventions~\cite{miller2016education}, but here we focus on robots that deliver the learning experience through social interaction~\cite{belpaeme2018education}, as opposed to construction kits~\cite{damico2020educational}.
Social robots have been shown to have a positive impact on  motivation~\cite{saerbeck2010expressivetutor,kose2015signlanguagetutor} and engagement~\cite{ahmad2019robotvocabulary, kose2015signlanguagetutor} in educational tasks, which can contribute to improved task performance~\cite{saerbeck2010expressivetutor, ahmad2019robotvocabulary, kose2015signlanguagetutor, belpaeme2018education}.
Physically embodied robots also improve task performance~\cite{kose2015signlanguagetutor} and enjoyment~\cite{fasola2013exercisecoachembodiment} over virtually embodied robots.

Common interaction paradigms used in education position the robot as a teacher, tutor, peer, or novice~\cite{belpaeme2018education}. Although these are appropriate for humanoid robots, they do not easily fit a zoomorphic robot.
Instead, a more appropriate approach may be constructionist learning, where children can explore, make hypotheses, and validate their beliefs via experimentation~\cite{weinberg2003roboticsineducation}, or experiential learning, where the robot is a safe substitute with which to practice skills~\cite{roberts2004infantsim}.

\subsection{Interactive Narratives}

Interactive narratives aim to immerse the user in the story so that they believe they are a key part of the story and their actions can have a significant impact on its direction~\cite{riedl2013interactive}. An early example of interactive narratives is the choose-your-own-adventure novel where the reader's choices made in response to events in the story could lead to different endings. Similar ideas have been used in role-playing and video games where the player's choice on, e.g., which character to befriend will lead to different story beats, customizing the experience to each individual. Interactive narratives have been used to encourage users to empathize with characters~\cite{hand2009interactive} and consider appropriate courses of action, such as in anti-bullying education~\cite{aylett2007fearnot}. 
Therefore, interactive narratives are promising for AWE, depending on appropriate narrative framing and story beats, so that children can connect their actions with consequences.

\subsection{Participatory Design}

PD aims to engage users in the design of systems, give them agency in the design process, and encourage active contribution to the design~\cite{lee2017steps}. \citet{druin2002role} outlines four different roles children can play in the design of new technology: user, tester, informant, and design partner. Children likely take a variety of roles during a PD workshop, but mainly act as informants and design partners. PD has been used successfully with a variety of user groups, including children~\cite{alves2021children,derboven2015multimodal}. With children, PD techniques have been used to develop robots for creativity~\cite{alves2021children}, math tutoring~\cite{elloumi2022exploring}, support in a special educational needs school~\cite{newbutt2022codesigning}, anti-bullying interventions~\cite{sanoubari2021remotecodesign}, pain management~\cite{foster2023paincodesign,zhang2022paincodesign}, and ethical reflection~\cite{mott2022codesign}.
These works have mostly been with children aged between six and fourteen, which fits our target age group.

Some works involve children as one of multiple user groups in the design process. For example, \citet{elloumi2022exploring} conducted focus groups with teachers and children to learn requirements for a social robot in mathematics education, finding some overlap between the groups. \citet{newbutt2022codesigning} sought input from autistic students and teachers to develop a robot for the school, with teachers suggesting changes to an idea developed following focus groups with students.

\section{Methods}


\subsection{Designing the Workshops and Activities}

The workshops were designed to mirror each other in content and structure, but activities were modified to be suitable for each group and probe relevant experience. As suggested by \citet{arevalo2021reflecting}, workshops were conducted with the children and educators separately to enable both groups to express their perspective without influence from the other, as well as to accommodate the different activities.
As part of these workshops we trialed some new activities for PD with children, which we motivate and describe here. These activities build on techniques used in HRI and in psychology to investigate child-animal interactions.
All materials were anonymous in an attempt to lessen participants' perceived pressure to respond in a `favorable' way.
The materials are available online\footnote{\label{footnote:osf} \url{https://osf.io/zd89e/?view_only=fd944f3a0f3a4da08ee31e8ae6749f66}}.

\subsubsection{Personas}
\label{sec:personas}

Personas have been used in PD for HRI with user groups like older adults (e.g., \cite{ostrowski2021codesign, uzor2012senior}) because they create engagement with discussions and allow participants to share personal experience in a removed way.
There is limited work using personas to engage children during PD, but we suggest that children in our age range (8--11 years old) are likely to engage well with them due to a familiarity with perspective-taking during imaginative play.

The personas created for this workshop captured different relationships and experiences with common pets, from positive to negative to limited/neutral. Personas' ages matched the age of the workshop participants and the planned intervention target group (8--12-years-old), and we created one male, one female, and one gender-neutral persona. We made the caring, affectionate cat-owner male so as not to reinforce stereotypical gender roles around caregiving, such as have been seen in Lego\textsuperscript\textregistered sets \cite{reich2018lego}, and as a gentle challenge to some boys' views that affectionate relationships with pets are `feminine' or `childish'~\cite{tipper2011relationships}.
The personas were as follows:

\begin{itemize}
    \item \textbf{Alex} is 10 years old. He loves animals. He has a cat at home that he likes to play with and cuddle.
    \item \textbf{Bonnie} is 11 years old. She has a pet rabbit. She is scared of dogs after being bitten by one a few years ago.
    \item \textbf{Charlie} is 9 years old. They don’t have any pets at home.
\end{itemize}

\subsubsection{Demonstrations}

Demonstrations are often used in PD for HRI (e.g., \cite{newbutt2022codesigning, zhang2022paincodesign}) to help participants to see what technology currently exists as a springboard for their later designs, since robotics is often a novel technology to many participants~\cite{arevalo2021reflecting}. Demonstrations can also help participants to provide more concrete feedback on the use of robots for the proposed application~\cite{winkle2018therapists}.
Children can often come up with fantastical ideas, which is part of the benefit of involving them in PD~\cite{guha2013reflections}, but demonstrations where children can interact with the robot and ask questions can help them to learn about the functionality of robots, give them a better sense of the design space, and ground their designs~\cite{voysey2023introducing, zhang2022paincodesign, barendregt2020demystifying}.

\subsubsection{Design Prompts}
\label{sec:designprompts}

The educators were simply prompted to sketch and annotate a robot for AWE, but for the children we introduced pre-design brainstorming prompts. Breaking the design task down into smaller chunks can help children with idea generation~\cite{guha2004mixing,sanoubari2021remotecodesign}, but it also served to refocus children's minds onto pet animals. This, in combination with the demonstrations, was designed to ground the children's designs in terms of pets and robots, in an attempt to avoid totally fantastical creations or ones that had no elements of pets---we expected some unfeasible designs, but wanted to avoid unicorns and dragons!

\paragraph{Brainstorming}

Brainstorming is a commonly used tool in PD to facilitate idea generation~\cite{elloumi2022exploring,sanoubari2021remotecodesign}.
\citet{sanoubari2021remotecodesign} used three brainstorming prompts on appearance, behavior, and name for a `student robot' prior to children starting the sketching portion of the design task to break the task down into smaller, more manageable chunks.
For this workshop children were given two sets of prompts. The first set of prompts drew on children's senses, prompting them to think about how a pet looks, feels, and sounds. The second set of prompts were about how pets behave. No reference was made to specific pet species (e.g., cats or dogs) nor was reference made to children's own pets so that children were free to use real and imagined experiences with a variety of pets.

\paragraph{Draw-and-write}

The main design activity for both groups used the draw-and-write technique. The draw-and-write method is generally viewed as an enjoyable, non-threatening way for children to communicate information without the need from them to immediately verbalize their thoughts~\cite{angell2015draw, gauntlett2006creative} and has been used to elicit ideas and input from children on a range of topics. Since the balance of drawing and writing can be adapted by each participant, it reduces the demand on participants' language skills and is inclusive of those with varying verbal comprehension and communication skills~\cite{georgiou2023refugees}. Therefore, draw-and-write is particularly appropriate for a workshop with a range of ages.
Sketching and annotations are also commonly used in PD workshops with adults~\cite{uzor2012senior}.

\subsubsection{Storyboards}

Storyboards are a classic tool used in PD, especially with children and adolescents~\cite{beelen2022designing,bjorling2019participatory, sanoubari2021remotecodesign}. They both elicit and document ideas, and in HRI can be used to understand how participants envision interactions with robots. The aim of the storyboard activity in this workshop was to investigate children's awareness and understanding of different scenarios surrounding pets and the different ways they felt people could respond to the robot pet.

A similar tool that uses narratives is the thematic apperception test (TAT). These prompt the participant to create a narrative around ambiguous images, which can reveal underlying motives, concerns, and ways they look at the world~\cite{murray1943thematic}. TATs have been used in many areas of psychological research and therapeutic practice, including interventions in cases of confirmed or suspected animal abuse \cite{shapiro2013anicare}. The `Animals-At-Risk' TAT uses images ``surrounding common situations that might create tension in the human–animal relationship''~\cite[p.~92]{shapiro2013anicare} and is used to understand children's experiences around pets within the family setting.

In this new tool, we combined storyboards with the ambiguous, open prompt aspect of TATs, focused on animal emotions.
This provided a scaffold for the children to develop their storyboards around---prior research has found that an intermediate degree of scaffolding facilitates more idea elicitation than completely unstructured storyboards~\cite{moraveji2007comicboarding}. However, we did not include any images of scenarios so that children could include the robot they had designed in the previous exercise.
To start, children were prompted to choose one of four emotions their robot pet could be feeling: happy, sad, angry, or scared. The panels of the storyboard were labeled with questions that prompted the child to fill in the events surrounding the emotion. Questions have been used to provide structure and prompts for participants~ \cite{beelen2022designing,disalvo2008neighborhood,mott2022codesign,sanoubari2021remotecodesign}, but have previously been kept separate from the storyboard template.

\begin{figure*}
    \centering
    \includegraphics[width=\textwidth]{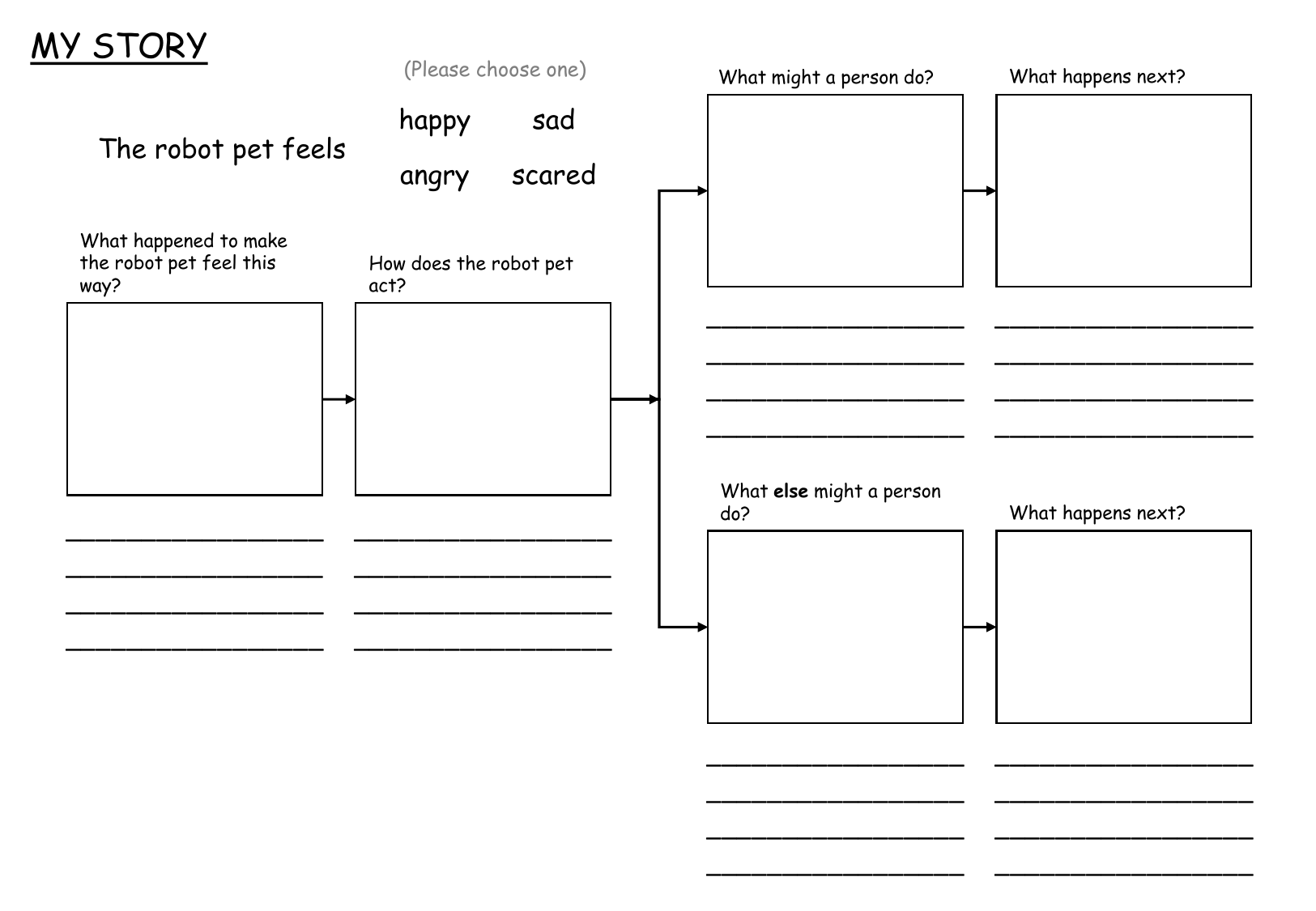}
    \caption{Storyboard template given to children}
    \label{fig:storyboard_template}
\end{figure*}

Another new technique used in this storyboard is branching. This enabled us to probe different potential scenarios surrounding the robot emotion.
The branching occurred at the point of human action, which also reflects the use of the `Animals-At-Risk' TAT; therapists can help teach self-management by prompting the child to think about how their and others' reactions to the situation could be different and what the potential outcome would then be~\cite[p.~67]{shapiro2013anicare}.
More branches could be added to probe different events that could have precipitated the emotion, for example, but we decided that this could become too complicated and potentially confusing.
The branching storyboard template given to children is shown in Fig.~\ref{fig:storyboard_template}.

\subsection{Conducting the Workshops}

\subsubsection{Participants}

Eleven educators took part in the first workshop (10 female, 1 male). Participants were recruited through partnership with the Scottish SPCA. All participants were part of the charity's Education Team, which is responsible for developing and delivering AW programs to the public. 
Two participants had 1 year of experience or less, seven had 3--5 years, and two had 10+ years of experience in AWE.
On average, they had 4.9 years of experience in AWE ($SD{=}\text{3.4 years}$).

Twenty-four children from a school on the outskirts of Edinburgh took part in the second workshop (12 female, 12 male).
The group was formed of three consecutive year groups from the school, so the children ranged in age from 8 years 9 months to 11 years 5 months ($M{=}\text{10 years 4 months}$, $SD{=}\text{10 months}$).
Of the 24 children, 17 owned pets. Dogs were the most common pet (13 owners), followed by cats and hamsters (6 owners each) and fish (4 owners). Ten of the 17 pet owners had more than one type of pet.

\subsubsection{Setup}

The research protocol received ethical approval from the University of Edinburgh School of Informatics. Educators, children, and guardians received information sheets before the study and signed physical consent forms.

The educators' workshop lasted 2.5 hrs and took place at the charity's headquarters. The discussions were held in two small groups, each of which was run by a researcher. The third researcher facilitated the running of the workshop and took additional notes.

The children's workshop lasted 2.5 hrs (1 hr 50 mins of activities and two breaks) and took place at the university. The discussions happened in five small groups of four or five children (suggested to be the optimal size for focus groups with children~\cite{heary2002focus}), each led by a researcher. Two other researchers facilitated the workshop, and the children were accompanied by three school staff who assisted when necessary, such as helping a younger child with writing.

\subsubsection{Procedure}

The two workshops followed the same structure: an introduction from the researchers; background discussion on AWE, robotics, and interactive narratives; demonstrations of two zoomorphic robots, MiRo and Qoobo, and participant feedback on the robots; a design session; and a plenary feedback session. The overall structure and breakdown of activities is shown in Table~\ref{tab:workshop_phases}.

\begin{table*}[]
    \centering
    \rowcolors{1}{white}{TableGray}
    \renewcommand{\tabularxcolumn}[1]{m{#1}}
    \caption{Workshop phases and activities}
    \begin{tabularx}{\linewidth}{>{\raggedright\arraybackslash}p{0.18\linewidth}XXl}
    \toprule
        \multirow{2}{*}{\textbf{Phase}} & \multicolumn{2}{c}{\textbf{Activities}} & \multirow{2}{*}{\textbf{Duration}}\\
    \cmidrule{2-3}
         & \centering \textbf{Educators} & \centering \textbf{Children} & \\
    \midrule
        Introduction & Introduction to research aims and key terms & Introduction to research aims and key terms & 10--15 mins\\
        Discussion of AWE, robotics, and interactive narratives & Identification of AWE goals \newline Discussion of personas \newline Thoughts on robotics for AWE \newline Impressions of interactive narratives & Self-rated knowledge of AW and perceived importance of AWE \newline Discussion of personas \newline Thoughts on robotics for AWE \newline Impressions of interactive narratives & 20--30 mins\\
        Demonstrations & Demonstrations of MiRo and Qoobo \newline Feedback on robots & Demonstrations of MiRo and Qoobo \newline Feedback on robots & 25--35 mins\\
        Design & Draw-and-write \newline Discussion of narrative ideas & Brainstorming on attributes of pets \newline Draw-and-write \newline Branching storyboards & 25--35 mins\\
        Feedback & Presentation of ideas to group \newline Plenary discussion \newline Thank you and goodbyes & Presentation of ideas to group \newline Voting on favorite robot \newline Thank you and goodbyes & 25--30 mins\\
    \bottomrule
    \end{tabularx}
    \label{tab:workshop_phases}
\end{table*}

\paragraph{Introduction}

We explained the aims of the workshop, key terms, like zoomorphic robot and interactive narrative, and the target intervention group: mainstream schoolchildren between 8 and 12 years old learning about pets.

\paragraph{Discussion of AWE, Robotics, and Interactive Narratives}

The first aspect we discussed was the background of and need for AWE. The children started by rating their own knowledge about AW and their perception of the importance of learning about AW, which they did by placing stickers on printed 5-point Likert scales.
With the educators we discussed the goals of AWE, including their desired learning outcomes for a session about pet mammals and important scenarios surrounding pet mammals children might need to be educated about. We also discussed currently used tools and approaches. 
Both sets of participants were then presented with three personas (see Section~\ref{sec:personas}). Participants were asked to think about why each persona might need to learn about AW and what activities might help them do that. Their answers were noted by the researcher leading the small group.

The second aspect discussed was the participants' perspective on using robotics in AWE. For the educators this included discussion of the pros and cons of the current robotics toolkit as well as envisioned benefits and challenges going forwards.

The final aspect discussed was the use of interactive narratives to structure an educational intervention. This discussion was started by giving more information on interactive narratives and an example of a choose-your-own-adventure novel. Educators were asked for their impressions of its use in AWE.
The children were asked some questions about their experience with and impressions of interactive narratives.

\paragraph{Demonstrations}

\begin{figure*}
    \centering
    \hspace{1em}
    \begin{subfigure}[b]{0.45\linewidth}
         \centering
         \includegraphics[height=5cm]{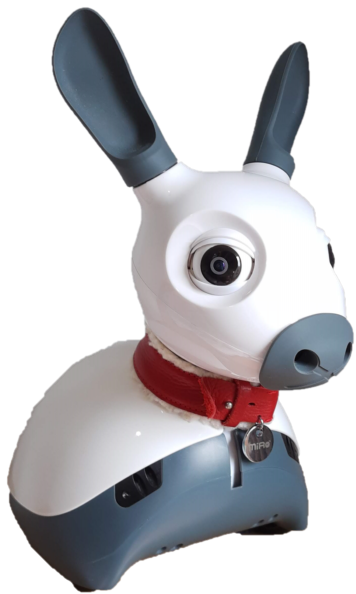}
         \caption{MiRo}
         \label{fig:miro}
    \end{subfigure}
    \hfill
    \begin{subfigure}[b]{0.45\linewidth}
         \centering
         \includegraphics[height=4cm]{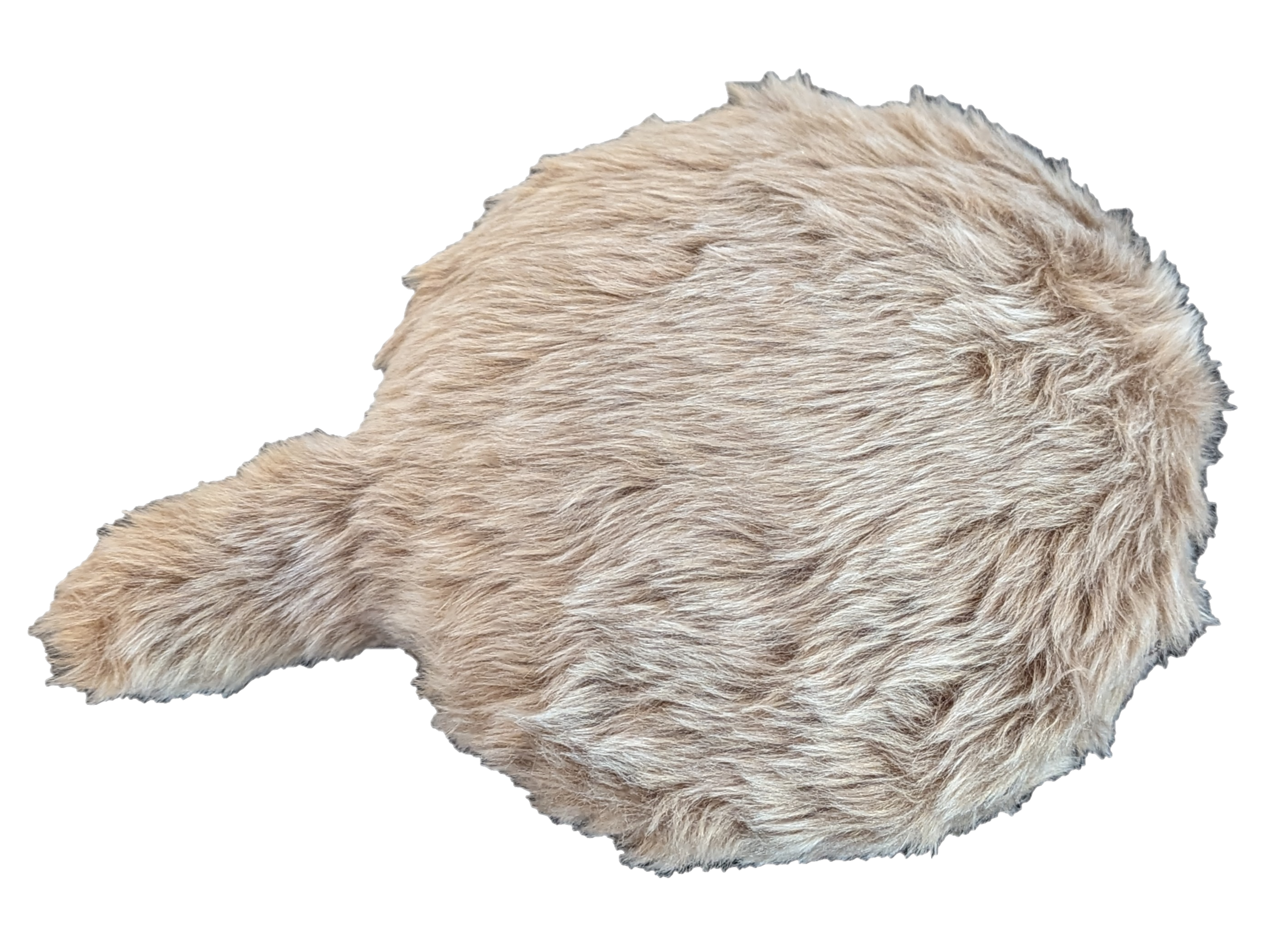}
         \caption{Qoobo}
         \label{fig:qoobo}
    \end{subfigure}
    \hspace{1em}
    \caption{Zoomorphic robots demonstrated to participants}
    \label{fig:demorobots}
\end{figure*}

Participants were shown MiRo\footnote{\url{https://www.miro-e.com/}} and Qoobo\footnote{\url{https://qoobo.info/index-en/}} (Fig.~\ref{fig:demorobots}).
MiRo was running a modified version of its `demo controller' that shows exaggerated displays of emotion through the ears, eyes, and tail, used in research into children's perceptions of mental attributes~\cite{voysey2022influence}. Qoobo was switched on, so the tail movements changed in response to sound and touch. After a description of the sensors, actuators, and capabilities, participants were invited to interact with the zoomorphic robot and ask questions.
They then returned to their small groups to give feedback on the zoomorphic robots, including likes, dislikes, what made them animal-like or not, and potential for use in AWE.

\paragraph{Design}

Following the demonstrations and feedback, the educators went back into their small groups to start the design session. This began with sketching and annotating individual designs.
They were then asked for some ideas for storylines or narrative framing to use.

The children's design session was formed of three main activities: brainstorming about the attributes and behavior of pets; designing a robot pet; and completing a branching storyboard. The rationale for each of these activities is given in Section~\ref{sec:designprompts}.

\paragraph{Feedback}

The final portion of the educators' workshop involved the two small groups presenting their combined ideas to each other. The workshop then united as a full group to discuss some final points, including suggestions for how the robot could be used with groups of children, how the robot could be made customizable, and thoughts on representing a specific mammal versus a generic mammal with the robot. There was also time for participants to raise any other points they wanted to discuss.

The children's feedback session involved each small group sharing one of their ideas with the rest of the group.
As a final activity, participants voted on their favorite robot of MiRo and Qoobo. The lead researcher then wrapped up the workshop, thanking the participants and allowing time for questions.


\subsection{Analysing the Workshop Data}

The scribes' notes and the materials produced in the workshops were analyzed using the constant comparative method of grounded theory~\cite{glaser1965constant}, where codes and categories are continually compared and redefined. This method works particularly well for comparing and contrasting data across and within two groups. Content analysis was also used for the drawings. 
We used the same rule of thumb as \citet{sanoubari2021remotecodesign} where something was considered a theme if it was seen in data from at least 20\% of participants (that is, 5 or more children or 2 or more educators) or included in the scribes' notes for more than one small group.
The themes were validated by a researcher separate from the project who coded 20\% of the data. Any major differences were discussed and resolved.

\section{Results}

This section covers the themes identified from the workshop data, split into perspectives on AWE and robotics, demonstrations, attributes of pets, pet robot designs, and narrative designs.
Participants are identified by E for educator or C for child and a number.

\subsection{Discussion of AWE, Robotics, and Interactive Narratives}

\subsubsection{AWE}

Table~\ref{tab:education_aims} shows six themes identified from the discussion with educators about the desired learning outcomes, along with some example quotes.
These six learning objectives targeted by the educators were: awareness of animal needs and proper care; awareness of animal emotions; understanding of animal behavior; awareness of appropriate handling; respect for animal boundaries; and awareness of the responsibilities of pet ownership.
Educators felt that addressing these areas would lead to increased compassion and empathy for animals.
This was linked as a consequence of understanding animal sentience and emotions, rather than directly targeted (“it all goes together”).
Additionally, children's responses to the personas touched on many of the same themes---see the final column of Table~\ref{tab:education_aims}.
The children also highly appreciated the importance of AWE, with 4 children rating it as important (the second strongest option) and 20 rating it as very important (the strongest option) to teach children about AW.
Children's self-rated knowledge of AW varied (poor: $N{=}2$, fair: $N{=}2$, good: $N{=}7$, very good: $N{=}6$, excellent: $N{=}7$).

\begin{table*}[]
    \centering
    \begin{threeparttable}
        \caption{Learning objectives in AWE}
        \label{tab:education_aims}
        \rowcolors{1}{white}{TableGray}
        \begin{tabularx}{\linewidth}{>{\raggedright\arraybackslash}p{0.01\linewidth}>{\raggedright\arraybackslash}p{0.19\linewidth}XX}
            \toprule
            \showrowcolors
            \textbf{\#} & \textbf{Description} & \textbf{Quotes from Educators} & \textbf{Quotes from Children}\\
            \midrule
            1 & Awareness of animal needs and proper care & “needs of an animal”, “five freedoms”\tnote{a}, “shouldn’t be neglecting pets’ needs” & “make sure he knows how to care for his cat”, “nice place to sleep in”, “how to keep a cat active” \\
        
            2 & Awareness of animal emotions & “teaching about animal emotions”, “happiness, sadness, fear, pain, anger” & “he has a cat and wants to learn about its emotions”, “her rabbit might have different emotions” \\
        
            3 & Understanding of animal behavior & “understanding why she was bitten”, “didn’t recognize the signs” & “she wants to find out why the dog bit her”, “his cat might bite if stressed” \\
            
            4 & Awareness of appropriate handling & “gentle handling, not grabbing” & “play in a gentle way”, “how to interact with a dog in a safe way” \\
        
            5 & Respect for animal boundaries & “not disturbing them when they’re eating or sleeping”, “have its own space” & “maybe if cat doesn’t want to cuddle he should know” \\
            
            6 & Awareness of responsibilities of pet ownership & “responsibilities of having a pet”, “broader responsibility”, “need to microchip”& “if sick, take to vet”, “how to do chores so that they get a pet” \\
            \bottomrule
            \hiderowcolors
        \end{tabularx}
        \begin{tablenotes}
            \item [a] The five freedoms for animals are: freedom from hunger or thirst; freedom from discomfort; freedom from pain, injury, or disease; freedom to express normal behavior; and freedom from fear and distress~\cite{fawc1993fivefreedoms}.
        \end{tablenotes}
    \end{threeparttable}
\end{table*}

Educators use a range of tools and activities in their programs including quizzes, memory games, scenario cards, and an emotion wheel. They have also used a modular robotic kit (Section~\ref{sec:bkg_awe}). Children suggested similar activities, like games and quizzes, as well as a writing exercise for Bonnie to engage with the feelings she experienced when bitten by a dog. Another of their ideas was learning through experience, through interactions with a pet in a safe environment or with a substitute like a zoomorphic robot or stuffed toy. 
Educators also explained that there are different education programs for different needs. Most children participate in the mainstream program, Animal WISE, but children who have shown concerning behavior towards animals may be recommended for another one-to-one program.

The educators stressed the importance of educating all children about AW, not just those with pets, as children without pets are still likely to interact with animals, whether in the wild or at a friend's house. The children also recognized this, saying that Charlie (the persona without pets) ``might have friends who have animals''.

\subsubsection{Robotics}

Educators were positive about the interactivity of zoomorphic robots, especially the ability to respond to sensory input, and the adaptability of zoomorphic robots to different situations and activities. They also anticipated high engagement from children due to the novelty of zoomorphic robots.
However, they were concerned that the excitement of new technology could be a distraction from the main purpose of the activity, i.e., learning about pets.
They also had mixed feelings about how realistic a zoomorphic robot could and needed to be for children to relate it to a pet and apply skills learned using the zoomorphic robot to real life.

A major concern for the educators was technical issues, centered around setup, connecting robots and controllers, and battery life---all of which they have experienced with the robotic toolkit. Issues like these require educators to adapt programs on the fly, wasting the limited time they have with classes and causing them stress.

Educators have experienced rough handling of toys and robots, which is a reason for using zoomorphic robots over live animals in schools. However, they expressed concerns about children being rough towards the zoomorphic robot if there is no immediate negative feedback from it. These concerns related both to damage to the zoomorphic robot and to the precedent it could set around acceptable behavior towards pets.

Children thought zoomorphic robots could be used to teach about pet care, including how and how not to interact with pets, and pets' responses to different actions. Two groups noted zoomorphic robots would let children learn through experience in a low-stakes environment and children who had had bad experiences with pets might be less scared of zoomorphic robots. 

\subsubsection{Interactive Narratives}

The educators liked the focus interactive narratives would give to the consequences of actions. For this to be effective, they stressed that the children would need to be able to see a clear response and have the opportunity to understand the link between the action and the response. They also felt that giving children control over the narrative would engage them and make them more likely to respond. The importance of interaction design for groups was raised as key to preventing arguments among children about narrative choices.

Children expressed interest in interactive narratives and several of them had experience of interactive narratives through choose-your-own-adventure books, video games, and interactive shows on streaming platforms. Some had been involved in collaborative storytelling games which, while not exactly the same, provide a similar sense of agency. They described interactive narratives as fun, interesting, and exciting because of the control it would afford them and the unpredictability of the story. However, some children preferred the structure provided by a traditional story.

\subsection{Demonstrations}
\label{sec:demos}


\subsubsection{MiRo}

Both educators and children liked Miro's interactivity, particularly its visible response to touch. They also both liked the lights on the back that represent emotions through color, which educators felt was an easy-to-understand way to communicate emotions. However, two groups of children noted the lights were not animal-like.
The educators further appreciated the expressive facial features (like the moving eyelids and ears) and how these were used in reaction to the environment. The facial features played a part in children connecting MiRo to pets, as children in three groups referenced its nose, ears, and eyes to compare MiRo to rabbits and dogs. 
The educators disliked toy-like aspects of MiRo, particularly the noises and wheels, and neither educators nor children liked MiRo's plastic and `robotic' appearance, which one child described as “creepy” and an educator described as “[not] nice to touch”.

\subsubsection{Qoobo}

Educators expressed that Qoobo had the feel of an animal due to its heartbeat, fur, and the responsiveness of its wagging tail. However, they struggled to see a use for Qoobo in AWE because it was distinctly unlike pets in certain ways due to its limited actuation and lack of facial features, meaning it cannot produce expressions. These were aspects that the children noted made Qoobo unlike pets, but they still felt Qoobo was more similar to pets than MiRo. This seemed to be due to its furry appearance, with one child saying “[you] might have to look twice to tell [Qoobo] is a robot”. Children also liked the feel of the fur and described it as “fluffy” ($N{=}8$). There were also multiple comments ($N{=}16$), often several per group, calling Qoobo “cute”.

\subsubsection{Use in AWE}

Both groups thought the zoomorphic robots could be used to show children how to treat animals.
Educators proposed two key modifications to the zoomorphic robots for use in AWE, namely the inclusion of negative reactions and more expressive features.
The negative reactions would help children understand how not to treat pets (e.g., being loud, rough handling), and the eyes, ears, tail, and noises would be used to communicate the zoomorphic robot's displeasure.
Another suggestion was items for the zoomorphic robot to interface with that could provide an objective for the interaction, like collecting items that pets need.
The educators also felt the zoomorphic robots' non-specific mammalian nature would make it hard for children to relate them to pets. However, children compared the zoomorphic robots to different pets, including their own, and seemed willing to view the zoomorphic robots as animals, albeit unusual ones.

\subsubsection{Preferences}

Children slightly preferred Qoobo to MiRo (14 votes to 10 votes). Educators were not asked explicitly for a preference, but scribes noted one for MiRo during group discussions.

\subsection{Attributes of Pets}
\label{sec:pet_attributes}

In brainstorming about pets' appearances, children listed features common to many species (e.g., eyes, ears, tails, and legs) and some specific to species (“udders” [C6], “feet like a chicken” [C15]). Most children who described pets' size wrote “small” ($N{=}8$), but some ($N{=}2$) used “big”.
Children made aesthetic judgments about pets' appearances, mostly that they were “cute” ($N{=}13$), though one child said “scary” [C20]. They also ascribed positive traits to appearances, like “happy” [C11], “friendly” [C1], and “calming” [C23].

Children referred to different textures that pets could be, including soft ($N{=}14$), fluffy ($N{=}13$), furry, scaly, rough ($N{=}2$), smooth, silky, and leathery ($N{=}1$). They also mentioned signs of vital processes, such as warmth ($N{=}3$) and a heartbeat. Children also extended this prompt about tactile sensation to the quality of tactile interaction (“cuddly”, “snuggly”, “cosy”) and how a pet makes them feel (“calming” [C18], “peaceful” [C1], “makes you feel good” [C24]).

Children listed a variety of sounds made by pets, including those specific to different species (dog noises, cat noises) and the sounds associated with the movement of feet and tails.
They used onomatopoeia to describe other noises.
The children also highlighted the volume of sounds; four said soft or quiet, one said loud, one said both. Some children also viewed the sounds as a method of communication, conveying emotion or a meaning, like “feed me”.

To describe how pets acted, children listed different movements, emotions (from simple to complex), and energy levels. They also gave examples of needs, like food, water, and sleep, and preferences that pets might display. Children pointed out  pets' responses to the environment and the actions of others. Pets' behavior was sometimes related to themselves and the home (“live with me” [C23], “cuddle with me” [C16], “jump on my lap” [C3, C4]).

Children noted the variety between different species commonly owned as pets, seen through individuals mentioning mutually exclusive textures (“fluffy, scaly, leathery” [C10]) or noises from different species (“barks, miaows”, $N{=}4$). However, children who seemed to focus on a single species still referred to variety within the pet, for example, that they could be happy and angry ($N{=}6$). This linked to an idea of temporality---that pets did things “sometimes” ($N{=}4$).

\subsection{Pet Robot Designs}

\begin{figure*}
    \centering
    \resizebox{\linewidth}{!}{
        \begin{tikzpicture}
            [%
            grow cyclic,
            every node/.style=
                {rounded rectangle,
                draw, 
                very thick, 
                inner xsep=0.75em,
                inner ysep=0.75em,
                },
            edge from parent/.style=
                {draw, black, very thick},
            level 1/.style={
                level distance = 2.5cm
            },
            level 2/.style={
                level distance = 1.5cm,
            },
            sibling distance=2cm
            ]
        
        \pgfdeclarepattern{
          name=children,
          parameters={\gapsize,\childangle,\hatchlinewidth},
          bottom left={\pgfpoint{-.1pt}{-.1pt}},
          top right={\pgfpoint{\gapsize+.1pt}{\gapsize+.1pt}},
          tile size={\pgfpoint{\gapsize}{\gapsize}},
          tile transformation={\pgftransformrotate{\childangle}},
          code={
            \pgfsetlinewidth{\hatchlinewidth}
            \pgfpathmoveto{\pgfpoint{-.1pt}{-.1pt}}
            \pgfpathlineto{\pgfpoint{\gapsize+.1pt}{\gapsize+.1pt}}
            \pgfusepath{stroke}
          }
        }
        \pgfdeclarepattern{
          name=educators,
          parameters={\gapsize,\educatorangle,\hatchlinewidth},
          bottom left={\pgfpoint{-.1pt}{-.1pt}},
          top right={\pgfpoint{\gapsize+.1pt}{\gapsize+.1pt}},
          tile size={\pgfpoint{\gapsize}{\gapsize}},
          tile transformation={\pgftransformrotate{\educatorangle}},
          code={
            \pgfsetlinewidth{\hatchlinewidth}
            \pgfpathmoveto{\pgfpoint{-.1pt}{-.1pt}}
            \pgfpathlineto{\pgfpoint{\gapsize+.1pt}{\gapsize+.1pt}}
            \pgfusepath{stroke}
          }
        }
        \pgfdeclarepattern{
          name=both,
          parameters={\gapsize,\childangle,\hatchlinewidth},
          bottom left={\pgfpoint{-.1pt}{-.1pt}},
          top right={\pgfpoint{\gapsize+.1pt}{\gapsize+.1pt}},
          tile size={\pgfpoint{\gapsize}{\gapsize}},
          tile transformation={\pgftransformrotate{\childangle}},
          code={
            \pgfsetlinewidth{\hatchlinewidth}
            \pgfpathmoveto{\pgfpoint{-.1pt}{-.1pt}}
            \pgfpathlineto{\pgfpoint{\gapsize+.1pt}{\gapsize+.1pt}}
            \pgfpathmoveto{\pgfpoint{-.1pt}{\gapsize+.1pt}}
            \pgfpathlineto{\pgfpoint{\gapsize+.1pt}{-.1pt}}
            \pgfusepath{stroke}
          }
        }
        
        \tikzset{
          gap size/.store in=\gapsize,
          child angle/.store in=\childangle,
          educator angle/.store in=\educatorangle,
          hatch line width/.store in=\hatchlinewidth,
          gap size=8pt,
          child angle=90,
          educator angle=0,
          hatch line width=.5pt,
        }
        
            
            \node[inner ysep=1.5em, inner xsep=1.5em, fill=gray!70] {\Large\textbf{Robot design}} [clockwise from=115, sibling angle=135]
            child {node [preaction={fill, violet!20}, pattern=both, pattern color=violet!25, yshift=-0.25cm]{\large \textbf{Appearance}} [clockwise from=190, sibling angle=60]
                child {node[preaction={fill, violet!20}, pattern=both, pattern color=violet!25, yshift=-0.5cm, xshift=-1.5cm] {Surface}}
                child {node[preaction={fill, violet!20}, pattern=both, pattern color=violet!25, xshift=-2cm] {Facial features}}
                child {node[preaction={fill, violet!20}, pattern=both, pattern color=violet!25] {Legs vs wheels}}
                child {node[preaction={fill, violet!20}, pattern=both, pattern color=violet!25, xshift=2cm] {Inspiration}}
                }
            child {node [preaction={fill, violet!20}, pattern=both, pattern color=violet!25, xshift=1.5cm, yshift=-1cm] {\large \textbf{Behaviour}} [clockwise from=40, sibling angle=60]
                child {node[preaction={fill, violet!20}, pattern=both, pattern color=violet!25, xshift=0.5cm, yshift=0.5cm] {Reactions} [clockwise from=120, sibling angle=60]
                    child {node[preaction={fill, red!20}, pattern=educators, pattern color=red!25, xshift=0cm, yshift=0cm] {Negative}}
                    child {node[preaction={fill, violet!20}, pattern=both, pattern color=violet!25, xshift=0.75cm, yshift=0cm] {To touch}}
                    child {node[preaction={fill, red!20}, pattern=educators, pattern color=red!25, xshift=1.5cm, yshift=0.1cm] {To environment}}
                }
                child {node[preaction={fill, violet!20}, pattern=both, pattern color=violet!25, xshift=2.5cm] {Communication} [clockwise from=30, sibling angle = 60]
                    child {node[preaction={fill, violet!20}, pattern=both, pattern color=violet!25, xshift=1.3cm] {Noises}}
                    child {node[preaction={fill, red!20}, pattern=educators, pattern color=red!25, xshift=2.4cm] {Signaling features}}
                }
                child {node[preaction={fill, violet!20}, pattern=both, pattern color=violet!25, xshift=1cm, yshift=-0.25cm] {Emotions}}
                child {node[preaction={fill, violet!20}, pattern=both, pattern color=violet!25, xshift=-0.5cm, yshift=-0.5cm] {Movement}}
            }
            child {node[preaction={fill, violet!20}, pattern=both, pattern color=violet!25, xshift=-1.5cm, yshift=-0.5cm] {\large \textbf{Features}} [counterclockwise from=140, sibling angle=90]
                child {node[preaction={fill, violet!20}, pattern=both, pattern color=violet!25, xshift=-1.5cm] {Accessories} [counterclockwise from=150, sibling angle = 60]
                    child {node[preaction={fill, red!20}, pattern=educators, pattern color=red!25, xshift=-0.8cm] {Needs}}
                    child {node[preaction={fill, cyan!20}, pattern=children, pattern color=cyan!25, xshift=-0.8cm] {Toys}}
                }
                child {node[preaction={fill, violet!20}, pattern=both, pattern color=violet!25, xshift=-1.0cm] {Identity} [counterclockwise from=190, sibling angle = 80]
                    child {node[preaction={fill, violet!20}, pattern=both, pattern color=violet!25, xshift=-0.6cm] {Name}}
                    child {node[preaction={fill, cyan!20}, pattern=children, pattern color=cyan!25, xshift=-0.4cm, yshift=0.1cm] {Sex}}
                }
            };
        
            \node[preaction={fill, red!20}, pattern=educators, pattern color=red!25, text width = 1.8cm, text centered, xshift=6.5cm, yshift=4.3cm]{Educators};
            \node[preaction={fill, cyan!20}, pattern=children, pattern color=cyan!25, text width = 1.8cm, text centered, xshift=9.25cm, yshift=4.3cm]{Children};
            \node[preaction={fill, violet!20}, pattern=both, pattern color=violet!25, text width = 1.8cm, text centered, xshift=12cm, yshift=4.3cm]{Both};
            
        \end{tikzpicture}
    }
    \caption{Developed thematic map for robot design, showing three main themes and sub-themes, which are colour-coded by the group of participants they were observed in.}
    \label{fig:mindmap_robot}
\end{figure*}
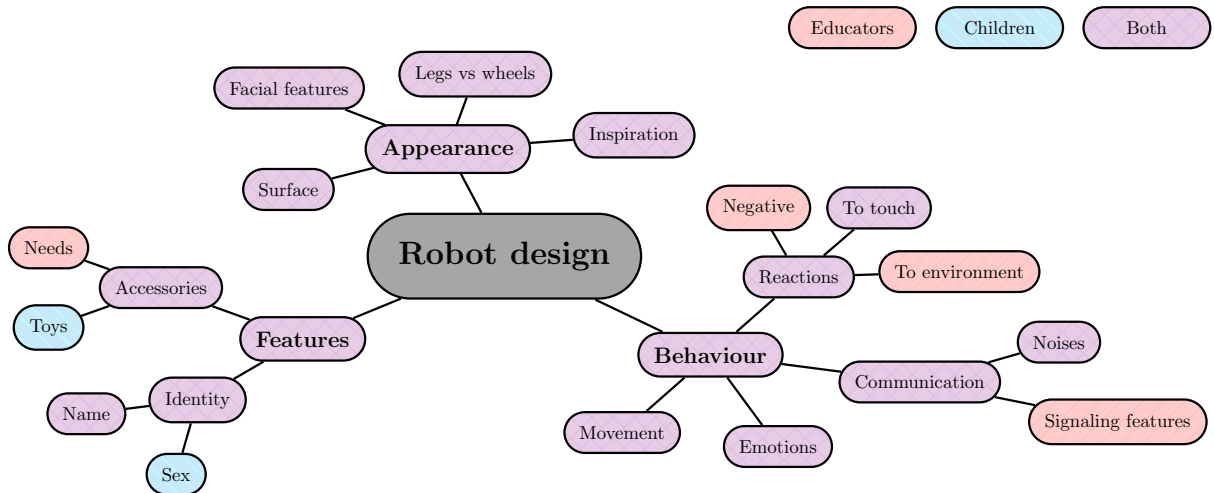

\begin{figure*}
    \centering
    \resizebox{0.9\linewidth}{!}{%
    \begin{tikzpicture}
        \node[anchor=south west,inner sep=0] (image) at (0,0) {\includegraphics[angle=-90,clip, trim=1cm 0cm 0cm 0cm, width=\linewidth]{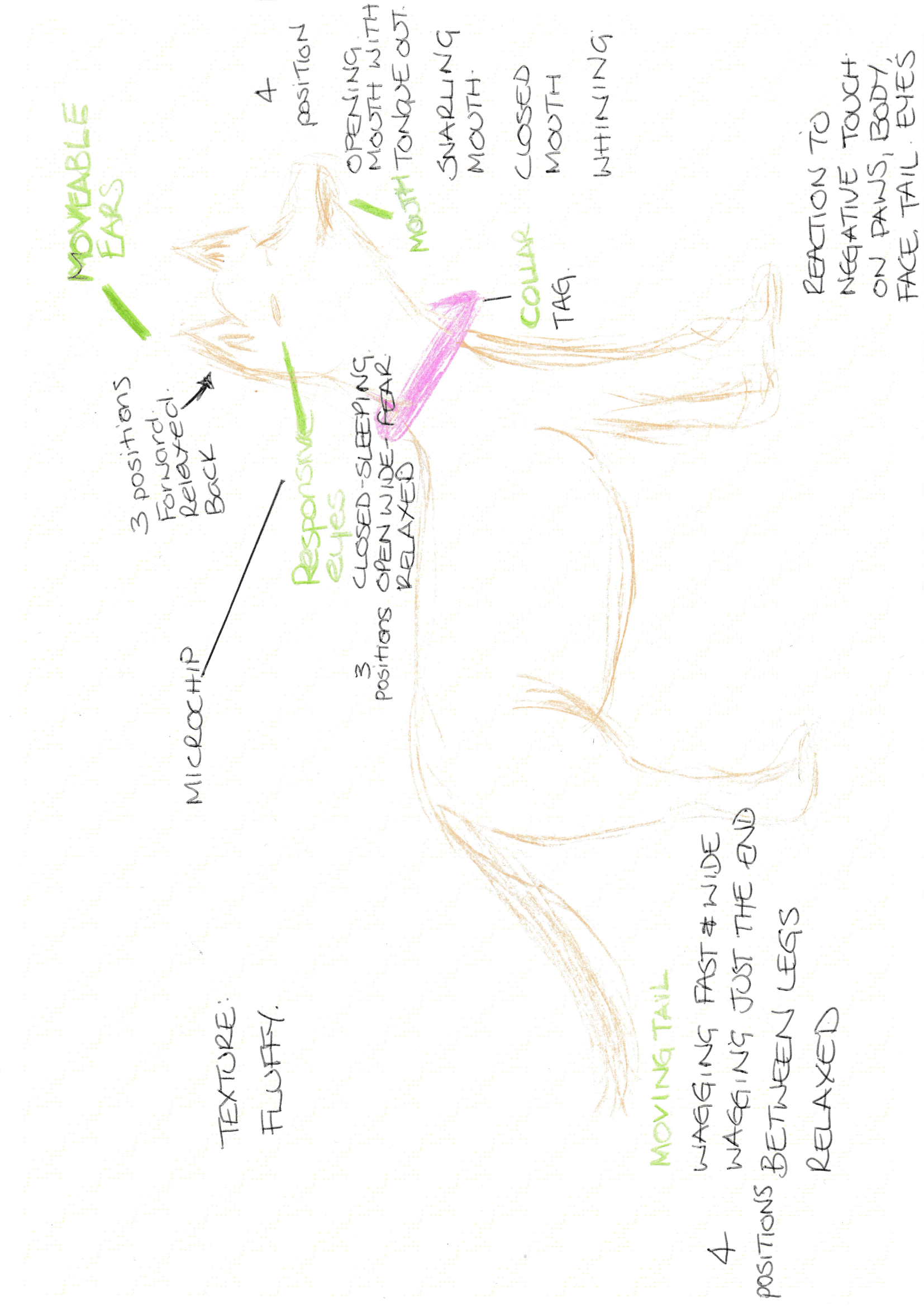}};
    \begin{scope}[x={(image.south east)},y={(image.north west)}]
        \draw[red,densely dotted,thick,rounded corners] (0.58,0.77) rectangle (0.94,0.96);
        \draw[red,densely dotted,thick,rounded corners] (0,0.09) rectangle (0.4,0.31);
        \draw[red,densely dotted,thick,rounded corners] (0.45,0.55) rectangle (0.74,0.71);
        \draw[red,densely dotted,thick,rounded corners] (0.8,0.34) rectangle (1.0,0.75);
    \end{scope}
    \end{tikzpicture}}
    \resizebox{0.9\linewidth}{!}{%
    \begin{tikzpicture}
        \node[anchor=south west,inner sep=0] (image) at (0,0) {\includegraphics[angle=-90,clip, trim=1cm 0cm 0cm 0cm,width=\linewidth]{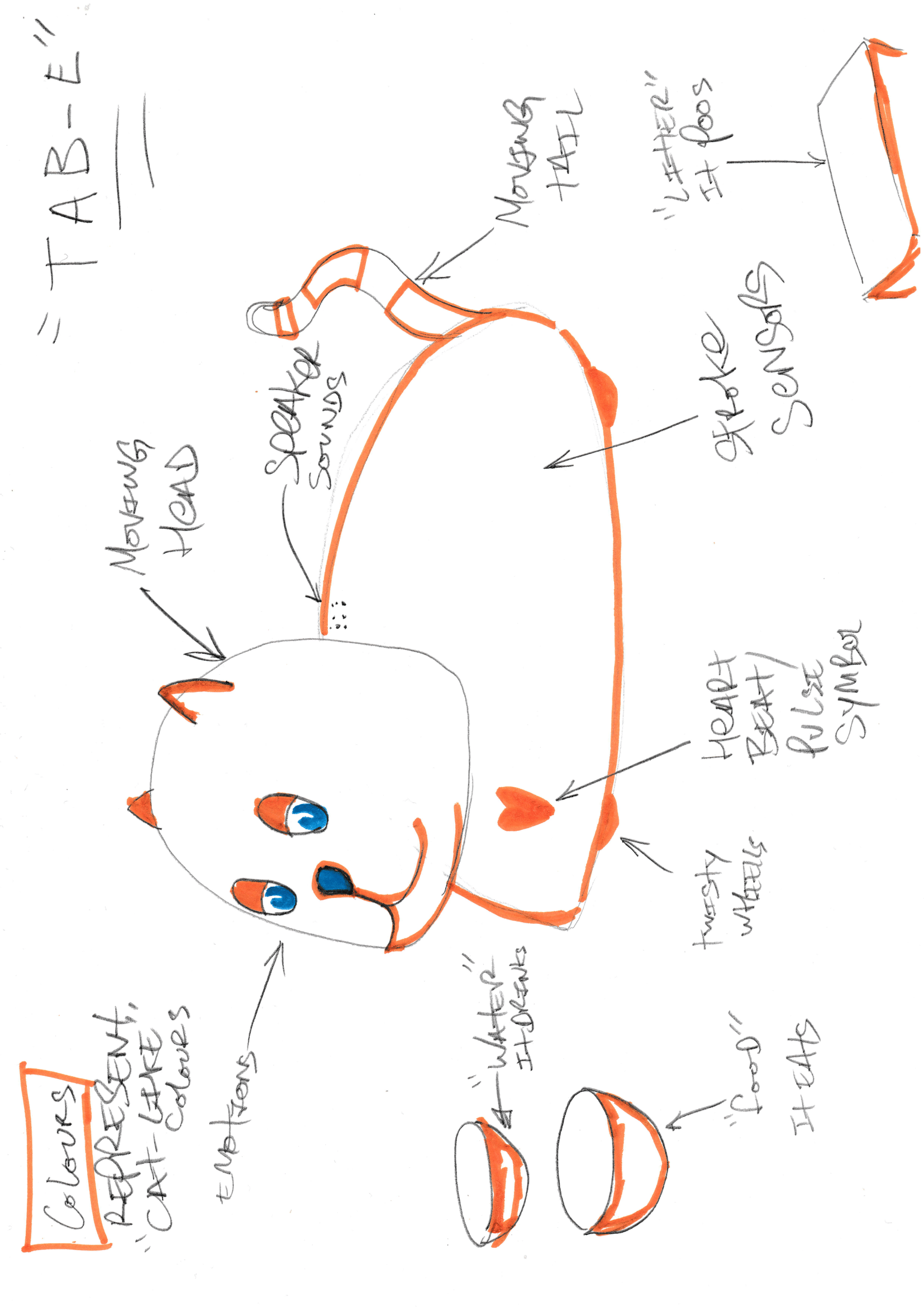}};
    \begin{scope}[x={(image.south east)},y={(image.north west)}]
        \draw[red,densely dotted,thick,rounded corners] (0.81,0.38) rectangle (0.94,0.50);
        \draw[red,densely dotted,thick,rounded corners] (0.07,0.75) rectangle (0.21,0.82);
    \end{scope}
    \end{tikzpicture}}
    \caption{Examples of educators' designs [E3, E6]. Movable signaling features are outlined in red.}
    \label{fig:educator_designs}
\end{figure*}

\begin{figure*}
    \centering
    \resizebox{0.48\linewidth}{!}{%
    \begin{tikzpicture}
        \node[anchor=south west,inner sep=0] (image) at (0,0) {\includegraphics[clip, trim=3.2cm 4cm 3.2cm 6.5cm, width=0.4\linewidth]{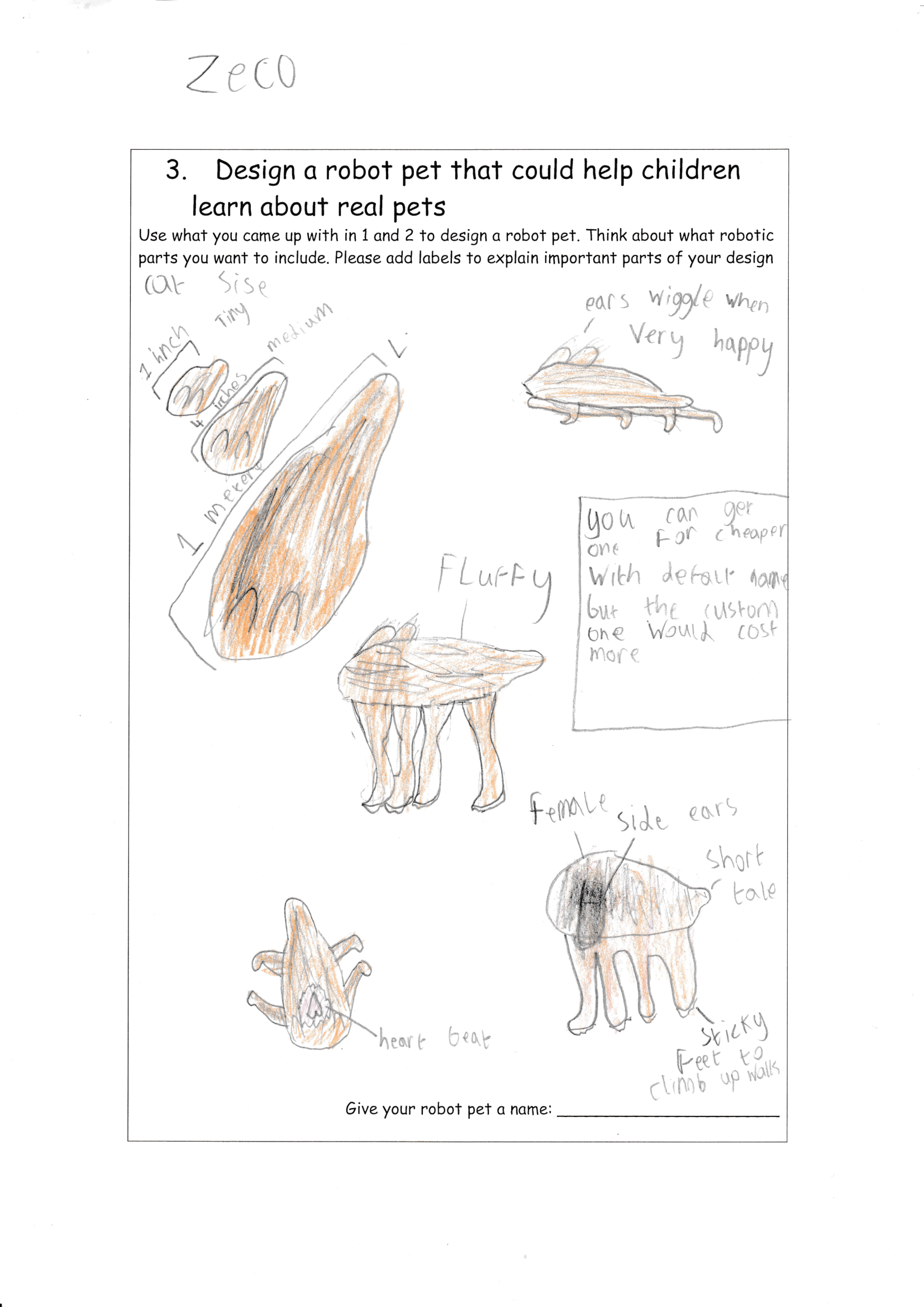}};
    \begin{scope}[x={(image.south east)},y={(image.north west)}]
        \draw[red,densely dotted,thick,rounded corners] (0.67,0.91) rectangle (1.0,1.0);
    \end{scope}
    \end{tikzpicture}}
    \hspace{0.01\linewidth}
    \resizebox{0.48\linewidth}{!}{%
    \begin{tikzpicture}
        \node[anchor=south west,inner sep=0] (image) at (0,0) {\includegraphics[clip, trim=3.0cm 4cm 3.8cm 6.5cm, width=0.4\linewidth]{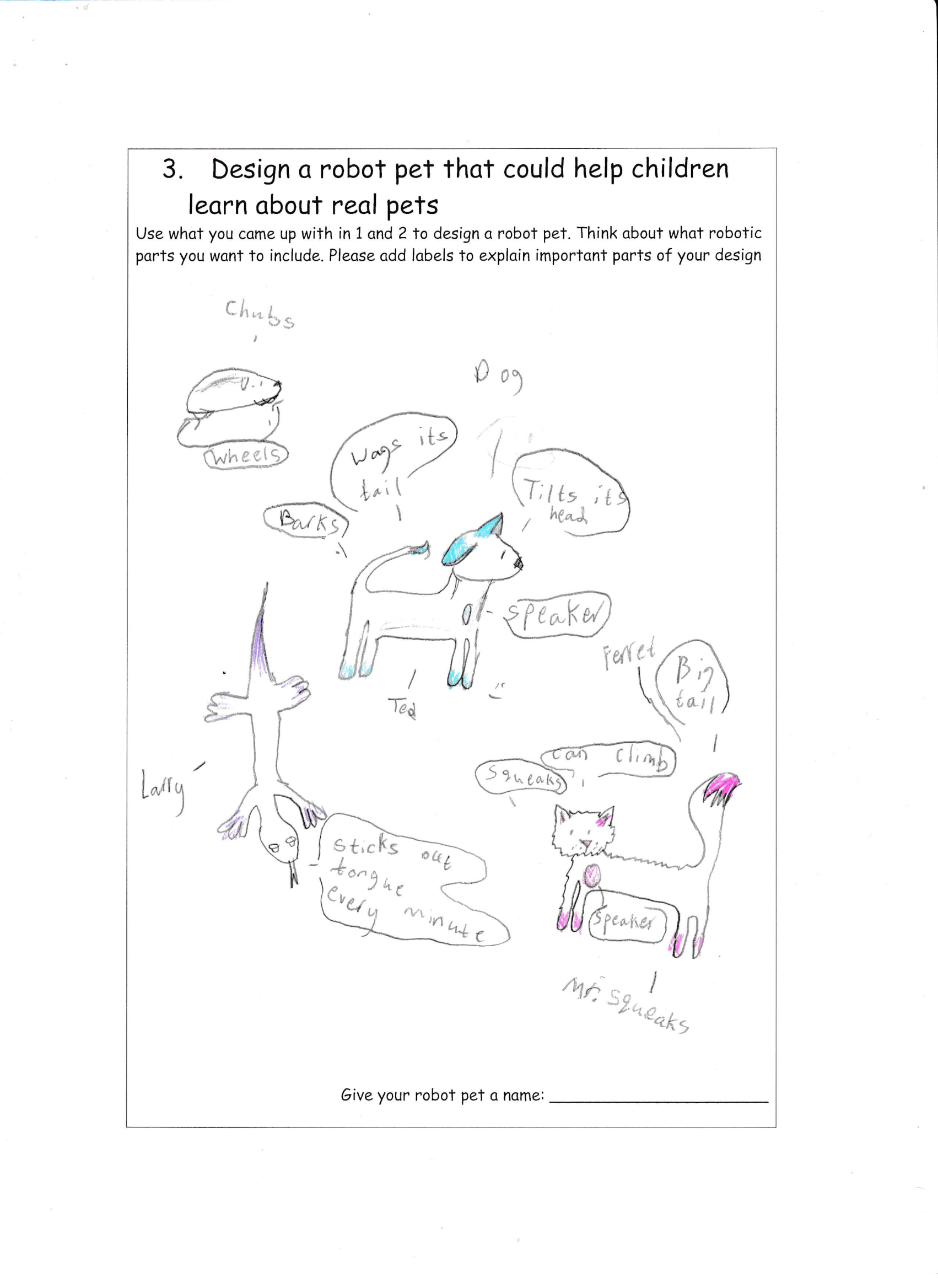}};
    \begin{scope}[x={(image.south east)},y={(image.north west)}]
        \draw[red,densely dotted,thick,rounded corners] (0.32,0.73) rectangle (0.51,0.84);
        \draw[red,densely dotted,thick,rounded corners] (0.3,0.2) rectangle (0.59,0.36);
    \end{scope}
    \end{tikzpicture}}
    \resizebox{0.48\linewidth}{!}{%
    \begin{tikzpicture}
        \node[anchor=south west,inner sep=0] (image) at (0,0) {\includegraphics[clip, trim=3.2cm 4cm 3.2cm 6.5cm, width=0.4\linewidth]{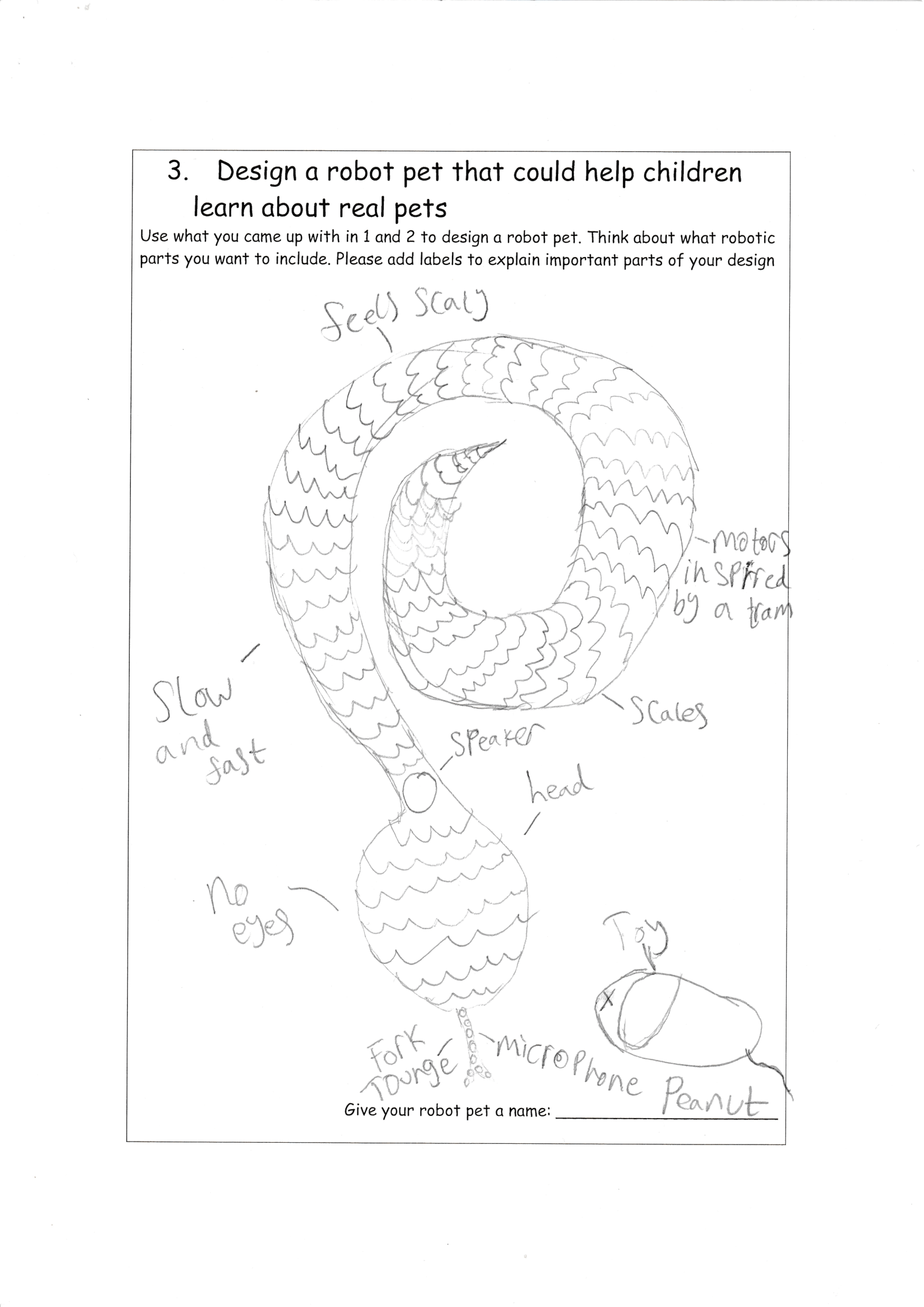}};
    \begin{scope}[x={(image.south east)},y={(image.north west)}]
    \end{scope}
    \end{tikzpicture}}
    \hspace{0.01\linewidth}
    \resizebox{0.48\linewidth}{!}{%
    \begin{tikzpicture}
        \node[anchor=south west,inner sep=0] (image) at (0,0) {\includegraphics[clip, trim=3.2cm 4cm 3.6cm 6.5cm, width=0.4\linewidth]{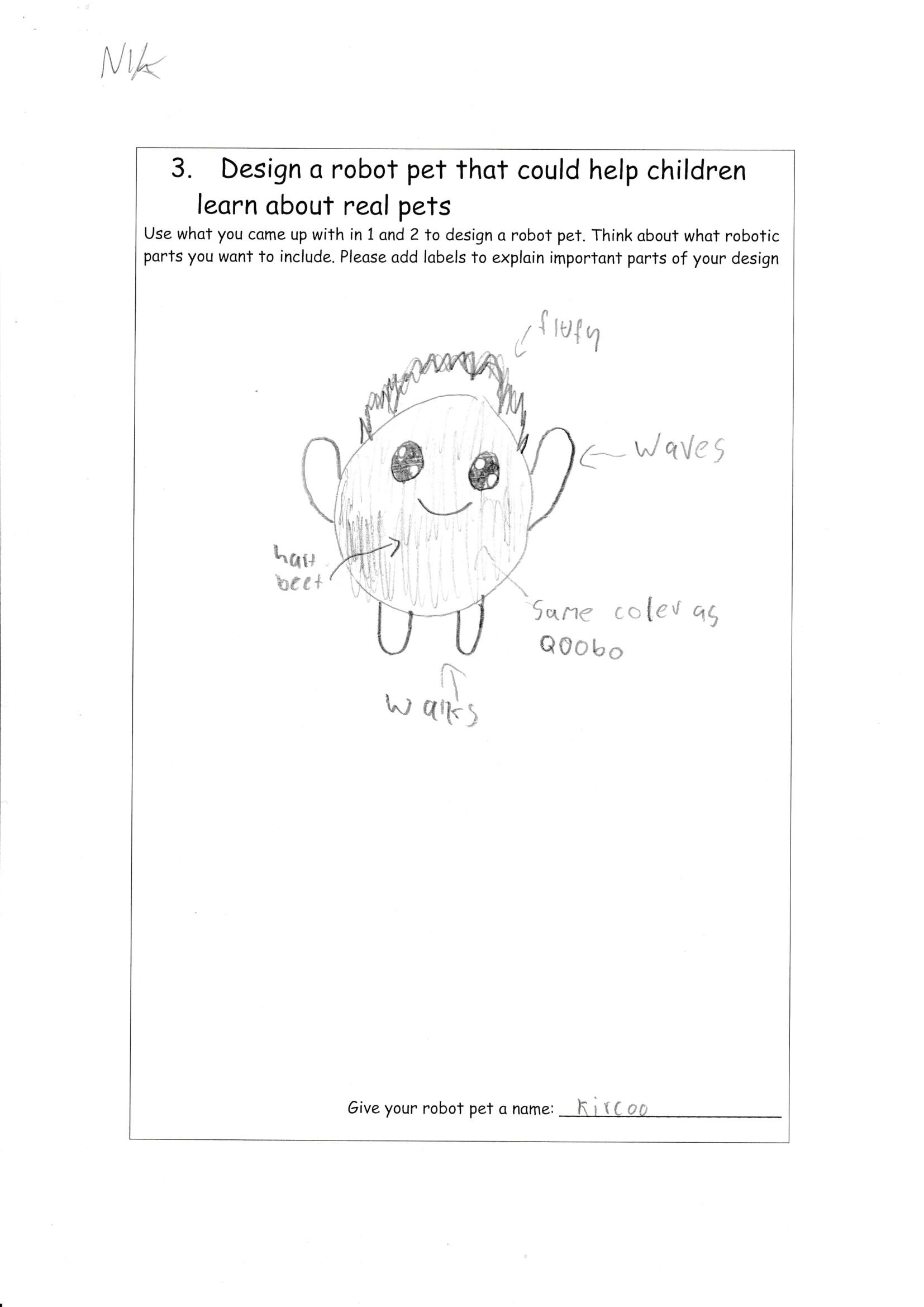}};
    \begin{scope}[x={(image.south east)},y={(image.north west)}]
    \end{scope}
    \end{tikzpicture}}
    \caption{Examples of children's designs [C10, C11, C12, C18]. Movable signaling features are outlined in red.}
    \label{fig:children_designs}
\end{figure*}

The thematic map in Fig.~\ref{fig:mindmap_robot} shows three high-level themes from the data, described as appearance, behavior, and features.
Examples of educators' and children's designs are shown in Fig.~\ref{fig:educator_designs} and Fig.~\ref{fig:children_designs} respectively. The full collection is available online$^{\ref{footnote:osf}}$.

Two children designed humanoid robots instead of pet robots, so these were excluded from analysis. Some children created multiple designs---features from all of them were collated for analysis.

\subsubsection{Appearance}
\label{sec:robotdesign_appearance}

\paragraph{Surface}

Similarities were seen in and across the groups for the zoomorphic robot's surface texture and color.
A design was counted as furry if it was labeled with soft, furry, or fluffy, or if the drawing style conveyed a furry texture through soft, feathery, or bumpy edges. 
Three educators and 12 children explicitly labeled the texture as furry, but a total of 7 educator and 15 child designs appeared furry through their drawing style.

The groups used color in a similar way. The drawings were mostly in pencil, but when color was used on the zoomorphic robot's body it was often brown or orange (natural fur colors) or labeled as ``same coler as Qoobo'' [\textit{sic}] [C18] (light brown). Other colors tended to be used for accessories, like collars and toys [E3, E8, C4, C19, C24], and to highlight small features, like eyes and ears [E6, C11, C15].

\paragraph{Legs vs wheels}
It did not seem to matter whether the zoomorphic robot used legs or wheels to move around as there were designs in both groups that used legs and designs that used wheels; roughly half of each group incorporated wheels. 
However, an educator raised the concern that children seem to link wheels to racing, which makes a zoomorphic robot more toy-like, so wheels need to be unobtrusive or ``hidden'' [E11].

\paragraph{Facial features}

Ten of the educators included eyes, ears, a mouth, and a nose in their design; the final design omitted only a mouth [E1]. These were also the four most common facial features included by children in their designs; eyes, ears, a mouth, and a nose were included by 82\%, 73\%, 55\%, and 50\% of children respectively.
Both groups included species-specific facial features, e.g., whiskers on cat-inspired zoomorphic robots [E1, E8, E11, C15, C19].
Five educators and six children labeled a facial feature with the associated sense, i.e., the eyes had sight/cameras and ears had hearing/microphones.

\paragraph{Inspiration}

All bar one of the educators' designs were analogous to specific pets currently targeted by AWE (5 dogs, 4 cats, 1 rabbit). One educator design was specific but with interchangeable ears for future years of the program [E1]. In contrast, the children tended to merge elements from different species and create non-specific animals. The children also drew on a wider range of species for their designs, including large mammals (cow, horse, goat), birds, and reptiles. The majority of designs ($N{=}17$) did seem to be inspired by mammals, as they had tails ($N{=}17$) and were furry ($N{=}15$). 

A couple of the children's designs were anthropomorphic, showing features or behaviors that are more human-like than animal-like. For example, the fourth design in Fig.~\ref{fig:children_designs} [C18] shows a bipedal creature waving, which is not a common animal behavior. Another anthropomorphic aspect is the smile, which was seen in a second child's design [C23]; animals may bare their teeth or appear to smile, but this does not generally indicate happiness or pleasure~\cite{meints2010prevent}. No anthropomorphic features were seen in the educators' designs.

\subsubsection{Behavior}
\label{sec:robotdesign_behavior}

\paragraph{Reactions}

A key component of 82\% of the educators' designs was negative reactions, which they framed as vital for teaching children what behavior is not appropriate towards pets. Negative reactions included pulling away from interactions, either removing the body part [E4] or retreating completely [E5], and displaying warning signs, such as growling [E1, E2, E7], hissing [E1, E8, E11], or baring teeth [E3, E4, E5, E7]. Educators wanted negative reactions even if they could be scary as, ultimately, the interaction is safe.

Tactile interactions are an important way children interact with pets. Five children included tactile interaction in their design, either through the zoomorphic robot sensing touch or `liking' tactile interaction (``likes to get petted'' [C24] or ``like[s] cuddles'' [C16]). Educators stressed the zoomorphic robot needed to distinguish gentle from rough touch, like pulling and poking, as well as sensing when children touch sensitive areas, like the paws, tail, and face, particularly around the eyes.

Educators also wanted the zoomorphic robot to be sensitive to its environment. This was broadly presented in a negative context, like the zoomorphic robot fleeing from loud noises, rough handling, and aggressive movements. While nine children included sensors that could detect these things, only C14 included corresponding reactions like ``reacts if people are shouting and get's scared'' [\textit{sic}] and ``when cold he shakes and wimpers'' [\textit{sic}].

\paragraph{Communication}

Children's designs frequently included noises or a speaker to produce them ($N{=}12$). The educators wanted the noises to fit the species being taught about---barking, whining, and growling for dogs and hissing, miaowing, and purring for cats---as opposed to MiRo's `mammalian voice'. Similarly, when children gave detail on the noise, it matched the apparent inspiration, e.g., barking for a dog. The educators wanted the noises to convey the pet's emotions and ask for attention when appropriate. No participant designs showed the zoomorphic robot communicating with speech.

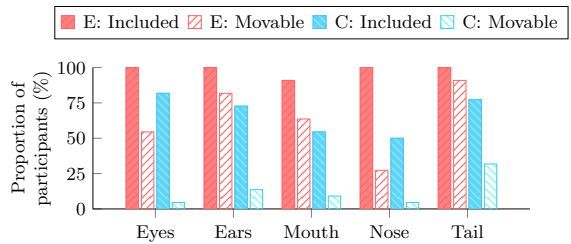
\begin{figure}[t]
    \centering
    \resizebox{\linewidth}{!}{
        \begin{tikzpicture}
            \pgfplotsset{every tick label/.append style={font=\small}}
            \pgfplotsset{every axis label/.append style={font=\small}}
                    
            \begin{axis} [ybar = .05cm,
                height=4cm,
                width=9cm,
                axis x line*=bottom,
                axis y line*=left,
                ylabel={Proportion of \\ participants (\%)},
                y label style = {align=center, at={(-0.14,0.5)}, anchor=center},
                bar width = 6pt,
                ymin = 0,
                ymax = 100,
                xtick = data,
                xticklabels={Eyes, Ears, Mouth, Nose, Tail},
                ytick = {0, 25, 50, 75, 100},
                enlarge x limits = {abs = .8},
                legend columns = 4,
                legend style={at={(0.5, 1.2)}, anchor=south, font=\scriptsize, /tikz/column sep=2pt},
                legend image code/.code={
                    \draw [#1] (0cm,-0.1cm) rectangle (0.22cm,0.12cm); },
            ]
            
            \addplot [preaction={fill, red!50}, draw = red!60, pattern=north east lines, pattern color = red!60] coordinates {(1,100) (2,100) (3,90.9) (4,100) (5,100)};
            \addplot [draw = red!60, pattern=north east lines, pattern color = red!60] coordinates {(1,54.5) (2,81.8) (3,63.6) (4,27.3) (5,90.9)};
            \addplot [preaction={fill, cyan!50}, draw = cyan!60, pattern=north west lines, pattern color = cyan!60] coordinates {(1,81.8) (2,72.7) (3,54.5) (4,50.0) (5,77.3)};
            \addplot [draw = cyan!60, pattern=north west lines, pattern color = cyan!60] coordinates {(1,4.5) (2,13.6) (3,9.1) (4,4.5) (5,31.8)};
             
            \legend {E: Included, E: Movable, C: Included, C: Movable};
             
            \end{axis}
            
        \end{tikzpicture}
    }
    \caption{Proportion of educators (E) and children (C) who included signaling features. Examples of movable signaling features are shown in Fig.~\ref{fig:educator_designs} and Fig.~\ref{fig:children_designs}}
    \label{fig:signaling_features}
\end{figure}

The educators highlighted the way a pet's facial features and tail can be used to communicate, which we refer to as signaling features. A large proportion labeled these features with specific positions or movements they wanted, examples of which can be seen in Fig.~\ref{fig:educator_designs}. For example, six educators labeled the eyes as able to blink or take different positions, from closed to relaxed, to wide open. Similar specifications were attached to the other features (ears: $N{=}9$, mouth: $N{=}7$, nose: $N{=}3$, tail: $N{=}10$).
When we remarked on this level of detail, educators explained that people must notice and understand the movement and positions of these features in order to have positive interactions with pets, since they communicate the internal state of a pet. 
Many children included the same features but were much less likely to provide further specification (Fig.~\ref{fig:signaling_features}). The feature they most commonly labeled as movable was the tail (by 32\%) where the description given was mostly ``wags''---compare this to E3 who gave four movements for the tail (Fig.~\ref{fig:educator_designs}).

\paragraph{Emotions}

The educators' designs suggested the zoomorphic robot would have five main emotions: happiness, sadness, anger, fear, and pain (Table~\ref{tab:education_aims}). These emotions would be influenced by the environment and interactions, and would be displayed through noises, body language, and/or colored lights in the eyes or back.
Additionally, five of the children's designs implied the zoomorphic robot would have emotions or ``mood''. Some labels specified reactions in response to emotions (“makes \textit{mmmmmm} sound when happy” [C9], “ears wiggle when very happy” [C10],  “when he’s scared his ears go back” [C14]). Two children included colored lights to display mood [C2, C4].

\paragraph{Movement}

Both educators and children described the zoomorphic robot's movement on a range of scales, from the large, like moving around in space, through the medium, like moving the head and wagging the tail, to the small, like twitching the nose and blinking the eyelids. A number of educators and children also included trembling [E1, E7, C14] and a heartbeat [E6, E11, C10, C17, C18, C19].

\subsubsection{Features}

\paragraph{Accessories}

Three of the educators' designs included peripheral items for the zoomorphic robot. Examples of these were a bed, a litter tray, food and water, and a tablet to display the zoomorphic robot's internal state.
The educators also explained that peripheral items could be used to split children into smaller groups doing different activities, rather than one large group.
Three of the children included peripheral items in their design, but while the educators focused on pets' needs, all the children's items were toys. An accessory common to both groups was a collar (seen in three educator and four child designs).

\paragraph{Identity}

Children were prompted to name their zoomorphic robot and 19 did so. There were names inspired by Qoobo and MiRo (e.g., Meqzo, Qute, Birbo) and pet-like ones (e.g., Bailey, Coco, Mickey). Designs with names in the former category were often more robot-like, with features like propellers and solar panels. Designs in the latter group tended to closely resemble specific pets, like cats or dogs.
Four children used the name to personalize the zoomorphic robot by the zoomorphic robot coming with a collar or birth certificate. Naming was also relevant for behavior, as two children wanted the zoomorphic robot to come when its name was called [C1, C14]. 
Only one educator named their design, ``TAB-E'' [E6], but both groups discussed the zoomorphic robot responding to its name.

59\% of the children referred to their zoomorphic robot with male pronouns, 32\% with gender-neutral pronouns (\textit{it}/\textit{they}/a mix of the two), and 9\% did not use pronouns (pronoun usage was cross-referenced with their storyboard if not discernible from their design alone).
No children used female pronouns for their zoomorphic robots, but two children said their zoomorphic robot could come as a female version [C10, C14]. 45\% of the educators used \textit{it} to refer to their zoomorphic robot and the rest did not use pronouns in their drawings.

\subsection{Narrative Designs}

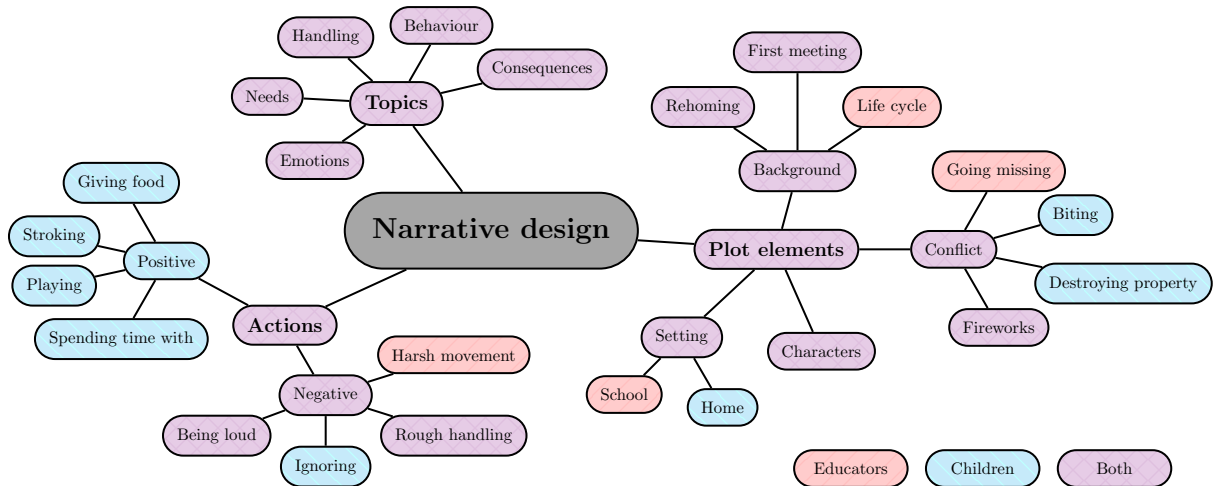
\begin{figure*}[t]
    \centering
    \resizebox{\linewidth}{!}{
        \begin{tikzpicture}
            [%
            grow cyclic,
            every node/.style=
                {rounded rectangle,
                draw, 
                very thick, 
                inner xsep=0.75em,
                inner ysep=0.75em,
                },
            edge from parent/.style=
                {draw, black, very thick},
            level 1/.style={
                level distance = 3.5cm
            },
            level 2/.style={
                level distance = 1.7cm,
            },
            sibling distance=2cm
            ]

        \tikzset{
          gap size/.store in=\gapsize,
          child angle/.store in=\childangle,
          educator angle/.store in=\educatorangle,
          hatch line width/.store in=\hatchlinewidth,
          gap size=8pt,
          child angle=90,
          educator angle=0,
          hatch line width=.5pt,
        }
            
            \node[inner ysep=1.5em, inner xsep=1.5em, fill=gray!70] {\Large\textbf{Narrative design}} [clockwise from=115, sibling angle=105]
            child {node[preaction={fill, violet!20}, pattern=both, pattern color=violet!25, xshift=-0.5cm, yshift=-0.5cm] {\large \textbf{Topics}} [clockwise from=225, sibling angle=50]
                child {node[preaction={fill, violet!20}, pattern=both, pattern color=violet!25, yshift=0cm, xshift=-0.5cm] {Emotions}}
                child {node[preaction={fill, violet!20}, pattern=both, pattern color=violet!25, xshift=-1.0cm] {Needs}}
                child {node[preaction={fill, violet!20}, pattern=both, pattern color=violet!25, xshift=-0.5cm] {Handling}}
                child {node[preaction={fill, violet!20}, pattern=both, pattern color=violet!25, xshift=0.5cm] {Behaviour}}
                child {node[preaction={fill, violet!20}, pattern=both, pattern color=violet!25, xshift=1.5cm] {Consequences}}
                }
            child {node [preaction={fill, violet!20}, pattern=both, pattern color=violet!25, xshift=2.5cm, yshift=-1cm] {\large \textbf{Plot elements}} [clockwise from=75, sibling angle=75]
                child {node[preaction={fill, violet!20}, pattern=both, pattern color=violet!25] {Background} [clockwise from=150, sibling angle=60]
                    child {node[preaction={fill, violet!20}, pattern=both, pattern color=violet!25, xshift=-0.5cm, yshift=0.5cm] {Rehoming}}
                    child {node[preaction={fill, violet!20}, pattern=both, pattern color=violet!25, yshift=0.8cm] {First meeting}}
                    child {node[preaction={fill, red!20}, pattern=educators, pattern color=red!25, xshift=0.5cm, yshift=0.5cm] {Life cycle}}
                }
                child {node[preaction={fill, violet!20}, pattern=both, pattern color=violet!25, xshift=2cm] {Conflict} [clockwise from=75, sibling angle = 50]
                    child {node[preaction={fill, red!20}, pattern=educators, pattern color=red!25, xshift=0.5cm] {Going missing}}
                    child {node[preaction={fill, cyan!20}, pattern=children, pattern color=cyan!25, xshift=1cm] {Biting}}
                    child {node[preaction={fill, cyan!20}, pattern=children, pattern color=cyan!25, xshift=2cm] {Destroying property}}
                    child {node[preaction={fill, violet!20}, pattern=both, pattern color=violet!25, xshift=0.5cm] {Fireworks}}
                }
                child {node[preaction={fill, violet!20}, pattern=both, pattern color=violet!25, xshift=0.5cm, yshift=-0.5cm] {Characters}}
                child {node[preaction={fill, violet!20}, pattern=both, pattern color=violet!25, xshift=-0.5cm, yshift=-1cm] {Setting} [clockwise from=300]
                    child {node[preaction={fill, cyan!20}, pattern=children, pattern color=cyan!25, xshift=0cm] {Home}}
                    child {node[preaction={fill, red!20}, pattern=educators, pattern color=red!25, xshift=0cm] {School}}
                }
            }
            child {node[preaction={fill, violet!20}, pattern=both, pattern color=violet!25, xshift=-4cm, yshift=1.5cm] {\large \textbf{Actions}} [counterclockwise from=150, sibling angle=150]
                child {node[preaction={fill, cyan!20}, pattern=children, pattern color=cyan!25, xshift=-1cm, yshift=0.5cm] {Positive} [counterclockwise from=105, sibling angle = 50]
                    child {node[preaction={fill, cyan!20}, pattern=children, pattern color=cyan!25, xshift=-0.5cm] {Giving food}}
                    child {node[preaction={fill, cyan!20}, pattern=children, pattern color=cyan!25, xshift=-0.8cm, yshift=-0.2cm] {Stroking}}
                    child {node[preaction={fill, cyan!20}, pattern=children, pattern color=cyan!25, xshift=-0.8cm, yshift=0.2cm] {Playing}}
                    child {node[preaction={fill, cyan!20}, pattern=children, pattern color=cyan!25, xshift=-0.5cm] {Spending time with}}
                }
                child {node[preaction={fill, violet!20}, pattern=both, pattern color=violet!25, xshift=0cm] {Negative} [counterclockwise from=210, sibling angle = 60]
                    child {node[preaction={fill, violet!20}, pattern=both, pattern color=violet!25, xshift=-0.8cm] {Being loud}}
                    child {node[preaction={fill, cyan!20}, pattern=children, pattern color=cyan!25, xshift=0cm, yshift=0.2cm] {Ignoring}}
                    child {node[preaction={fill, violet!20}, pattern=both, pattern color=violet!25, xshift=1.2cm] {Rough handling}}
                    child {node[preaction={fill, red!20}, pattern=educators, pattern color=red!25, xshift=1.2cm] {Harsh movement}}
                }
            };
        
            \node[preaction={fill, red!20}, pattern=educators, pattern color=red!25, text width = 1.8cm, text centered, xshift=7.5cm, yshift=-5cm]{Educators};
            \node[preaction={fill, cyan!20}, pattern=children, pattern color=cyan!25, text width = 1.8cm, text centered, xshift=10.25cm, yshift=-5cm]{Children};
            \node[preaction={fill, violet!20}, pattern=both, pattern color=violet!25, text width = 1.8cm, text centered, xshift=13cm, yshift=-5cm]{Both};
            
        \end{tikzpicture}
    }
    \caption{Developed thematic map for narrative design, showing three main themes and sub-themes, which are colour-coded by the group of participants they were observed in.}
    \label{fig:mindmap_narrative}
\end{figure*}

\begin{figure*}
    \centering
    \resizebox{0.9\linewidth}{!}{\includegraphics[clip, trim=1cm 2cm 1cm 1cm, width=\linewidth]{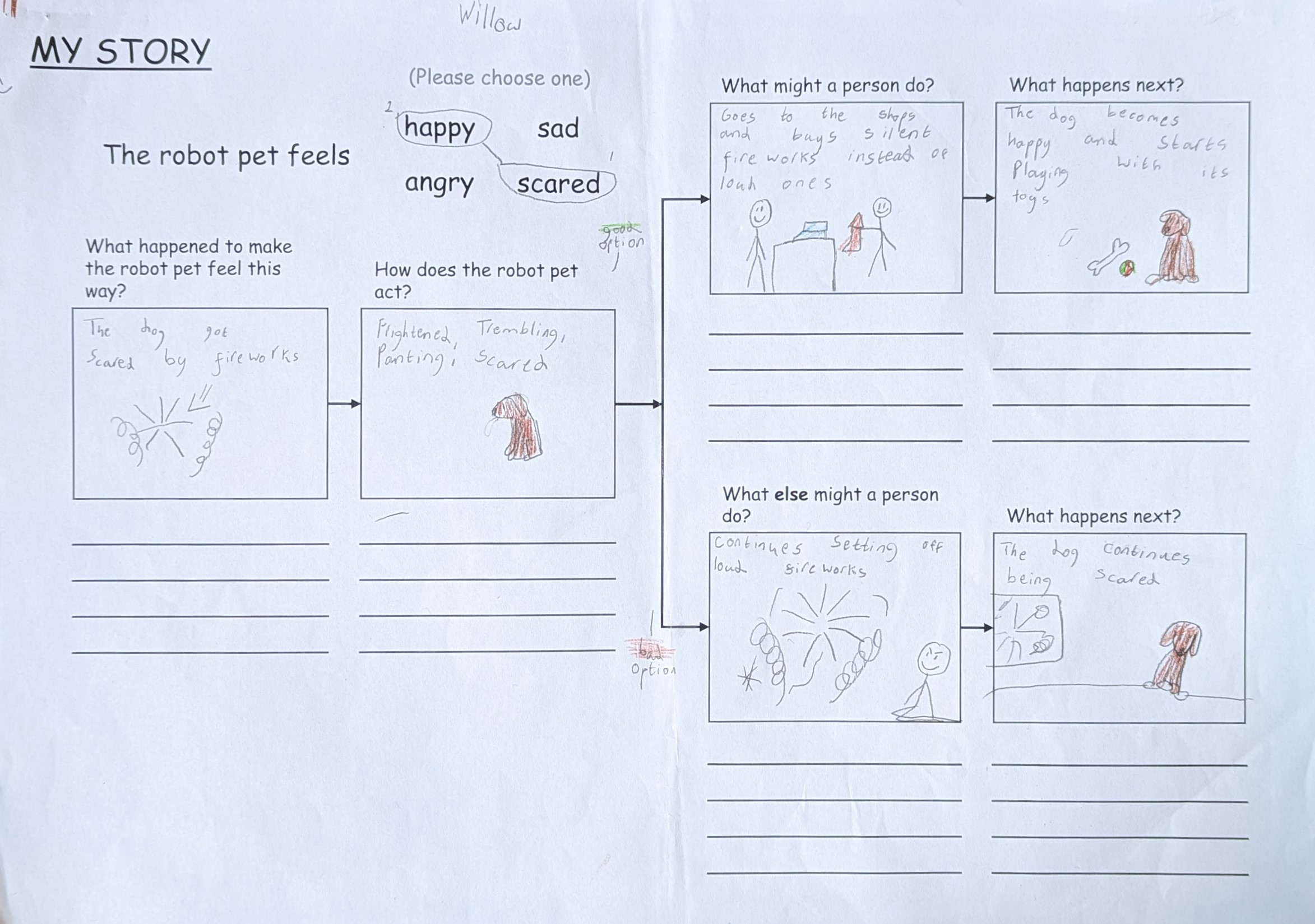}}
    \resizebox{0.9\linewidth}{!}{\includegraphics[clip, trim=1cm 2cm 1cm 1cm, width=\linewidth]{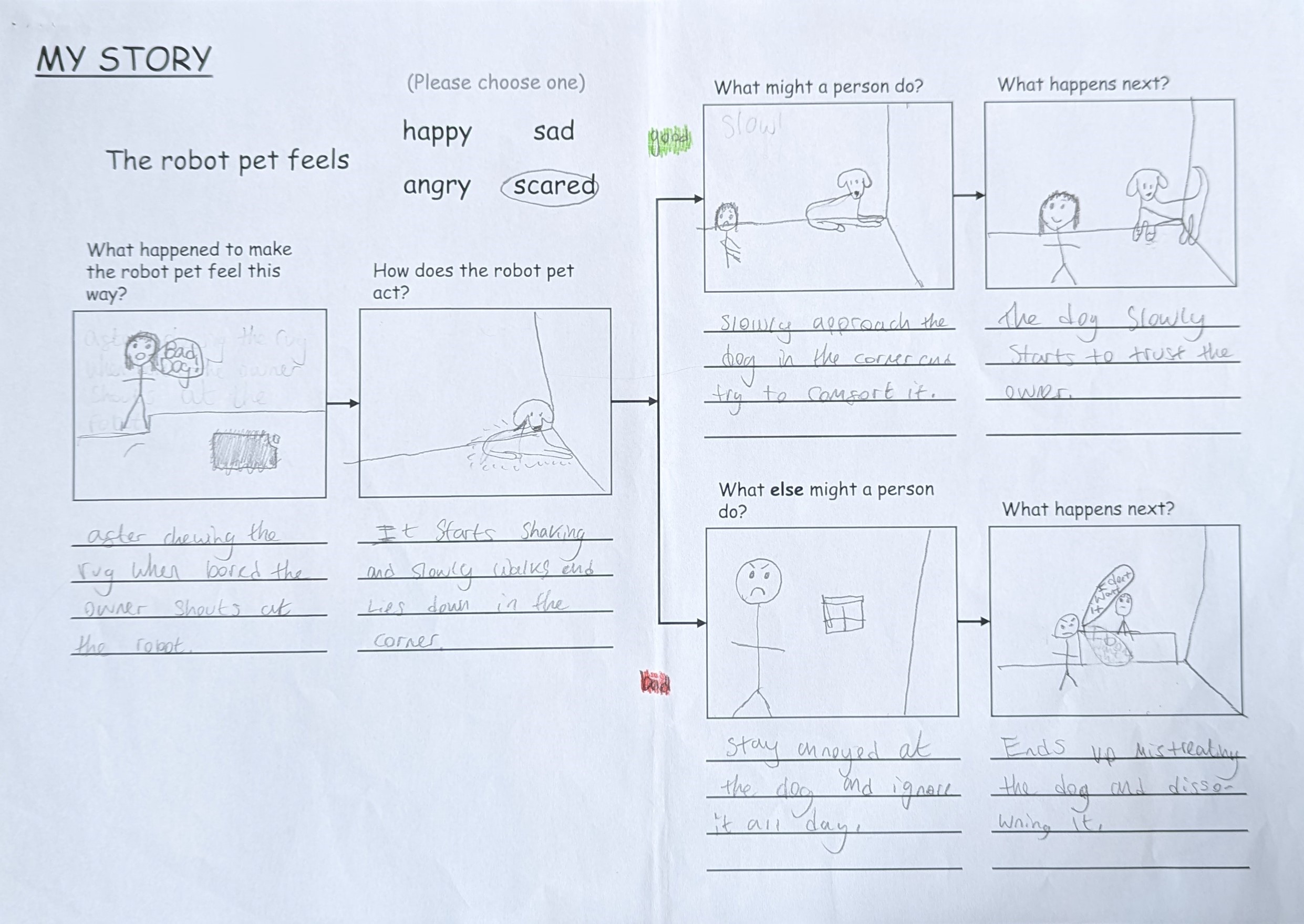}}    \caption{Examples of completed storyboards [C11, C15]}
    \label{fig:storyboard_example}
\end{figure*}

This section presents the themes coded to the educators' small group discussions about the narrative for the robot and the children's storyboards.
Fig.~\ref{fig:mindmap_narrative} shows three themes, identified as topics, plot elements, and actions. 

As with the robot designs, only the narratives from the 22 children who designed a robot for AWE were analyzed. Children could choose from four emotions that the robot was experiencing at the start of the story: happy, sad, angry, or scared. 10 children chose happy, 1 chose sad, 2 chose angry, and 9 chose scared. 17 storyboards were complete, 2 were almost complete (missing 1 panel), and 3 were half complete (missing 2--3 panels). 18 children used writing and drawing to complete the storyboards, 3 used only writing, and 1 used only drawing.

\subsubsection{Topics}
\label{sec:narrative_topics}

This concept related to what the key topics and takeaways of the narrative would be.

\paragraph{Emotions}

Educators talked about the way the robot could convey emotions through appropriate reactions to events in a story. The goal would be for the children to understand what emotion the robot was experiencing and why that emotion had occurred.
All the children's storyboards included an emotion due to the prompt, but half of them ($N{=}11$) had an additional, different emotion as the outcome of the story as a result of the events in the story.

\paragraph{Needs}
The educators looked for ways the narrative could be used to convey animals' needs, as well as emotions. They also highlighted providing needs as an easy way children could make choices in the narrative, such as the type of food to provide or where to take it if it were ill. One group of educators suggested a story around a journey to collect what an animal needs, using devices to track progress, like a heart that fills up.
Five children's stories were coded with the theme of needs, four of which focused on providing food, but two of which extended to environmental and behavioral needs, such as the need to play or spend time outside [C12, C18]. The food in the children's stories was a mixture of regular pet food and explicitly ``robotic'' food, like batteries. 

\paragraph{Handling}
Following on from the robot designs, the educators highlighted the importance of teaching children handling skills.
They suggested the narrative could include instances of handling, acted out by the children, and the robot would react in different ways, with positive and negative reactions as appropriate. 
Handling was also a common theme in the children's storyboards; nine of them pictured or described the person handling the robot animal in some way, mostly petting or stroking it but sometimes carrying it ($N{=}3$).
Additionally, three narratives used the branching to show the differences between rough and gentle handling and their effect on the animal [C1, C2, C16]. The stories showed that grabbing an animal or petting it in the wrong place (e.g., the tail) would distress the animal and might make it run away, contrasted with gentle stroking that would make the animal happy and facilitate bonding.

\paragraph{Behavior}
Educators wanted the narrative to have displays of animal behavior in different situations. They would then use these to teach children how to interpret these behaviors and react appropriately. 
Several storyboards showed that some children already have a good grasp of animal behavior. Examples of this included the potential causes of emotional states, like loud noises causing fear, but also how the pet robot would display the emotion. For example, several children who had designed a robot pet inspired by dogs gave correct examples of dog behavior for fear, like cowering, shaking, and whimpering.
The educators further stressed that it would be key that the behavior of the robot matched what an animal would do at each point in the story and be appropriate to the inspiration (i.e., rabbit behaviors for a rabbit-like robot).

\paragraph{Consequences}
Educators wanted the narrative to show children the consequences of their actions. One such example involved the robot being initially shy but through ``the class making the right choices'' the personality would change to more confident and playful.
They also wanted negative outcomes and reactions to poor choices, such as giving a dog chocolate or not getting a pet vaccinated.
The children's storyboards also showed a connection from people's actions to the outcome of the story.
Children frequently ($N{=}10$) presented polarized storylines, where there were distinct positive and negative options and the outcome of the story varied accordingly, even though the prompts were neutral (``What might a person do?'' and ``What else might a person do?''). On two occasions, these storylines were explicitly labeled `good' and `bad' [C11, C15] (see Fig.~\ref{fig:storyboard_example}).

\subsubsection{Plot Elements}
\label{sec:narrative_plotelements}

This concept related to the aspects that make up the sequence of events in a story.

\paragraph{Background}

The educators came up with several different ideas for the narrative the robot would form a part of. 
They wanted realistic, animal-centered scenarios that could conceivably happen to a pet, such as encountering new people or going missing on a night with fireworks. They considered scenarios that linked to their organization, such as an animal's journey from rescue to rehoming.
The children's storyboards covered similar scenarios to the educators, with three narratives revolving around the rehoming or abandonment of a pet.

A couple of the educators' ideas involved seeing changes over time. One such idea revolved around short-term changes; the robot's personality would change over the course of a session, from shy and easily startled to more adventurous and playful if the class treat it correctly. Another idea focused on the long-term changes experienced in an animal's lifetime, for example the differences in behavior and needs between a puppy, an adult, and an elderly dog.

\paragraph{Conflict}

There were some suggestions from educators for narratives with negative events, like a dog going missing or getting ill, but some cautioned against being too negative as they considered this could dampen enthusiasm and lose some children. However, children themselves did not avoid negative scenarios, and some stories of theirs revolved around people's potential reactions to the robotic animal biting a person or destroying property.
As with educators, the children included the distressing effect of loud noises, like fireworks or shouting, as events that caused the robot to become scared.

\paragraph{Characters}

The main human character in the children's storyboards was generally, but not always, the owner. Children also included other animals in the interactions. Some of the educators' ideas positioned the children as the pet owner, but in other ideas the children were meeting a robot owned by someone else.

\paragraph{Setting}

The children's storyboards suggested they mostly envisioned interactions with a robot pet within the home (eight had drawings showing interiors and furniture).
In contrast, educators described interactions within a classroom environment, explained by the animal visiting the school with the educator.

\subsubsection{Actions}

This concept related to the actions a person takes during interaction with the robot, both positive and negative.

\paragraph{Positive}

Children particularly focused on the positive ways that people could interact with the robot animal. These included giving it food, stroking it or engaging in gentle touch, playing with it, and simply spending time with it. This last interaction was seen in four storyboards where the robot sat next to the child on a sofa or bed (e.g., ``let him sit on the couch with them'' led to ``he would be happy'' [C3]). In five storyboards, children presented only positive options, like patting the robot or taking it outside [C18], or playing with it or letting it sleep [C9].

\paragraph{Negative}
In contrast to the children, the educators focused mainly on negative interactions and teaching children what not to do. Negative actions included being loud, moving suddenly or crowding the robot, and rough handling.
Children also included negative actions; nine storyboards featured some form of negative action, including being loud, ignoring or neglecting the robot, and handling the robot roughly.

\section{Discussion}

As AWE presents a new use case for zoomorphic robots, it is important to understand the perspective of AW experts when approaching their design, and this work marks the first instance of involving AW educators in the design of a zoomorphic robot. This is also the first work to get children's input on the design of a zoomorphic robot. On top of this, the paralleled structure of the workshops and activities allowed us to identify similarities and differences between the two groups. Our findings reflect some aspects from \citet{collins2023skinzoomorphic} about a zoomorphic robot for adults with depression, like a preference for a familiar mammalian appearance, a soft, natural-colored covering, natural noises, and touch sensors for calming tactile interactions, suggesting a zoomorphic robot developed for AWE could have other uses. However, the perspective presented by the AW educators emphasizes the importance of other aspects, like signaling features and aversive reactions. The findings in this work have implications for the design of zoomorphic robots in general and in particular those positioned as household pets or those for educating children about animals.

\subsection{Robot Design Considerations}

In this section we outline some important considerations when designing a zoomorphic robot.

\subsubsection{Natural-Colored Fur}

The majority of both educator and child designs included fur, mostly in orange or brown colors (Section~\ref{sec:robotdesign_appearance}). Fur was also noted by the children as a common aspect of real pets, so including fur in a zoomorphic robot for AWE may help strengthen the connection to real pets. It was also associated with pleasing qualities, like softness and cuddliness. Fur also links to the children's focus on tactile interactions in their robot designs and narratives. Incorporating soft fur may thereby encourage gentle tactile interactions.
The combination of a natural, furry appearance and a large repertoire of animal behaviors is rarely seen in existing zoomorphic robots; neither zoomorphic robot demonstrated in these workshops had both.
Zoomorphic robots seem to fall into two groups: highly responsive with extensive actuation but distinctly robotic in appearance (e.g., MiRo, AIBO) versus natural in appearance but with limited sensing and/or actuation (e.g., Qoobo, Hasbro's Joy for All cat, Paro), with surface properties a key divide (plastic or metal vs fur).
We recognize the challenge of sensing and actuating through fur, but our findings stress its importance for linking zoomorphic robots to pets, just as \citet{loffler2020uncanny} found surface properties a strong predictor of animal-likeness.

\subsubsection{Signaling Features}
As shown in Fig.~\ref{fig:signaling_features}, the educators focused on how the zoomorphic robot could include and use signaling features, due to their importance in communicating internal state. 
Previous work suggests recognizable emotions are possible with zoomorphic robots. For example, videos of MiRo producing 11 affective expressions (including happy, sad, scared, and angry) were recognized significantly above the chance level~\cite{ghafurian2022zoomorphic}. Other work has used characteristics like ear stiffness, breathing rate and depth, and purring to successfully convey the arousal of a zoomorphic robot~\cite{yohanan2011designaffect}. Displays of affect can also change perceptions of zoomorphic robots; children who interacted with a version of MiRo that expressed arousal and valence through its ears, eyes, tail, and voice significantly increased in their beliefs about its emotive, cognitive, and perceptive abilities compared to a control~\cite{voysey2022influence}.
Therefore, a focus on the zoomorphic robot's signaling features could prove highly beneficial for AWE.

\subsubsection{Aversive Reactions}

While the children focused mainly on positive interactions with pets, especially how tactile interaction was pleasant, both for them (Section~\ref{sec:pet_attributes}) and the zoomorphic robot (Section~\ref{sec:robotdesign_behavior}), the educators centered negative situations where the pet is frightened or stressed---situations that are dangerous and need to be addressed to prevent harm to pets and children. Four of the five emotions educators teach about are negative (Section~\ref{sec:bkg_awe} and Section~\ref{sec:robotdesign_behavior}), and children's belief in animals' ability to experience these emotions is an important predictor of their attitudes towards animals~ \cite{hawkins2016children_bam}. However, in the study by \citet{ghafurian2022zoomorphic}, negative expressions were more likely to be miscategorized, with fearful and angry only successfully recognized by 32\% and 28\% of participants respectively. Similarly, participants in the study by \citet{yohanan2011designaffect} failed to identify low valence. This could be due to the expression design and actuation limitations, but may also reflect a wider difficulty identifying negative affect in pets \cite{wan2012dogfear, demirbas2016dogfear, dawson2019cataffect}.
The educators' focus on negative reactions may also be due to their experience with zoomorphic robots that only have positive reactions.
This is not to say a zoomorphic robot for AWE should forego positive interactions, but there is a key need for negative interactions that is unsatisfied by existing zoomorphic robots.

\subsubsection{Species-Specific Features}

The physical features included by most participants (children and educators) in their zoomorphic robot were eyes, ears, a mouth, a nose, and a tail (Fig.~\ref{fig:signaling_features}). These align with features of prototypical mammals selected by children~\cite{allen2015preschooltaxonomic, rusca1992developmentconceptsanimal}.
While several children mixed aspects of mammals, the educators' designs were usually a specific mammal. This may be because educators recognize that needs and behaviors are species-specific. For example, a tail thumping on the ground may signify excitement in dogs but irritation in cats~\cite{catsprotection2021behaviourguide}. Therefore, while a non-specific zoomorphic robot could be used to simulate a range of pets, its behaviors would need to be common to multiple species or modified for the specific species taught about.

\subsubsection{Role-Based Accessories}

A major difference between educators and children was the emphasis on responsibility versus fun.
This came across in the accessories included in the designs, with educators including items to fulfill pets' needs and children including items for play. While play is an important component of pet ownership and facilitates good animal well-being, AWE seeks to promote responsible pet care, as this is the area of pet ownership that children are less likely to engage in~\cite{muldoon2015ijustplayed}.
We suggest using pet care and play accessories to draw on children’s existing role in relation to pets (play) and to guide them to the desired role (responsible care).

\subsubsection{Cuteness}

A recurring theme in the children's workshop was cuteness. This came up in the feedback on the zoomorphic robots (Section~\ref{sec:demos}) and in the brainstorming about pets (Section~\ref{sec:pet_attributes}). 
Animals, particularly domestic mammals, are widely viewed as cute~\cite{archer2011preferences, little2012manipulation} and have been used as cute stimuli in studies~\cite{nittono2012power, sherman2009viewing}, but it is interesting to see cuteness applied to the zoomorphic robots, albeit to different extents.
In the brainstorming about pets, ``cute'' tended to coincide with physical attributes like fluffy, furry, soft, and small, which may explain why ``cute'' was applied to Qoobo more than MiRo ($N{=}16$ vs $N{=}2$).
Research on cuteness shows it captures and secures attention~\cite{brosch2007baby, kringelbach2008neural}, and influences preferences~\cite{golle2015preference}. Therefore, it may also have played a part in 14 of the children preferring Qoobo to MiRo.
Given the engaging appeal of cuteness, as well as the idea that harm to cute beings is morally proscribed~\cite{sherman2011cuteness}, it seems ideal for a zoomorphic robot for AWE to be cute.
However, many features associated with cuteness---big eyes, round flat faces, etc.~\cite{lorenz1971behaviour}---have negative health consequences in pets~\cite{packer2017purchasingpopular, packer2019truthdog}.
This raises a key question over if cuteness should be a design goal for zoomorphic robots, particularly when the focus is AW. Rejection of cuteness would go against the trend in many products~\cite{miesler2011cuteevolutionary, hinde1985evolution}, and a further question arises over a suitable design alternative. 
This finding merits in-depth interdisciplinary discussion with designers, psychologists, and AW experts.

\subsection{Narrative Design Considerations}

In this section we outline some important considerations when designing a narrative to accompany a zoomorphic robot.

\subsubsection{Conveying Changes}

Returning to the theme ``actions have consequences'', educators wanted to convey both the immediate changes as a result of interactions (e.g., change in mood as a result of environment and interactions) and the delayed or longer-term changes from pet care decisions (e.g., illness caused as a result of choosing not to vaccinate or skittishness from limited socialization as a young animal). Children's storyboards frequently picked up on the former, with their polarized storylines (Section~\ref{sec:narrative_topics}), but the latter may require repeated or longer-term interactions with the robot to convey these changes.

\subsubsection{Combining the Positive and Negative}

Despite the educators' concerns that negative narratives could alienate children (Section~\ref{sec:narrative_plotelements}), the children conceived both positive and negative events in their stories. Interestingly, despite the prompts asking for two actions without any other stipulations, children frequently presented polarized options in their storyboards, so it may be important to convey the nuances of choices in an educational intervention.
Also, while children conceived these negative actions for their stories, they may not want to act these out with the robot. \citet{voysey2022influence} found that children largely thought it was unacceptable to engage in acts of cruelty towards a zoomorphic robot (though groups of children have been known to abuse robots~\cite{brvsvcic2015escapingrobotabuse}). Thus the framing of the story may need to involve a character whose actions the children act out.

There is a balance to be struck between the positive and negative within an educational narrative---educators need the negative events to educate about dangerous interactions but are likely to want to end on positive note.
Therefore, it is worth considering common narrative structures and how to structure events in a suitable way, which may be challenging with some approaches to interactive narratives.
\citet{ligthart2020designpatterns} suggest some design patterns to keep the branching contained, including shallow branching, alternate endings, and world state choices, which may be relevant to employ here.

\subsubsection{Framing and Backstories}

The difference in narratives between children and educators raises challenges for the framing. Children often have strong personal relationships with their pets \cite[p.~27--28]{muldoon2009promotingdutyofcare} and these relationships exist in the context of their home, which we saw in the setting of their storyboards (Section~\ref{sec:narrative_plotelements}). However, in this use case, the robot will be used in a school, so we need to consider how to navigate this incongruity. This could be done by developing a robot backstory. Backstories can engender empathy towards a robot~\cite{darling2015empathicstories} and help children form an initial connection to it~\cite{lee2022unboxingbackstory}.
However, \citet{michaelis2023offscript} find that children can be disbelieving about certain backstory comments made by a robot, and suggest backstories should be designed to have clear ways of making sense of them and fit with the robot's capabilities.
For our purpose, the robot could be positioned as a new arrival from the rescue center that needs to get used to people before being rehomed or as the educator's own pet that they take into schools, much like Martha and Gibson, existing mascots for the Scottish SPCA~\cite{sspca2021marthagibson}.

Framing is also important for the interpretation of behaviors.
\citet{bucci2018furrynarrativeframe} found that if a participant's mental framing was positive, then a robot's shaking was perceived as excitement, whereas if it was negative, then the shaking was perceived as fear.
This can be advantageous---with the correct framing an ambiguous behavior can be explained.

\subsection{Lessons Learned}
\subsubsection{Reflections on Design Process}
Using a similar workshop structure and activities for educators and children worked well. In particular, it allowed us to appreciate children's contributions to the design.
While children do not have expert knowledge about features needed for an AWE intervention, they are the experts on their own experience with pets, and the data gave insights into their perceptions of pets (e.g., the importance of personal relationship, variety, and cuteness).

The workshop activities helped the children produce grounded, realistic designs that followed the brief. The designs did not simply reproduce the demonstrations, but did incorporate features from them, like different sensors, suggesting the demonstrations influenced them to include technical features in their designs. The prompts about pet attributes and behavior functioned similarly, as several children used an attribute they had brainstormed directly in their design, as in \citet{sanoubari2021remotecodesign}.

The draw-and-write approach allowed the children to customize the balance of drawing and writing, from one drawing with no writing to one with 24 labels. 
There were some labels that were difficult to decipher, which is to be expected when working with children this age, so it would be beneficial to look into methods like draw, write, and tell~\cite{angell2015draw} to give the children a chance to clarify and expand on aspects of their designs. Unfortunately, this method is very time intensive, which is why it was not used here.

The branching storyboards were a highly effective tool for gathering rich data from the children. Firstly, the level of scaffolding seemed to strike the happy medium described by \citet{moraveji2007comicboarding}, where there was enough guidance to avoid children feeling overwhelmed by options, but enough freedom to share a range of ideas.
Secondly, the storyboards provided insights into the extent of children's AW knowledge, potential actions they could take, and potential consequences to these actions.
Several storyboards gave correct interpretations for the behavior of common pets and the underlying reason for this behavior, like a dog-inspired robot cowering in fear due to being shouted at (Fig.~\ref{fig:storyboard_example}).
Finally, the new storyboards were designed to combine a traditional storyboard with the `Animals-At-Risk' thematic apperception test (AAR TAT). We found that the themes collated from the data reflected material the designers of the AAR TAT expected it to elicit, i.e., ``attachment, loss, separation, discipline, and conflicts over pet care''~\cite[p.~67]{shapiro2013anicare}. For example, three of the children's storyboards centered around reactions to the pet breaking belongings [C5, C15, C21], which is the scenario of a prompt in the AAR TAT.

\subsubsection{Achieving Consensus}

We found similarities across the educators and the children, but also differences (Fig.~\ref{fig:mindmap_robot}). A key challenge for PD with multiple user groups is achieving consensus and navigating conflicting design requirements. 
A common approach to achieve consensus involves bringing stakeholders together to discuss and exchange views, which requires careful mediation~\cite{coleman2014conflictresolution}. This may be especially tricky where there is a power imbalance between the groups (e.g., `experts' and users, or adults and children) as participants may feel a pressure to acquiesce~\cite{leung2019buildingconsensus}.
Alternatively, consensus can be achieved through sequential design workshops where one group builds on the ideas of the other, like \citet{newbutt2022codesigning}.
However, this step constrains the second group to the ideas already expressed by the first, and researchers may be less likely to pick up on differences, or indeed similarities, between the groups. 
While our methodology does not help us to reach consensus at this stage, the mirrored structure and activities of our workshops meant we got the perspectives and designs of both groups, uninfluenced by the other.
To extend this work and build a level of consensus, it would be helpful to combine the two groups for discussion.
Consensus will not necessarily mean all requirements are satisfied, but instead will mean the final prototype is satisfactory to all, given the existing constraints~\cite{briggs2005consensusbuilding}.

\subsubsection{Children's AW Knowledge}

This group of children demonstrated a solid understanding of animals and animal behavior. They recognized many of the same aspects as educators when it came to the importance of AWE (Table~\ref{tab:education_aims}) and the needs of pets. Furthermore, several storyboards gave correct interpretations for the behavior of common pets and the underlying reason for this behavior, like a dog-inspired robot cowering in fear due to being shouted at (Fig.~\ref{fig:storyboard_example}).
However, this level of AW knowledge may be specific to this group of children.
Other studies on AWE interventions in a similar region find that 63--67\% of children own a pet~\cite{williams2022rabbit, hawkins2020seriousgame, hawkins2017childhoodattachment}, close to the 71\% figure here, but in the sample from \citet{williams2022rabbit}, only 11\% of children owned more than one species of pet, compared to 42\% of children in our group.
A chi-square independence test showed the variety of pets owned by a child differed significantly between the two samples ($\chi^2(2, N{=}147){=}15.1$, $p{<}0.001$) meaning a greater proportion of children owned more than one species of pet in our sample.
Owning multiple species of pet could indicate greater interest in animals and thereby potentially greater knowledge than average in our sample.
Nevertheless, the designs revealed some gaps in children's understanding; the children's intuition of animal communication was limited to the noises they make, whereas educators included cues from the facial features and tail. Children are more likely to misinterpret dog behavior when they do not take a full range of cues (movement, posture, sound, face, and tail) into account~\cite{lakestani2014interpretationdog}, which can increase their risk of being bitten~\cite{overall2001dogbites}.
Our findings reinforce the work of \citet{muldoon2016perspectiveswelfare} that children are uncertain about identifying behavioral cues and interactive education methods would be helpful.

\subsection{Limitations}

These findings are specific to the country and context (e.g., high income country, high levels of pet ownership, high expressed regard for AW), so it would be good to involve participants from other regions.
Furthermore, the children involved were all from one school. Prior research has shown differences in attitudes to animals between adolescents in urban and rural environments \cite{hawkins2020children, bjerke1998attitudesnorwegian} (although other studies do not find differences~\cite{marsa2016short}), so it could be beneficial to involve a group from another school in a different area. 
However, involving another group of participants is unlikely to remove recurrent themes.

The findings for the narrative design provide only an initial flavor of the setting and scenarios for the educational intervention, but defining the structure and events of the story would need further work. A possible extension is running a writing workshop with the two groups to further develop different story ideas when a prototype exists.
Future work with the branching storyboard tool could compare it with a plain storyboard by looking at the difference in the number of ideas generated, as in \citet{moraveji2007comicboarding}.

\subsection{Future Work}

These design insights will be used to develop a prototype zoomorphic robot for AWE, focusing on how it can facilitate the learning objectives identified in Table~\ref{tab:education_aims}. 
Incorporating the following aspects should help the system achieve the goal of promoting kind, caring behavior to pets:

\begin{itemize}
    \item Species-specific realistic aversive reactions to discourage unwanted behaviors
    \item Facial features, ears, and tail to convey positive and negative emotions
    \item Natural-colored fur to promote the link to real pets and encourage gentle tactile interaction
    \item Pet care and play accessories to draw on children’s existing role in relation to pets (play) and to guide them to the desired role (responsible care)
    \item Narrative developments to highlight the consequences of actions
\end{itemize}

The zoomorphic robot will be developed in tandem with an appropriate interactive narrative and other educational materials. Once developed and refined based on feedback from AW educators and children, the zoomorphic robot will be tested in an in-school AWE program to see if it results in improved learning outcomes compared to existing tools.

\section{Conclusion}

We conducted two PD workshops, one with 11 educators from an AW organization and one with 24 primary school children, to inform the design of a zoomorphic robot and narrative for AWE. 
Their discussions and designs have highlighted important considerations for the appearance, behavior, and features of such a robot, including furriness, negative reactions to undesirable behavior, and facial features and a tail that can signal an animal's internal state.
When it comes to the narrative around a zoomorphic robot for AWE, it is important to consider how to convey changes (``actions have consequences''), balance positive and negative experiences, and frame interactions with the zoomorphic robot.
Through conducting separate PD workshops with educators and children, but mirroring their structure and content, we found key similarities to build on (e.g., pets' use of noises to communicate emotions, a focus on tactile interactions) and key differences to navigate (e.g., the focus on responsibility or fun in pet ownership).
In all, this work marks an important step in responding to a call from AWE researchers for more interactive teaching tools~\cite{muldoon2016perspectiveswelfare,williams2022rabbit}.

\backmatter

\bmhead{Acknowledgements}

Many thanks to the Education Team at the Scottish SPCA and the children and teachers who participated in this work. Thanks also to those who provided feedback on the activities and the members of the Interactive and Trustworthy Technologies research group who were facilitators.


\section*{Declarations}


\bmhead{Funding}

This work was supported by the United Kingdom Research and Innovation (grant EP/S023208/1), Engineering and Physical Sciences Research Council Centre for Doctoral Training in Robotics and Autonomous Systems.

\bmhead{Ethics approval and consent to participate}

The research protocol received ethical approval from the University of Edinburgh School of Informatics. Educators, children, and guardians received information sheets before the study and signed physical consent forms.

\bmhead{Data availability}

The materials used in the workshops and data generated by this research are available online via the Open Science Framework: \url{https://osf.io/zd89e/?view_only=fd944f3a0f3a4da08ee31e8ae6749f66}.



\bibliography{ijsr_pd_refs}


\begin{thebibliography}{109}
\ifx \bisbn   \undefined \def \bisbn  #1{ISBN #1}\fi
\ifx \binits  \undefined \def \binits#1{#1}\fi
\ifx \bauthor  \undefined \def \bauthor#1{#1}\fi
\ifx \batitle  \undefined \def \batitle#1{#1}\fi
\ifx \bjtitle  \undefined \def \bjtitle#1{#1}\fi
\ifx \bvolume  \undefined \def \bvolume#1{\textbf{#1}}\fi
\ifx \byear  \undefined \def \byear#1{#1}\fi
\ifx \bissue  \undefined \def \bissue#1{#1}\fi
\ifx \bfpage  \undefined \def \bfpage#1{#1}\fi
\ifx \blpage  \undefined \def \blpage #1{#1}\fi
\ifx \burl  \undefined \def \burl#1{\textsf{#1}}\fi
\ifx \doiurl  \undefined \def \doiurl#1{\url{https://doi.org/#1}}\fi
\ifx \betal  \undefined \def \betal{\textit{et al.}}\fi
\ifx \binstitute  \undefined \def \binstitute#1{#1}\fi
\ifx \binstitutionaled  \undefined \def \binstitutionaled#1{#1}\fi
\ifx \bctitle  \undefined \def \bctitle#1{#1}\fi
\ifx \beditor  \undefined \def \beditor#1{#1}\fi
\ifx \bpublisher  \undefined \def \bpublisher#1{#1}\fi
\ifx \bbtitle  \undefined \def \bbtitle#1{#1}\fi
\ifx \bedition  \undefined \def \bedition#1{#1}\fi
\ifx \bseriesno  \undefined \def \bseriesno#1{#1}\fi
\ifx \blocation  \undefined \def \blocation#1{#1}\fi
\ifx \bsertitle  \undefined \def \bsertitle#1{#1}\fi
\ifx \bsnm \undefined \def \bsnm#1{#1}\fi
\ifx \bsuffix \undefined \def \bsuffix#1{#1}\fi
\ifx \bparticle \undefined \def \bparticle#1{#1}\fi
\ifx \barticle \undefined \def \barticle#1{#1}\fi
\bibcommenthead
\ifx \bconfdate \undefined \def \bconfdate #1{#1}\fi
\ifx \botherref \undefined \def \botherref #1{#1}\fi
\ifx \url \undefined \def \url#1{\textsf{#1}}\fi
\ifx \bchapter \undefined \def \bchapter#1{#1}\fi
\ifx \bbook \undefined \def \bbook#1{#1}\fi
\ifx \bcomment \undefined \def \bcomment#1{#1}\fi
\ifx \oauthor \undefined \def \oauthor#1{#1}\fi
\ifx \citeauthoryear \undefined \def \citeauthoryear#1{#1}\fi
\ifx \endbibitem  \undefined \def \endbibitem {}\fi
\ifx \bconflocation  \undefined \def \bconflocation#1{#1}\fi
\ifx \arxivurl  \undefined \def \arxivurl#1{\textsf{#1}}\fi
\csname PreBibitemsHook\endcsname

\bibitem[\protect\citeauthoryear{Marsa-Sambola et~al.}{2016a}]{marsa2016sociodemographics}
\begin{barticle}
\bauthor{\bsnm{Marsa-Sambola}, \binits{F.}},
\bauthor{\bsnm{Williams}, \binits{J.}},
\bauthor{\bsnm{Muldoon}, \binits{J.}},
\bauthor{\bsnm{Lawrence}, \binits{A.}},
\bauthor{\bsnm{Connor}, \binits{M.}},
\bauthor{\bsnm{Roberts}, \binits{C.}},
\bauthor{\bsnm{Brooks}, \binits{F.}},
\bauthor{\bsnm{Currie}, \binits{C.}}:
\batitle{{Sociodemographics of pet ownership among adolescents in Great Britain: Findings from the HBSC Study in England, Scotland, and Wales}}.
\bjtitle{Anthrozo{\"o}s}
\bvolume{29}(\bissue{4}),
\bfpage{559}--\blpage{580}
(\byear{2016})
\doiurl{10.1080/08927936.2016.1228756}
\end{barticle}
\endbibitem

\bibitem[\protect\citeauthoryear{Marsa-Sambola et~al.}{2016b}]{marsa2016short}
\begin{barticle}
\bauthor{\bsnm{Marsa-Sambola}, \binits{F.}},
\bauthor{\bsnm{Muldoon}, \binits{J.}},
\bauthor{\bsnm{Williams}, \binits{J.}},
\bauthor{\bsnm{Lawrence}, \binits{A.}},
\bauthor{\bsnm{Connor}, \binits{M.}},
\bauthor{\bsnm{Currie}, \binits{C.}}:
\batitle{The short attachment to pets scale ({SAPS}) for children and young people: Development, psychometric qualities and demographic and health associations}.
\bjtitle{Child Indicators Research}
\bvolume{9}(\bissue{1}),
\bfpage{111}--\blpage{131}
(\byear{2016})
\doiurl{10.1007/s12187-015-9303-9}
\end{barticle}
\endbibitem

\bibitem[\protect\citeauthoryear{Purewal et~al.}{2017}]{purewal2017companionanimals}
\begin{barticle}
\bauthor{\bsnm{Purewal}, \binits{R.}},
\bauthor{\bsnm{Christley}, \binits{R.}},
\bauthor{\bsnm{Kordas}, \binits{K.}},
\bauthor{\bsnm{Joinson}, \binits{C.}},
\bauthor{\bsnm{Meints}, \binits{K.}},
\bauthor{\bsnm{Gee}, \binits{N.}},
\bauthor{\bsnm{Westgarth}, \binits{C.}}:
\batitle{Companion animals and child/adolescent development: A systematic review of the evidence}.
\bjtitle{International journal of environmental research and public health}
\bvolume{14}(\bissue{3}),
\bfpage{234}
(\byear{2017})
\doiurl{10.3390/ijerph14030234}
\end{barticle}
\endbibitem

\bibitem[\protect\citeauthoryear{Ladny and Meyer}{2020}]{ladny2020traumatizedwitnesses}
\begin{barticle}
\bauthor{\bsnm{Ladny}, \binits{R.T.}},
\bauthor{\bsnm{Meyer}, \binits{L.}}:
\batitle{Traumatized witnesses: Review of childhood exposure to animal cruelty}.
\bjtitle{Journal of child \& adolescent trauma}
\bvolume{13},
\bfpage{527}--\blpage{537}
(\byear{2020})
\doiurl{10.1007/s40653-019-00277-x}
\end{barticle}
\endbibitem

\bibitem[\protect\citeauthoryear{Ferreira and Williams}{2023}]{ferreira2023understanding}
\begin{bbook}
\bauthor{\bsnm{Ferreira}, \binits{G.M.}},
\bauthor{\bsnm{Williams}, \binits{J.M.}}:
\bbtitle{Understanding Animal Abuse and How to Intervene with Children and Young People: A Practical Guide for Professionals Working with People and Animals}.
\bpublisher{Taylor \& Francis},
\blocation{London}
(\byear{2023}).
\doiurl{10.4324/9781003165552}
\end{bbook}
\endbibitem

\bibitem[\protect\citeauthoryear{Wauthier and Williams}{2022}]{wauthier2022understanding}
\begin{barticle}
\bauthor{\bsnm{Wauthier}, \binits{L.M.}},
\bauthor{\bsnm{Williams}, \binits{J.M.}}:
\batitle{Understanding and conceptualizing childhood animal harm: A meta-narrative systematic review}.
\bjtitle{Anthrozo{\"o}s}
\bvolume{35}(\bissue{2}),
\bfpage{165}--\blpage{202}
(\byear{2022})
\doiurl{10.1080/08927936.2021.1986262}
\end{barticle}
\endbibitem

\bibitem[\protect\citeauthoryear{{Scottish SPCA}}{2019}]{sspca2019about}
\begin{botherref}
\oauthor{\bsnm{{Scottish SPCA}}}:
About the Scottish SPCA
(2019).
\url{https://www.scottishspca.org/about-the-scottish-spca}
\end{botherref}
\endbibitem

\bibitem[\protect\citeauthoryear{Glenk}{2017}]{glenk2017therapydog}
\begin{barticle}
\bauthor{\bsnm{Glenk}, \binits{L.M.}}:
\batitle{Current perspectives on therapy dog welfare in animal-assisted interventions}.
\bjtitle{Animals}
\bvolume{7}(\bissue{2}),
\bfpage{7}
(\byear{2017})
\doiurl{10.3390/ani7020007}
\end{barticle}
\endbibitem

\bibitem[\protect\citeauthoryear{Muldoon and Williams}{2021a}]{muldoon2021delphi1establishing}
\begin{barticle}
\bauthor{\bsnm{Muldoon}, \binits{J.C.}},
\bauthor{\bsnm{Williams}, \binits{J.M.}}:
\batitle{Establishing consensus on the best ways to educate children about animal welfare and prevent harm: An online delphi study}.
\bjtitle{Animal Welfare}
\bvolume{30}(\bissue{2}),
\bfpage{179}--\blpage{195}
(\byear{2021})
\doiurl{10.7120/09627286.30.2.179}
\end{barticle}
\endbibitem

\bibitem[\protect\citeauthoryear{Muldoon and Williams}{2021b}]{muldoon2021delphi2challenges}
\begin{barticle}
\bauthor{\bsnm{Muldoon}, \binits{J.C.}},
\bauthor{\bsnm{Williams}, \binits{J.M.}}:
\batitle{The challenges and future development of animal welfare education in the uk}.
\bjtitle{Animal Welfare}
\bvolume{30}(\bissue{2}),
\bfpage{197}--\blpage{209}
(\byear{2021})
\doiurl{10.7120/09627286.30.2.197}
\end{barticle}
\endbibitem

\bibitem[\protect\citeauthoryear{Belpaeme et~al.}{2018}]{belpaeme2018education}
\begin{barticle}
\bauthor{\bsnm{Belpaeme}, \binits{T.}},
\bauthor{\bsnm{Kennedy}, \binits{J.}},
\bauthor{\bsnm{Ramachandran}, \binits{A.}},
\bauthor{\bsnm{Scassellati}, \binits{B.}},
\bauthor{\bsnm{Tanaka}, \binits{F.}}:
\batitle{Social robots for education: A review}.
\bjtitle{Science robotics}
\bvolume{3}(\bissue{21}),
\bfpage{5954}
(\byear{2018})
\doiurl{10.1126/scirobotics.aat5954}
\end{barticle}
\endbibitem

\bibitem[\protect\citeauthoryear{Riedl and Bulitko}{2013}]{riedl2013interactive}
\begin{barticle}
\bauthor{\bsnm{Riedl}, \binits{M.O.}},
\bauthor{\bsnm{Bulitko}, \binits{V.}}:
\batitle{Interactive narrative: An intelligent systems approach}.
\bjtitle{Ai Magazine}
\bvolume{34}(\bissue{1}),
\bfpage{67}--\blpage{67}
(\byear{2013})
\doiurl{10.1609/aimag.v34i1.2449}
\end{barticle}
\endbibitem

\bibitem[\protect\citeauthoryear{Steel et~al.}{2022}]{steel2022teacherreadingdog}
\begin{barticle}
\bauthor{\bsnm{Steel}, \binits{J.}},
\bauthor{\bsnm{Williams}, \binits{J.M.}},
\bauthor{\bsnm{McGeown}, \binits{S.}}:
\batitle{Teacher-researcher collaboration in animal-assisted education: Co-designing a reading to dogs intervention}.
\bjtitle{Educational Research}
\bvolume{64}(\bissue{1}),
\bfpage{113}--\blpage{131}
(\byear{2022})
\doiurl{10.1080/00131881.2021.2016061}
\end{barticle}
\endbibitem

\bibitem[\protect\citeauthoryear{Elloumi et~al.}{2022}]{elloumi2022exploring}
\begin{bchapter}
\bauthor{\bsnm{Elloumi}, \binits{L.}},
\bauthor{\bsnm{Bossema}, \binits{M.}},
\bauthor{\bsnm{De~Droog}, \binits{S.M.}},
\bauthor{\bsnm{Smakman}, \binits{M.H.}},
\bauthor{\bsnm{Van~Ginkel}, \binits{S.}},
\bauthor{\bsnm{Ligthart}, \binits{M.E.}},
\bauthor{\bsnm{Hoogland}, \binits{K.}},
\bauthor{\bsnm{Hindriks}, \binits{K.V.}},
\bauthor{\bsnm{Allouch}, \binits{S.B.}}:
\bctitle{Exploring requirements and opportunities for social robots in primary mathematics education}.
In: \bbtitle{2022 31st IEEE International Conference on Robot and Human Interactive Communication (RO-MAN)},
pp. \bfpage{316}--\blpage{322}
(\byear{2022}).
\doiurl{10.1109/ro-man53752.2022.9900569} .
\bcomment{IEEE}
\end{bchapter}
\endbibitem

\bibitem[\protect\citeauthoryear{Morgan et~al.}{2002}]{morgan2002focus}
\begin{barticle}
\bauthor{\bsnm{Morgan}, \binits{M.}},
\bauthor{\bsnm{Gibbs}, \binits{S.}},
\bauthor{\bsnm{Maxwell}, \binits{K.}},
\bauthor{\bsnm{Britten}, \binits{N.}}:
\batitle{Hearing children's voices: methodological issues in conducting focus groups with children aged 7-11 years}.
\bjtitle{Qualitative research}
\bvolume{2}(\bissue{1}),
\bfpage{5}--\blpage{20}
(\byear{2002})
\doiurl{10.1177/1468794102002001636}
\end{barticle}
\endbibitem

\bibitem[\protect\citeauthoryear{Alves-Oliveira et~al.}{2021}]{alves2021children}
\begin{bchapter}
\bauthor{\bsnm{Alves-Oliveira}, \binits{P.}},
\bauthor{\bsnm{Arriaga}, \binits{P.}},
\bauthor{\bsnm{Paiva}, \binits{A.}},
\bauthor{\bsnm{Hoffman}, \binits{G.}}:
\bctitle{Children as robot designers}.
In: \bbtitle{Proceedings of the 2021 ACM/IEEE International Conference on Human-Robot Interaction},
pp. \bfpage{399}--\blpage{408}
(\byear{2021}).
\doiurl{10.1145/3434073.3444650}
\end{bchapter}
\endbibitem

\bibitem[\protect\citeauthoryear{Collins et~al.}{2023}]{collins2023skinzoomorphic}
\begin{bchapter}
\bauthor{\bsnm{Collins}, \binits{S.}},
\bauthor{\bsnm{Hicks}, \binits{D.}},
\bauthor{\bsnm{Henkel}, \binits{Z.}},
\bauthor{\bsnm{Baugus~Henkel}, \binits{K.}},
\bauthor{\bsnm{Piatt}, \binits{J.A.}},
\bauthor{\bsnm{Bethel}, \binits{C.L.}},
\bauthor{\bsnm{Sabanovic}, \binits{S.}}:
\bctitle{What skin is your robot in?}
In: \bbtitle{Companion of the 2023 ACM/IEEE International Conference on Human-Robot Interaction},
pp. \bfpage{511}--\blpage{515}
(\byear{2023}).
\doiurl{10.1145/3568294.3580137}
\end{bchapter}
\endbibitem

\bibitem[\protect\citeauthoryear{Bradwell et~al.}{2019}]{bradwell2019companion}
\begin{barticle}
\bauthor{\bsnm{Bradwell}, \binits{H.L.}},
\bauthor{\bsnm{Edwards}, \binits{K.J.}},
\bauthor{\bsnm{Winnington}, \binits{R.}},
\bauthor{\bsnm{Thill}, \binits{S.}},
\bauthor{\bsnm{Jones}, \binits{R.B.}}:
\batitle{Companion robots for older people: importance of user-centred design demonstrated through observations and focus groups comparing preferences of older people and roboticists in south west england}.
\bjtitle{BMJ open}
\bvolume{9}(\bissue{9}),
\bfpage{032468}
(\byear{2019})
\doiurl{10.1136/bmjopen-2019-032468}
\end{barticle}
\endbibitem

\bibitem[\protect\citeauthoryear{Zhang et~al.}{2022}]{zhang2022paincodesign}
\begin{bchapter}
\bauthor{\bsnm{Zhang}, \binits{F.}},
\bauthor{\bsnm{Broz}, \binits{F.}},
\bauthor{\bsnm{Dertien}, \binits{E.}},
\bauthor{\bsnm{Kousi}, \binits{N.}},
\bauthor{\bsnm{Van~Gurp}, \binits{J.A.}},
\bauthor{\bsnm{Ferrari}, \binits{O.I.}},
\bauthor{\bsnm{Malagon}, \binits{I.}},
\bauthor{\bsnm{Barakova}, \binits{E.I.}}:
\bctitle{Understanding design preferences for robots for pain management: A co-design study}.
In: \bbtitle{2022 17th ACM/IEEE International Conference on Human-Robot Interaction (HRI)},
pp. \bfpage{1124}--\blpage{1129}
(\byear{2022}).
\doiurl{10.1109/hri53351.2022.9889542} .
\bcomment{IEEE}
\end{bchapter}
\endbibitem

\bibitem[\protect\citeauthoryear{Lohse et~al.}{2007}]{lohse2007appearance}
\begin{bchapter}
\bauthor{\bsnm{Lohse}, \binits{M.}},
\bauthor{\bsnm{Hegel}, \binits{F.}},
\bauthor{\bsnm{Swadzba}, \binits{A.}},
\bauthor{\bsnm{Rohlfing}, \binits{K.}},
\bauthor{\bsnm{Wachsmuth}, \binits{S.}},
\bauthor{\bsnm{Wrede}, \binits{B.}}:
\bctitle{What can i do for you? appearance and application of robots}.
In: \bbtitle{Proceedings of AISB},
vol. \bseriesno{7},
pp. \bfpage{121}--\blpage{126}
(\byear{2007}).
\doiurl{10.5040/9781784606770.00000006}
\end{bchapter}
\endbibitem

\bibitem[\protect\citeauthoryear{Marchetti et~al.}{2022}]{marchetti2022petfloorwashing}
\begin{bchapter}
\bauthor{\bsnm{Marchetti}, \binits{E.}},
\bauthor{\bsnm{Grimme}, \binits{S.}},
\bauthor{\bsnm{Hornecker}, \binits{E.}},
\bauthor{\bsnm{Kollakidou}, \binits{A.}},
\bauthor{\bsnm{Graf}, \binits{P.}}:
\bctitle{Pet-robot or appliance? care home residents with dementia respond to a zoomorphic floor washing robot}.
In: \bbtitle{Proceedings of the 2022 CHI Conference on Human Factors in Computing Systems},
pp. \bfpage{1}--\blpage{21}
(\byear{2022}).
\doiurl{10.1145/3491102.3517463}
\end{bchapter}
\endbibitem

\bibitem[\protect\citeauthoryear{Newbutt et~al.}{2022}]{newbutt2022codesigning}
\begin{barticle}
\bauthor{\bsnm{Newbutt}, \binits{N.}},
\bauthor{\bsnm{Rice}, \binits{L.}},
\bauthor{\bsnm{Lemaignan}, \binits{S.}},
\bauthor{\bsnm{Daly}, \binits{J.}},
\bauthor{\bsnm{Charisi}, \binits{V.}},
\bauthor{\bsnm{Conley}, \binits{I.}}:
\batitle{Co-designing a social robot in a special educational needs school: Listening to the ambitions of autistic children and their teachers}.
\bjtitle{Interaction Studies}
\bvolume{23}(\bissue{2}),
\bfpage{204}--\blpage{242}
(\byear{2022})
\doiurl{10.1075/is.21031.new}
\end{barticle}
\endbibitem

\bibitem[\protect\citeauthoryear{{Scottish SPCA}}{2018}]{sspca2018robowunderkind}
\begin{botherref}
\oauthor{\bsnm{{Scottish SPCA}}}:
Scottish SPCA Launches New Partnership with Robo Wunderkind
(2018).
\url{https://www.scottishspca.org/news/scottish-spca-launches-new-partnership-with-robo-wunderkind}
\end{botherref}
\endbibitem

\bibitem[\protect\citeauthoryear{Williams and Ferreira}{2023}]{williams2023animal}
\begin{bchapter}
\bauthor{\bsnm{Williams}, \binits{J.M.}},
\bauthor{\bsnm{Ferreira}, \binits{G.M.}}:
\bctitle{Animal abuse: What we know, what we can do and what we need to know}.
In: \bbtitle{Understanding Animal Abuse and How to Intervene with Children and Young People},
pp. \bfpage{167}--\blpage{182}.
\bpublisher{Taylor \& Francis},
\blocation{London}
(\byear{2023}).
\doiurl{10.4324/9781003165552-10}
\end{bchapter}
\endbibitem

\bibitem[\protect\citeauthoryear{Muldoon and Williams}{2021}]{muldoon2021awetoolkit}
\begin{botherref}
\oauthor{\bsnm{Muldoon}, \binits{J.}},
\oauthor{\bsnm{Williams}, \binits{J.}}:
The Animal Welfare Education for Children Toolkit.
Children, adolescents and animals research (Caar), University of Edinburgh,
(2021).
\doiurl{10.4324/9781003165552-9} .
Children, adolescents and animals research (Caar), University of Edinburgh
\end{botherref}
\endbibitem

\bibitem[\protect\citeauthoryear{Williams et~al.}{2022}]{williams2022rabbit}
\begin{barticle}
\bauthor{\bsnm{Williams}, \binits{J.M.}},
\bauthor{\bsnm{Cardoso}, \binits{M.P.}},
\bauthor{\bsnm{Zumaglini}, \binits{S.}},
\bauthor{\bsnm{Finney}, \binits{A.L.}},
\bauthor{\bsnm{{Scottish SPCA}}},
\bauthor{\bsnm{Knoll}, \binits{M.A.}}:
\batitle{“rabbit rescuers”: A school-based animal welfare education intervention for young children}.
\bjtitle{Anthrozo{\"o}s}
\bvolume{35}(\bissue{1}),
\bfpage{55}--\blpage{73}
(\byear{2022})
\doiurl{10.1080/08927936.2021.1944561}
\end{barticle}
\endbibitem

\bibitem[\protect\citeauthoryear{Hawkins et~al.}{2020}]{hawkins2020seriousgame}
\begin{barticle}
\bauthor{\bsnm{Hawkins}, \binits{R.D.}},
\bauthor{\bsnm{{Scottish SPCA}}},
\bauthor{\bsnm{Williams}, \binits{J.M.}}:
\batitle{The development and pilot evaluation of a ‘serious game’to promote positive child-animal interactions}.
\bjtitle{Human-Animal Interaction Bulletin}
(\bissue{2022})
(\byear{2020})
\doiurl{10.1079/hai.2020.0008}
\end{barticle}
\endbibitem

\bibitem[\protect\citeauthoryear{Wauthier et~al.}{2023}]{wauthier2023preliminary}
\begin{barticle}
\bauthor{\bsnm{Wauthier}, \binits{L.M.}},
\bauthor{\bsnm{Farnfield}, \binits{S.}},
\bauthor{\bsnm{{Scottish SPCA}}},
\bauthor{\bsnm{Williams}, \binits{J.M.}}:
\batitle{A preliminary exploration of the psychological risk factors for childhood animal cruelty: The roles of attachment, self-regulation, and empathy}.
\bjtitle{Anthrozo{\"o}s}
\bvolume{36}(\bissue{3}),
\bfpage{447}--\blpage{469}
(\byear{2023})
\doiurl{10.1080/08927936.2022.2125197}
\end{barticle}
\endbibitem

\bibitem[\protect\citeauthoryear{Miller and Nourbakhsh}{2016}]{miller2016education}
\begin{bchapter}
\bauthor{\bsnm{Miller}, \binits{D.P.}},
\bauthor{\bsnm{Nourbakhsh}, \binits{I.}}:
\bctitle{Robotics for education}.
In: \bbtitle{Springer Handbook of Robotics},
pp. \bfpage{2115}--\blpage{2134}.
\bpublisher{Springer},
\blocation{Switzerland}
(\byear{2016}).
\doiurl{10.1007/978-3-319-32552-1_79}
\end{bchapter}
\endbibitem

\bibitem[\protect\citeauthoryear{D'Amico et~al.}{2020}]{damico2020educational}
\begin{botherref}
\oauthor{\bsnm{D'Amico}, \binits{A.}},
\oauthor{\bsnm{Guastella}, \binits{D.}},
\oauthor{\bsnm{Chella}, \binits{A.}}:
A playful experiential learning system with educational robotics.
Frontiers in Robotics and AI,
33
(2020)
\doiurl{10.3389/frobt.2020.00033}
\end{botherref}
\endbibitem

\bibitem[\protect\citeauthoryear{Saerbeck et~al.}{2010}]{saerbeck2010expressivetutor}
\begin{bchapter}
\bauthor{\bsnm{Saerbeck}, \binits{M.}},
\bauthor{\bsnm{Schut}, \binits{T.}},
\bauthor{\bsnm{Bartneck}, \binits{C.}},
\bauthor{\bsnm{Janse}, \binits{M.D.}}:
\bctitle{Expressive robots in education: varying the degree of social supportive behavior of a robotic tutor}.
In: \bbtitle{Proceedings of the SIGCHI Conference on Human Factors in Computing Systems},
pp. \bfpage{1613}--\blpage{1622}
(\byear{2010}).
\doiurl{10.1145/1753326.1753567}
\end{bchapter}
\endbibitem

\bibitem[\protect\citeauthoryear{K{\"o}se et~al.}{2015}]{kose2015signlanguagetutor}
\begin{barticle}
\bauthor{\bsnm{K{\"o}se}, \binits{H.}},
\bauthor{\bsnm{Uluer}, \binits{P.}},
\bauthor{\bsnm{Akal{\i}n}, \binits{N.}},
\bauthor{\bsnm{Yorganc{\i}}, \binits{R.}},
\bauthor{\bsnm{{\"O}zkul}, \binits{A.}},
\bauthor{\bsnm{Ince}, \binits{G.}}:
\batitle{The effect of embodiment in sign language tutoring with assistive humanoid robots}.
\bjtitle{International Journal of Social Robotics}
\bvolume{7},
\bfpage{537}--\blpage{548}
(\byear{2015})
\doiurl{10.1007/s12369-015-0311-1}
\end{barticle}
\endbibitem

\bibitem[\protect\citeauthoryear{Ahmad et~al.}{2019}]{ahmad2019robotvocabulary}
\begin{barticle}
\bauthor{\bsnm{Ahmad}, \binits{M.I.}},
\bauthor{\bsnm{Mubin}, \binits{O.}},
\bauthor{\bsnm{Shahid}, \binits{S.}},
\bauthor{\bsnm{Orlando}, \binits{J.}}:
\batitle{Robot’s adaptive emotional feedback sustains children’s social engagement and promotes their vocabulary learning: a long-term child--robot interaction study}.
\bjtitle{Adaptive Behavior}
\bvolume{27}(\bissue{4}),
\bfpage{243}--\blpage{266}
(\byear{2019})
\doiurl{10.1177/1059712319844182}
\end{barticle}
\endbibitem

\bibitem[\protect\citeauthoryear{Fasola and Matari{\'c}}{2013}]{fasola2013exercisecoachembodiment}
\begin{barticle}
\bauthor{\bsnm{Fasola}, \binits{J.}},
\bauthor{\bsnm{Matari{\'c}}, \binits{M.J.}}:
\batitle{A socially assistive robot exercise coach for the elderly}.
\bjtitle{Journal of Human-Robot Interaction}
\bvolume{2}(\bissue{2}),
\bfpage{3}--\blpage{32}
(\byear{2013})
\doiurl{10.5898/jhri.2.2.fasola}
\end{barticle}
\endbibitem

\bibitem[\protect\citeauthoryear{Weinberg et~al.}{2003}]{weinberg2003roboticsineducation}
\begin{barticle}
\bauthor{\bsnm{Weinberg}, \binits{J.B.}},
\bauthor{\bsnm{Yu}, \binits{X.}}, \betal:
\batitle{Robotics in education: Low-cost platforms for teaching integrated systems}.
\bjtitle{IEEE Robotics \& automation magazine}
\bvolume{10}(\bissue{2}),
\bfpage{4}--\blpage{6}
(\byear{2003})
\doiurl{10.1109/mra.2003.1213610}
\end{barticle}
\endbibitem

\bibitem[\protect\citeauthoryear{Roberts and McCowan}{2004}]{roberts2004infantsim}
\begin{botherref}
\oauthor{\bsnm{Roberts}, \binits{S.W.}},
\oauthor{\bsnm{McCowan}, \binits{R.J.}}:
The effectiveness of infant simulators.
Adolescence
\textbf{39}(155)
(2004)
\end{botherref}
\endbibitem

\bibitem[\protect\citeauthoryear{Hand and Varan}{2009}]{hand2009interactive}
\begin{barticle}
\bauthor{\bsnm{Hand}, \binits{S.}},
\bauthor{\bsnm{Varan}, \binits{D.}}:
\batitle{Interactive stories and the audience: Why empathy is important}.
\bjtitle{Computers in Entertainment (CIE)}
\bvolume{7}(\bissue{3}),
\bfpage{1}--\blpage{14}
(\byear{2009})
\doiurl{10.1145/1594943.1594951}
\end{barticle}
\endbibitem

\bibitem[\protect\citeauthoryear{Aylett et~al.}{2007}]{aylett2007fearnot}
\begin{bchapter}
\bauthor{\bsnm{Aylett}, \binits{R.}},
\bauthor{\bsnm{Vala}, \binits{M.}},
\bauthor{\bsnm{Sequeira}, \binits{P.}},
\bauthor{\bsnm{Paiva}, \binits{A.}}:
\bctitle{Fearnot!--an emergent narrative approach to virtual dramas for anti-bullying education}.
In: \bbtitle{International Conference on Virtual Storytelling},
pp. \bfpage{202}--\blpage{205}
(\byear{2007}).
\doiurl{10.1007/978-3-540-77039-8_19} .
\bcomment{Springer}
\end{bchapter}
\endbibitem

\bibitem[\protect\citeauthoryear{Lee et~al.}{2017}]{lee2017steps}
\begin{bchapter}
\bauthor{\bsnm{Lee}, \binits{H.R.}},
\bauthor{\bsnm{{\v{S}}abanovi{\'c}}, \binits{S.}},
\bauthor{\bsnm{Chang}, \binits{W.-L.}},
\bauthor{\bsnm{Nagata}, \binits{S.}},
\bauthor{\bsnm{Piatt}, \binits{J.}},
\bauthor{\bsnm{Bennett}, \binits{C.}},
\bauthor{\bsnm{Hakken}, \binits{D.}}:
\bctitle{Steps toward participatory design of social robots: mutual learning with older adults with depression}.
In: \bbtitle{Proceedings of the 2017 ACM/IEEE International Conference on Human-robot Interaction},
pp. \bfpage{244}--\blpage{253}
(\byear{2017}).
\doiurl{10.1145/2909824.3020237}
\end{bchapter}
\endbibitem

\bibitem[\protect\citeauthoryear{Druin}{2002}]{druin2002role}
\begin{barticle}
\bauthor{\bsnm{Druin}, \binits{A.}}:
\batitle{The role of children in the design of new technology}.
\bjtitle{Behaviour and information technology}
\bvolume{21}(\bissue{1}),
\bfpage{1}--\blpage{25}
(\byear{2002})
\doiurl{10.1080/01449290110108659}
\end{barticle}
\endbibitem

\bibitem[\protect\citeauthoryear{Derboven et~al.}{2015}]{derboven2015multimodal}
\begin{bchapter}
\bauthor{\bsnm{Derboven}, \binits{J.}},
\bauthor{\bsnm{Van~Mechelen}, \binits{M.}},
\bauthor{\bsnm{Slegers}, \binits{K.}}:
\bctitle{Multimodal analysis in participatory design with children: A primary school case study}.
In: \bbtitle{Proceedings of the 33rd Annual ACM Conference on Human Factors in Computing Systems},
pp. \bfpage{2825}--\blpage{2828}
(\byear{2015}).
\doiurl{10.1145/2702123.2702475}
\end{bchapter}
\endbibitem

\bibitem[\protect\citeauthoryear{Sanoubari et~al.}{2021}]{sanoubari2021remotecodesign}
\begin{bchapter}
\bauthor{\bsnm{Sanoubari}, \binits{E.}},
\bauthor{\bsnm{Mu{\~n}oz~Cardona}, \binits{J.E.}},
\bauthor{\bsnm{Mahdi}, \binits{H.}},
\bauthor{\bsnm{Young}, \binits{J.E.}},
\bauthor{\bsnm{Houston}, \binits{A.}},
\bauthor{\bsnm{Dautenhahn}, \binits{K.}}:
\bctitle{Robots, bullies and stories: a remote co-design study with children}.
In: \bbtitle{Interaction Design and Children},
pp. \bfpage{171}--\blpage{182}
(\byear{2021}).
\doiurl{10.1145/3459990.3460725}
\end{bchapter}
\endbibitem

\bibitem[\protect\citeauthoryear{Foster et~al.}{2023}]{foster2023paincodesign}
\begin{bchapter}
\bauthor{\bsnm{Foster}, \binits{M.E.}},
\bauthor{\bsnm{Candelaria}, \binits{P.}},
\bauthor{\bsnm{Dwyer}, \binits{L.J.}},
\bauthor{\bsnm{Hudson}, \binits{S.}},
\bauthor{\bsnm{Lindsay}, \binits{A.}},
\bauthor{\bsnm{Nishat}, \binits{F.}},
\bauthor{\bsnm{Pacquing}, \binits{M.}},
\bauthor{\bsnm{Petrick}, \binits{R.P.}},
\bauthor{\bsnm{Ram{\'\i}rez-Duque}, \binits{A.A.}},
\bauthor{\bsnm{Stinson}, \binits{J.}}, \betal:
\bctitle{Co-design of a social robot for distraction in the paediatric emergency department}.
In: \bbtitle{Companion of the 2023 ACM/IEEE International Conference on Human-Robot Interaction},
pp. \bfpage{461}--\blpage{465}
(\byear{2023}).
\doiurl{10.1145/3568294.3580127}
\end{bchapter}
\endbibitem

\bibitem[\protect\citeauthoryear{Mott et~al.}{2022}]{mott2022codesign}
\begin{bchapter}
\bauthor{\bsnm{Mott}, \binits{T.}},
\bauthor{\bsnm{Bejarano}, \binits{A.}},
\bauthor{\bsnm{Williams}, \binits{T.}}:
\bctitle{Robot co-design can help us engage child stakeholders in ethical reflection}.
In: \bbtitle{2022 17th ACM/IEEE International Conference on Human-Robot Interaction (HRI)},
pp. \bfpage{14}--\blpage{23}
(\byear{2022}).
\doiurl{10.1109/hri53351.2022.9889430} .
\bcomment{IEEE}
\end{bchapter}
\endbibitem

\bibitem[\protect\citeauthoryear{Arevalo~Arboleda et~al.}{2021}]{arevalo2021reflecting}
\begin{bchapter}
\bauthor{\bsnm{Arevalo~Arboleda}, \binits{S.}},
\bauthor{\bsnm{Pascher}, \binits{M.}},
\bauthor{\bsnm{Baumeister}, \binits{A.}},
\bauthor{\bsnm{Klein}, \binits{B.}},
\bauthor{\bsnm{Gerken}, \binits{J.}}:
\bctitle{Reflecting upon participatory design in human-robot collaboration for people with motor disabilities: Challenges and lessons learned from three multiyear projects}.
In: \bbtitle{The 14th PErvasive Technologies Related to Assistive Environments Conference},
pp. \bfpage{147}--\blpage{155}
(\byear{2021}).
\doiurl{10.1145/3453892.3458044}
\end{bchapter}
\endbibitem

\bibitem[\protect\citeauthoryear{Ostrowski et~al.}{2021}]{ostrowski2021codesign}
\begin{bchapter}
\bauthor{\bsnm{Ostrowski}, \binits{A.K.}},
\bauthor{\bsnm{Breazeal}, \binits{C.}},
\bauthor{\bsnm{Park}, \binits{H.W.}}:
\bctitle{Long-term co-design guidelines: empowering older adults as co-designers of social robots}.
In: \bbtitle{2021 30th IEEE International Conference on Robot \& Human Interactive Communication (RO-MAN)},
pp. \bfpage{1165}--\blpage{1172}
(\byear{2021}).
\doiurl{10.1109/ro-man50785.2021.9515559} .
\bcomment{IEEE}
\end{bchapter}
\endbibitem

\bibitem[\protect\citeauthoryear{Uzor et~al.}{2012}]{uzor2012senior}
\begin{bchapter}
\bauthor{\bsnm{Uzor}, \binits{S.}},
\bauthor{\bsnm{Baillie}, \binits{L.}},
\bauthor{\bsnm{Skelton}, \binits{D.}}:
\bctitle{Senior designers: empowering seniors to design enjoyable falls rehabilitation tools}.
In: \bbtitle{Proceedings of the SIGCHI Conference on Human Factors in Computing Systems},
pp. \bfpage{1179}--\blpage{1188}
(\byear{2012}).
\doiurl{10.1145/2207676.2208568}
\end{bchapter}
\endbibitem

\bibitem[\protect\citeauthoryear{Reich et~al.}{2018}]{reich2018lego}
\begin{barticle}
\bauthor{\bsnm{Reich}, \binits{S.M.}},
\bauthor{\bsnm{Black}, \binits{R.W.}},
\bauthor{\bsnm{Foliaki}, \binits{T.}}:
\batitle{Constructing difference: Lego{\textregistered} set narratives promote stereotypic gender roles and play}.
\bjtitle{Sex Roles}
\bvolume{79},
\bfpage{285}--\blpage{298}
(\byear{2018})
\doiurl{10.1007/s11199-017-0868-2}
\end{barticle}
\endbibitem

\bibitem[\protect\citeauthoryear{Tipper}{2011}]{tipper2011relationships}
\begin{barticle}
\bauthor{\bsnm{Tipper}, \binits{B.}}:
\batitle{‘a dog who i know quite well’: Everyday relationships between children and animals}.
\bjtitle{Children's Geographies}
\bvolume{9}(\bissue{2}),
\bfpage{145}--\blpage{165}
(\byear{2011})
\doiurl{10.1080/14733285.2011.562378}
\end{barticle}
\endbibitem

\bibitem[\protect\citeauthoryear{Winkle et~al.}{2018}]{winkle2018therapists}
\begin{bchapter}
\bauthor{\bsnm{Winkle}, \binits{K.}},
\bauthor{\bsnm{Caleb-Solly}, \binits{P.}},
\bauthor{\bsnm{Turton}, \binits{A.}},
\bauthor{\bsnm{Bremner}, \binits{P.}}:
\bctitle{Social robots for engagement in rehabilitative therapies: Design implications from a study with therapists}.
In: \bbtitle{Proceedings of the 2018 Acm/ieee International Conference on Human-robot Interaction},
pp. \bfpage{289}--\blpage{297}
(\byear{2018}).
\doiurl{10.1145/3171221.3171273}
\end{bchapter}
\endbibitem

\bibitem[\protect\citeauthoryear{Guha et~al.}{2013}]{guha2013reflections}
\begin{barticle}
\bauthor{\bsnm{Guha}, \binits{M.L.}},
\bauthor{\bsnm{Druin}, \binits{A.}},
\bauthor{\bsnm{Fails}, \binits{J.A.}}:
\batitle{Cooperative inquiry revisited: Reflections of the past and guidelines for the future of intergenerational co-design}.
\bjtitle{International Journal of Child-Computer Interaction}
\bvolume{1}(\bissue{1}),
\bfpage{14}--\blpage{23}
(\byear{2013})
\doiurl{10.1016/j.ijcci.2012.08.003}
\end{barticle}
\endbibitem

\bibitem[\protect\citeauthoryear{Voysey et~al.}{2023}]{voysey2023introducing}
\begin{bchapter}
\bauthor{\bsnm{Voysey}, \binits{I.}},
\bauthor{\bsnm{Bettosi}, \binits{C.}},
\bauthor{\bsnm{Nault}, \binits{E.}},
\bauthor{\bsnm{Stals}, \binits{S.}},
\bauthor{\bsnm{Baillie}, \binits{L.}}:
\bctitle{Introducing children and young people with sight loss to social robots: A preliminary workshop}.
In: \bbtitle{Companion of the 2023 ACM/IEEE International Conference on Human-Robot Interaction},
pp. \bfpage{384}--\blpage{388}
(\byear{2023}).
\doiurl{10.1145/3568294.3580111}
\end{bchapter}
\endbibitem

\bibitem[\protect\citeauthoryear{Barendregt et~al.}{2020}]{barendregt2020demystifying}
\begin{barticle}
\bauthor{\bsnm{Barendregt}, \binits{W.}},
\bauthor{\bsnm{Ekstr{\"o}m}, \binits{S.}},
\bauthor{\bsnm{Kiesewetter}, \binits{S.}},
\bauthor{\bsnm{Pareto}, \binits{L.}},
\bauthor{\bsnm{Serholt}, \binits{S.}}:
\batitle{Demystifying robots in the co-design of a tutee robot with primary school children.}
\bjtitle{Interaction Design and Architecture(s) IxD{\&}A}
\bvolume{44},
\bfpage{109}--\blpage{128}
(\byear{2020})
\doiurl{10.55612/s-5002-044-006}
\end{barticle}
\endbibitem

\bibitem[\protect\citeauthoryear{Guha et~al.}{2004}]{guha2004mixing}
\begin{bchapter}
\bauthor{\bsnm{Guha}, \binits{M.L.}},
\bauthor{\bsnm{Druin}, \binits{A.}},
\bauthor{\bsnm{Chipman}, \binits{G.}},
\bauthor{\bsnm{Fails}, \binits{J.A.}},
\bauthor{\bsnm{Simms}, \binits{S.}},
\bauthor{\bsnm{Farber}, \binits{A.}}:
\bctitle{Mixing ideas: a new technique for working with young children as design partners}.
In: \bbtitle{Proceedings of the 2004 Conference on Interaction Design and Children: Building a Community},
pp. \bfpage{35}--\blpage{42}
(\byear{2004}).
\doiurl{10.1145/1017833.1017838}
\end{bchapter}
\endbibitem

\bibitem[\protect\citeauthoryear{Angell et~al.}{2015}]{angell2015draw}
\begin{barticle}
\bauthor{\bsnm{Angell}, \binits{C.}},
\bauthor{\bsnm{Alexander}, \binits{J.}},
\bauthor{\bsnm{Hunt}, \binits{J.A.}}:
\batitle{‘draw, write and tell’: A literature review and methodological development on the ‘draw and write’research method}.
\bjtitle{Journal of Early Childhood Research}
\bvolume{13}(\bissue{1}),
\bfpage{17}--\blpage{28}
(\byear{2015})
\doiurl{10.1177/1476718x14538592}
\end{barticle}
\endbibitem

\bibitem[\protect\citeauthoryear{Gauntlett and Holzwarth}{2006}]{gauntlett2006creative}
\begin{barticle}
\bauthor{\bsnm{Gauntlett}, \binits{D.}},
\bauthor{\bsnm{Holzwarth}, \binits{P.}}:
\batitle{Creative and visual methods for exploring identities}.
\bjtitle{Visual studies}
\bvolume{21}(\bissue{01}),
\bfpage{82}--\blpage{91}
(\byear{2006})
\doiurl{10.1080/14725860600613261}
\end{barticle}
\endbibitem

\bibitem[\protect\citeauthoryear{Georgiou et~al.}{2023}]{georgiou2023refugees}
\begin{barticle}
\bauthor{\bsnm{Georgiou}, \binits{T.}},
\bauthor{\bsnm{Baillie}, \binits{L.}},
\bauthor{\bsnm{Netto}, \binits{G.}},
\bauthor{\bsnm{Paterson}, \binits{S.}}:
\batitle{Investigating technology concepts to support rohingya refugees in malaysia}.
\bjtitle{Interacting with Computers}
(\byear{2023})
\doiurl{10.1093/iwc/iwad026}
\end{barticle}
\endbibitem

\bibitem[\protect\citeauthoryear{Beelen et~al.}{2022}]{beelen2022designing}
\begin{barticle}
\bauthor{\bsnm{Beelen}, \binits{T.}},
\bauthor{\bsnm{Velner}, \binits{E.}},
\bauthor{\bsnm{Ordelman}, \binits{R.}},
\bauthor{\bsnm{Truong}, \binits{K.P.}},
\bauthor{\bsnm{Evers}, \binits{V.}},
\bauthor{\bsnm{Huibers}, \binits{T.}}:
\batitle{Designing conversational robots with children during the pandemic}.
\bjtitle{arXiv preprint arXiv:2205.11300}
(\byear{2022})
\doiurl{arXiv:2205.11300}
\end{barticle}
\endbibitem

\bibitem[\protect\citeauthoryear{Bj{\"o}rling and Rose}{2019}]{bjorling2019participatory}
\begin{barticle}
\bauthor{\bsnm{Bj{\"o}rling}, \binits{E.A.}},
\bauthor{\bsnm{Rose}, \binits{E.}}:
\batitle{Participatory research principles in human-centered design: engaging teens in the co-design of a social robot}.
\bjtitle{Multimodal Technologies and Interaction}
\bvolume{3}(\bissue{1}),
\bfpage{8}
(\byear{2019})
\doiurl{10.3390/mti3010008}
\end{barticle}
\endbibitem

\bibitem[\protect\citeauthoryear{Murray}{1943}]{murray1943thematic}
\begin{botherref}
\oauthor{\bsnm{Murray}, \binits{H.A.}}:
Thematic apperception test.
(1943)
\end{botherref}
\endbibitem

\bibitem[\protect\citeauthoryear{Shapiro et~al.}{2013}]{shapiro2013anicare}
\begin{bbook}
\bauthor{\bsnm{Shapiro}, \binits{K.}},
\bauthor{\bsnm{Randour}, \binits{M.L.}},
\bauthor{\bsnm{Krinsk}, \binits{S.}},
\bauthor{\bsnm{Wolf}, \binits{J.L.}}:
\bbtitle{The Assessment and Treatment of Children Who Abuse Animals: The AniCare Child Approach}.
\bpublisher{Springer},
\blocation{Switzerland}
(\byear{2013}).
\doiurl{10.1007/978-3-319-01089-2}
\end{bbook}
\endbibitem

\bibitem[\protect\citeauthoryear{Moraveji et~al.}{2007}]{moraveji2007comicboarding}
\begin{bchapter}
\bauthor{\bsnm{Moraveji}, \binits{N.}},
\bauthor{\bsnm{Li}, \binits{J.}},
\bauthor{\bsnm{Ding}, \binits{J.}},
\bauthor{\bsnm{O'Kelley}, \binits{P.}},
\bauthor{\bsnm{Woolf}, \binits{S.}}:
\bctitle{Comicboarding: using comics as proxies for participatory design with children}.
In: \bbtitle{Proceedings of the SIGCHI Conference on Human Factors in Computing Systems},
pp. \bfpage{1371}--\blpage{1374}
(\byear{2007}).
\doiurl{10.1145/1240624.1240832}
\end{bchapter}
\endbibitem

\bibitem[\protect\citeauthoryear{DiSalvo et~al.}{2008}]{disalvo2008neighborhood}
\begin{bchapter}
\bauthor{\bsnm{DiSalvo}, \binits{C.}},
\bauthor{\bsnm{Nourbakhsh}, \binits{I.}},
\bauthor{\bsnm{Holstius}, \binits{D.}},
\bauthor{\bsnm{Akin}, \binits{A.}},
\bauthor{\bsnm{Louw}, \binits{M.}}:
\bctitle{The neighborhood networks project: a case study of critical engagement and creative expression through participatory design}.
In: \bbtitle{Proceedings of the Tenth Anniversary Conference on Participatory Design 2008},
pp. \bfpage{41}--\blpage{50}
(\byear{2008}).
\doiurl{10.5555/1795234.1795241}
\end{bchapter}
\endbibitem

\bibitem[\protect\citeauthoryear{Heary and Hennessy}{2002}]{heary2002focus}
\begin{barticle}
\bauthor{\bsnm{Heary}, \binits{C.M.}},
\bauthor{\bsnm{Hennessy}, \binits{E.}}:
\batitle{The use of focus group interviews in pediatric health care research}.
\bjtitle{Journal of pediatric psychology}
\bvolume{27}(\bissue{1}),
\bfpage{47}--\blpage{57}
(\byear{2002})
\doiurl{10.1093/jpepsy/27.1.47}
\end{barticle}
\endbibitem

\bibitem[\protect\citeauthoryear{Voysey et~al.}{2022}]{voysey2022influence}
\begin{bchapter}
\bauthor{\bsnm{Voysey}, \binits{I.}},
\bauthor{\bsnm{Baillie}, \binits{L.}},
\bauthor{\bsnm{Williams}, \binits{J.}},
\bauthor{\bsnm{Herrmann}, \binits{J.M.}}:
\bctitle{Influence of animallike affective non-verbal behavior on children’s perceptions of a zoomorphic robot}.
In: \bbtitle{2022 31st IEEE International Conference on Robot and Human Interactive Communication (RO-MAN)},
pp. \bfpage{1443}--\blpage{1450}
(\byear{2022}).
\doiurl{10.1109/ro-man53752.2022.9900621} .
\bcomment{IEEE}
\end{bchapter}
\endbibitem

\bibitem[\protect\citeauthoryear{Glaser}{1965}]{glaser1965constant}
\begin{barticle}
\bauthor{\bsnm{Glaser}, \binits{B.G.}}:
\batitle{The constant comparative method of qualitative analysis}.
\bjtitle{Social problems}
\bvolume{12}(\bissue{4}),
\bfpage{436}--\blpage{445}
(\byear{1965})
\doiurl{10.1525/sp.1965.12.4.03a00070}
\end{barticle}
\endbibitem

\bibitem[\protect\citeauthoryear{{Farm Animal Welfare Council}}{1993}]{fawc1993fivefreedoms}
\begin{botherref}
\oauthor{\bsnm{{Farm Animal Welfare Council}}}:
Second report on priorities for research and development in farm animal welfare.
Technical report,
Department for Environment, Food and Rural Affairs,
London, UK
(1993)
\end{botherref}
\endbibitem

\bibitem[\protect\citeauthoryear{Meints et~al.}{2010}]{meints2010prevent}
\begin{barticle}
\bauthor{\bsnm{Meints}, \binits{K.}},
\bauthor{\bsnm{Racca}, \binits{A.}},
\bauthor{\bsnm{Hickey}, \binits{N.}}:
\batitle{How to prevent dog bite injuries? children misinterpret dogs facial expressions}.
\bjtitle{Injury Prevention}
\bvolume{16}(\bissue{Suppl 1}),
\bfpage{68}--\blpage{68}
(\byear{2010})
\doiurl{10.1136/ip.2010.029215.246}
\end{barticle}
\endbibitem

\bibitem[\protect\citeauthoryear{L{\"o}ffler et~al.}{2020}]{loffler2020uncanny}
\begin{bchapter}
\bauthor{\bsnm{L{\"o}ffler}, \binits{D.}},
\bauthor{\bsnm{D{\"o}rrenb{\"a}cher}, \binits{J.}},
\bauthor{\bsnm{Hassenzahl}, \binits{M.}}:
\bctitle{The uncanny valley effect in zoomorphic robots: The u-shaped relation between animal likeness and likeability}.
In: \bbtitle{Proceedings of the 2020 ACM/IEEE International Conference on Human-robot Interaction},
pp. \bfpage{261}--\blpage{270}
(\byear{2020}).
\doiurl{10.1145/3319502.3374788}
\end{bchapter}
\endbibitem

\bibitem[\protect\citeauthoryear{Ghafurian et~al.}{2022}]{ghafurian2022zoomorphic}
\begin{botherref}
\oauthor{\bsnm{Ghafurian}, \binits{M.}},
\oauthor{\bsnm{Lakatos}, \binits{G.}},
\oauthor{\bsnm{Dautenhahn}, \binits{K.}}:
The zoomorphic {MiRo} robot’s affective expression design and perceived appearance.
International Journal of Social Robotics,
1--18
(2022)
\doiurl{10.1007/s12369-021-00832-3}
\end{botherref}
\endbibitem

\bibitem[\protect\citeauthoryear{Yohanan and MacLean}{2011}]{yohanan2011designaffect}
\begin{bchapter}
\bauthor{\bsnm{Yohanan}, \binits{S.}},
\bauthor{\bsnm{MacLean}, \binits{K.E.}}:
\bctitle{Design and assessment of the haptic creature's affect display}.
In: \bbtitle{Proceedings of the 6th International Conference on Human-robot Interaction},
pp. \bfpage{473}--\blpage{480}
(\byear{2011}).
\doiurl{10.1145/1957656.1957820}
\end{bchapter}
\endbibitem

\bibitem[\protect\citeauthoryear{Hawkins and Williams}{2016}]{hawkins2016children_bam}
\begin{barticle}
\bauthor{\bsnm{Hawkins}, \binits{R.D.}},
\bauthor{\bsnm{Williams}, \binits{J.M.}}:
\batitle{Children’s beliefs about animal minds ({Child-BAM}): Associations with positive and negative child--animal interactions}.
\bjtitle{Anthrozo{\"o}s}
\bvolume{29}(\bissue{3}),
\bfpage{503}--\blpage{519}
(\byear{2016})
\doiurl{10.1080/08927936.2016.1189749}
\end{barticle}
\endbibitem

\bibitem[\protect\citeauthoryear{Wan et~al.}{2012}]{wan2012dogfear}
\begin{barticle}
\bauthor{\bsnm{Wan}, \binits{M.}},
\bauthor{\bsnm{Bolger}, \binits{N.}},
\bauthor{\bsnm{Champagne}, \binits{F.A.}}:
\batitle{Human perception of fear in dogs varies according to experience with dogs}.
\bjtitle{PLoS one}
\bvolume{7}(\bissue{12}),
\bfpage{51775}
(\byear{2012})
\doiurl{10.1371/journal.pone.0051775}
\end{barticle}
\endbibitem

\bibitem[\protect\citeauthoryear{Demirbas et~al.}{2016}]{demirbas2016dogfear}
\begin{barticle}
\bauthor{\bsnm{Demirbas}, \binits{Y.S.}},
\bauthor{\bsnm{Ozturk}, \binits{H.}},
\bauthor{\bsnm{Emre}, \binits{B.}},
\bauthor{\bsnm{Kockaya}, \binits{M.}},
\bauthor{\bsnm{Ozvardar}, \binits{T.}},
\bauthor{\bsnm{Scott}, \binits{A.}}:
\batitle{Adults’ ability to interpret canine body language during a dog--child interaction}.
\bjtitle{Anthrozo{\"o}s}
\bvolume{29}(\bissue{4}),
\bfpage{581}--\blpage{596}
(\byear{2016})
\doiurl{10.1080/08927936.2016.1228750}
\end{barticle}
\endbibitem

\bibitem[\protect\citeauthoryear{Dawson et~al.}{2019}]{dawson2019cataffect}
\begin{barticle}
\bauthor{\bsnm{Dawson}, \binits{L.C.}},
\bauthor{\bsnm{Cheal}, \binits{J.}},
\bauthor{\bsnm{Niel}, \binits{L.}},
\bauthor{\bsnm{Mason}, \binits{G.}}:
\batitle{Humans can identify cats’ affective states from subtle facial expressions}.
\bjtitle{Animal Welfare}
\bvolume{28}(\bissue{4}),
\bfpage{519}--\blpage{531}
(\byear{2019})
\doiurl{10.7120/09627286.28.4.519}
\end{barticle}
\endbibitem

\bibitem[\protect\citeauthoryear{Allen}{2015}]{allen2015preschooltaxonomic}
\begin{barticle}
\bauthor{\bsnm{Allen}, \binits{M.}}:
\batitle{Preschool children's taxonomic knowledge of animal species}.
\bjtitle{Journal of Research in Science Teaching}
\bvolume{52}(\bissue{1}),
\bfpage{107}--\blpage{134}
(\byear{2015})
\doiurl{10.1002/tea.21191}
\end{barticle}
\endbibitem

\bibitem[\protect\citeauthoryear{Rusca and Tonucci}{1992}]{rusca1992developmentconceptsanimal}
\begin{barticle}
\bauthor{\bsnm{Rusca}, \binits{G.}},
\bauthor{\bsnm{Tonucci}, \binits{F.}}:
\batitle{Development of the concepts of living and animal in the child}.
\bjtitle{European journal of psychology of education}
\bvolume{7},
\bfpage{151}--\blpage{176}
(\byear{1992})
\doiurl{10.1007/bf03172891}
\end{barticle}
\endbibitem

\bibitem[\protect\citeauthoryear{{Cats Protection}}{2021}]{catsprotection2021behaviourguide}
\begin{bbook}
\bauthor{\bsnm{{Cats Protection}}}:
\bbtitle{The Behaviour Guide}.
\bpublisher{Cats Protection},
\blocation{Haywards Heath}
(\byear{2021})
\end{bbook}
\endbibitem

\bibitem[\protect\citeauthoryear{Muldoon et~al.}{2015}]{muldoon2015ijustplayed}
\begin{barticle}
\bauthor{\bsnm{Muldoon}, \binits{J.C.}},
\bauthor{\bsnm{Williams}, \binits{J.M.}},
\bauthor{\bsnm{Lawrence}, \binits{A.}}:
\batitle{{‘Mum cleaned it and I just played with it’: Children’s perceptions of their roles and responsibilities in the care of family pets}}.
\bjtitle{Childhood}
\bvolume{22}(\bissue{2}),
\bfpage{201}--\blpage{216}
(\byear{2015})
\doiurl{10.1177/0907568214524457}
\end{barticle}
\endbibitem

\bibitem[\protect\citeauthoryear{Archer and Monton}{2011}]{archer2011preferences}
\begin{barticle}
\bauthor{\bsnm{Archer}, \binits{J.}},
\bauthor{\bsnm{Monton}, \binits{S.}}:
\batitle{Preferences for infant facial features in pet dogs and cats}.
\bjtitle{Ethology}
\bvolume{117}(\bissue{3}),
\bfpage{217}--\blpage{226}
(\byear{2011})
\doiurl{10.1111/j.1439-0310.2010.01863.x}
\end{barticle}
\endbibitem

\bibitem[\protect\citeauthoryear{Little}{2012}]{little2012manipulation}
\begin{barticle}
\bauthor{\bsnm{Little}, \binits{A.C.}}:
\batitle{Manipulation of infant-like traits affects perceived cuteness of infant, adult and cat faces}.
\bjtitle{Ethology}
\bvolume{118}(\bissue{8}),
\bfpage{775}--\blpage{782}
(\byear{2012})
\doiurl{10.1111/j.1439-0310.2012.02068.x}
\end{barticle}
\endbibitem

\bibitem[\protect\citeauthoryear{Nittono et~al.}{2012}]{nittono2012power}
\begin{barticle}
\bauthor{\bsnm{Nittono}, \binits{H.}},
\bauthor{\bsnm{Fukushima}, \binits{M.}},
\bauthor{\bsnm{Yano}, \binits{A.}},
\bauthor{\bsnm{Moriya}, \binits{H.}}:
\batitle{The power of kawaii: Viewing cute images promotes a careful behavior and narrows attentional focus}.
\bjtitle{PloS one}
\bvolume{7}(\bissue{9}),
\bfpage{46362}
(\byear{2012})
\doiurl{10.1371/journal.pone.0046362}
\end{barticle}
\endbibitem

\bibitem[\protect\citeauthoryear{Sherman et~al.}{2009}]{sherman2009viewing}
\begin{barticle}
\bauthor{\bsnm{Sherman}, \binits{G.D.}},
\bauthor{\bsnm{Haidt}, \binits{J.}},
\bauthor{\bsnm{Coan}, \binits{J.A.}}:
\batitle{Viewing cute images increases behavioral carefulness.}
\bjtitle{Emotion}
\bvolume{9}(\bissue{2}),
\bfpage{282}
(\byear{2009})
\doiurl{10.1037/a0014904}
\end{barticle}
\endbibitem

\bibitem[\protect\citeauthoryear{Brosch et~al.}{2007}]{brosch2007baby}
\begin{botherref}
\oauthor{\bsnm{Brosch}, \binits{T.}},
\oauthor{\bsnm{Sander}, \binits{D.}},
\oauthor{\bsnm{Scherer}, \binits{K.R.}}:
That baby caught my eye... attention capture by infant faces.
(2007)
\doiurl{10.1037/1528-3542.7.3.685}
\end{botherref}
\endbibitem

\bibitem[\protect\citeauthoryear{Kringelbach et~al.}{2008}]{kringelbach2008neural}
\begin{barticle}
\bauthor{\bsnm{Kringelbach}, \binits{M.L.}},
\bauthor{\bsnm{Lehtonen}, \binits{A.}},
\bauthor{\bsnm{Squire}, \binits{S.}},
\bauthor{\bsnm{Harvey}, \binits{A.G.}},
\bauthor{\bsnm{Craske}, \binits{M.G.}},
\bauthor{\bsnm{Holliday}, \binits{I.E.}},
\bauthor{\bsnm{Green}, \binits{A.L.}},
\bauthor{\bsnm{Aziz}, \binits{T.Z.}},
\bauthor{\bsnm{Hansen}, \binits{P.C.}},
\bauthor{\bsnm{Cornelissen}, \binits{P.L.}}, \betal:
\batitle{A specific and rapid neural signature for parental instinct}.
\bjtitle{PloS one}
\bvolume{3}(\bissue{2}),
\bfpage{1664}
(\byear{2008})
\doiurl{10.1371/journal.pone.0001664}
\end{barticle}
\endbibitem

\bibitem[\protect\citeauthoryear{Golle et~al.}{2015}]{golle2015preference}
\begin{barticle}
\bauthor{\bsnm{Golle}, \binits{J.}},
\bauthor{\bsnm{Probst}, \binits{F.}},
\bauthor{\bsnm{Mast}, \binits{F.W.}},
\bauthor{\bsnm{Lobmaier}, \binits{J.S.}}:
\batitle{Preference for cute infants does not depend on their ethnicity or species: evidence from hypothetical adoption and donation paradigms}.
\bjtitle{PloS one}
\bvolume{10}(\bissue{4}),
\bfpage{0121554}
(\byear{2015})
\doiurl{10.1371/journal.pone.0121554}
\end{barticle}
\endbibitem

\bibitem[\protect\citeauthoryear{Sherman and Haidt}{2011}]{sherman2011cuteness}
\begin{barticle}
\bauthor{\bsnm{Sherman}, \binits{G.D.}},
\bauthor{\bsnm{Haidt}, \binits{J.}}:
\batitle{Cuteness and disgust: The humanizing and dehumanizing effects of emotion}.
\bjtitle{Emotion Review}
\bvolume{3}(\bissue{3}),
\bfpage{245}--\blpage{251}
(\byear{2011})
\doiurl{10.1177/1754073911402396}
\end{barticle}
\endbibitem

\bibitem[\protect\citeauthoryear{Lorenz}{1971}]{lorenz1971behaviour}
\begin{bbook}
\bauthor{\bsnm{Lorenz}, \binits{K.}}:
\bbtitle{Studies in Animal and Human Behaviour. Volume II / Konrad Lorenz.}
\bsertitle{Studies in Animal and Human Behaviour ; Volume II}.
\bpublisher{Harvard University Press,},
\blocation{Cambridge, MA}
(\byear{1971}).
\doiurl{10.4159/harvard.9780674430426} .
\bcomment{Trans. Robert Martin}
\end{bbook}
\endbibitem

\bibitem[\protect\citeauthoryear{Packer et~al.}{2017}]{packer2017purchasingpopular}
\begin{barticle}
\bauthor{\bsnm{Packer}, \binits{R.}},
\bauthor{\bsnm{Murphy}, \binits{D.}},
\bauthor{\bsnm{Farnworth}, \binits{M.}}:
\batitle{Purchasing popular purebreds: investigating the influence of breed-type on the pre-purchase motivations and behaviour of dog owners}.
\bjtitle{Animal welfare}
\bvolume{26}(\bissue{2}),
\bfpage{191}--\blpage{201}
(\byear{2017})
\doiurl{10.7120/09627286.26.2.191}
\end{barticle}
\endbibitem

\bibitem[\protect\citeauthoryear{Packer et~al.}{2019}]{packer2019truthdog}
\begin{barticle}
\bauthor{\bsnm{Packer}, \binits{R.M.}},
\bauthor{\bsnm{O’Neill}, \binits{D.G.}},
\bauthor{\bsnm{Fletcher}, \binits{F.}},
\bauthor{\bsnm{Farnworth}, \binits{M.J.}}:
\batitle{Great expectations, inconvenient truths, and the paradoxes of the dog-owner relationship for owners of brachycephalic dogs}.
\bjtitle{PLoS One}
\bvolume{14}(\bissue{7}),
\bfpage{0219918}
(\byear{2019})
\doiurl{10.1371/journal.pone.0219918}
\end{barticle}
\endbibitem

\bibitem[\protect\citeauthoryear{Miesler et~al.}{2011}]{miesler2011cuteevolutionary}
\begin{botherref}
\oauthor{\bsnm{Miesler}, \binits{L.}},
\oauthor{\bsnm{Leder}, \binits{H.}},
\oauthor{\bsnm{Herrmann}, \binits{A.}}:
Isn't it cute: An evolutionary perspective of baby-schema effects in visual product designs.
International Journal of Design
\textbf{5}(3)
(2011)
\end{botherref}
\endbibitem

\bibitem[\protect\citeauthoryear{Hinde and Barden}{1985}]{hinde1985evolution}
\begin{barticle}
\bauthor{\bsnm{Hinde}, \binits{R.A.}},
\bauthor{\bsnm{Barden}, \binits{L.A.}}:
\batitle{The evolution of the teddy bear}.
\bjtitle{Animal Behaviour}
\bvolume{33}(\bissue{4}),
\bfpage{1371}--\blpage{1373}
(\byear{1985})
\doiurl{10.1016/s0003-3472(85)80205-0}
\end{barticle}
\endbibitem

\bibitem[\protect\citeauthoryear{Br{\v{s}}{\v{c}}i{\'c} et~al.}{2015}]{brvsvcic2015escapingrobotabuse}
\begin{bchapter}
\bauthor{\bsnm{Br{\v{s}}{\v{c}}i{\'c}}, \binits{D.}},
\bauthor{\bsnm{Kidokoro}, \binits{H.}},
\bauthor{\bsnm{Suehiro}, \binits{Y.}},
\bauthor{\bsnm{Kanda}, \binits{T.}}:
\bctitle{Escaping from children's abuse of social robots}.
In: \bbtitle{Proceedings of the Tenth Annual ACM/IEEE International Conference on Human-robot Interaction},
pp. \bfpage{59}--\blpage{66}
(\byear{2015}).
\doiurl{10.1145/2696454.2696468}
\end{bchapter}
\endbibitem

\bibitem[\protect\citeauthoryear{Ligthart et~al.}{2020}]{ligthart2020designpatterns}
\begin{bchapter}
\bauthor{\bsnm{Ligthart}, \binits{M.E.}},
\bauthor{\bsnm{Neerincx}, \binits{M.A.}},
\bauthor{\bsnm{Hindriks}, \binits{K.V.}}:
\bctitle{Design patterns for an interactive storytelling robot to support children's engagement and agency}.
In: \bbtitle{Proceedings of the 2020 ACM/IEEE International Conference on Human-robot Interaction},
pp. \bfpage{409}--\blpage{418}
(\byear{2020}).
\doiurl{10.1145/3319502.3374826}
\end{bchapter}
\endbibitem

\bibitem[\protect\citeauthoryear{Muldoon et~al.}{2009}]{muldoon2009promotingdutyofcare}
\begin{botherref}
\oauthor{\bsnm{Muldoon}, \binits{J.}},
\oauthor{\bsnm{Williams}, \binits{J.}},
\oauthor{\bsnm{Lawrence}, \binits{A.}},
\oauthor{\bsnm{Lakestani}, \binits{N.}},
\oauthor{\bsnm{Currie}, \binits{C.}}:
Promoting a duty of care towards animals among children and young people: A literature review and findings from initial research to inform the development of interventions.
Technical report,
Child and Adolescent Health Research Unit, University of Edinburgh, Defra
(2009)
\end{botherref}
\endbibitem

\bibitem[\protect\citeauthoryear{Darling et~al.}{2015}]{darling2015empathicstories}
\begin{bchapter}
\bauthor{\bsnm{Darling}, \binits{K.}},
\bauthor{\bsnm{Nandy}, \binits{P.}},
\bauthor{\bsnm{Breazeal}, \binits{C.}}:
\bctitle{Empathic concern and the effect of stories in human-robot interaction}.
In: \bbtitle{2015 24th IEEE International Symposium on Robot and Human Interactive Communication (RO-MAN)},
pp. \bfpage{770}--\blpage{775}
(\byear{2015}).
\doiurl{10.1109/roman.2015.7333675} .
\bcomment{IEEE}
\end{bchapter}
\endbibitem

\bibitem[\protect\citeauthoryear{Lee et~al.}{2022}]{lee2022unboxingbackstory}
\begin{bchapter}
\bauthor{\bsnm{Lee}, \binits{C.P.}},
\bauthor{\bsnm{Cagiltay}, \binits{B.}},
\bauthor{\bsnm{Mutlu}, \binits{B.}}:
\bctitle{The unboxing experience: Exploration and design of initial interactions between children and social robots}.
In: \bbtitle{Proceedings of the 2022 CHI Conference on Human Factors in Computing Systems},
pp. \bfpage{1}--\blpage{14}
(\byear{2022}).
\doiurl{10.1145/3491102.3501955}
\end{bchapter}
\endbibitem

\bibitem[\protect\citeauthoryear{Michaelis et~al.}{2023}]{michaelis2023offscript}
\begin{bchapter}
\bauthor{\bsnm{Michaelis}, \binits{J.E.}},
\bauthor{\bsnm{Cagiltay}, \binits{B.}},
\bauthor{\bsnm{Ibtasar}, \binits{R.}},
\bauthor{\bsnm{Mutlu}, \binits{B.}}:
\bctitle{"off script:" design opportunities emerging from long-term social robot interactions in-the-wild}.
In: \bbtitle{Proceedings of the 2023 ACM/IEEE International Conference on Human-Robot Interaction},
pp. \bfpage{378}--\blpage{387}
(\byear{2023}).
\doiurl{10.1145/3568162.3576978}
\end{bchapter}
\endbibitem

\bibitem[\protect\citeauthoryear{{Scottish SPCA}}{2021}]{sspca2021marthagibson}
\begin{botherref}
\oauthor{\bsnm{{Scottish SPCA}}}:
Introducing Martha and Gibson
(2021).
\url{https://www.scottishspca.org/sites/default/files/2021-11/Martha%20and%20Gibson%20Activity%20Pack.pdf}
\end{botherref}
\endbibitem

\bibitem[\protect\citeauthoryear{Bucci et~al.}{2018}]{bucci2018furrynarrativeframe}
\begin{bchapter}
\bauthor{\bsnm{Bucci}, \binits{P.}},
\bauthor{\bsnm{Zhang}, \binits{L.}},
\bauthor{\bsnm{Cang}, \binits{X.L.}},
\bauthor{\bsnm{MacLean}, \binits{K.E.}}:
\bctitle{Is it happy? behavioural and narrative frame complexity impact perceptions of a simple furry robot's emotions}.
In: \bbtitle{Proceedings of the 2018 CHI Conference on Human Factors in Computing Systems},
pp. \bfpage{1}--\blpage{11}
(\byear{2018}).
\doiurl{10.1145/3173574.3174083}
\end{bchapter}
\endbibitem

\bibitem[\protect\citeauthoryear{Coleman et~al.}{2014}]{coleman2014conflictresolution}
\begin{bbook}
\bauthor{\bsnm{Coleman}, \binits{P.T.}},
\bauthor{\bsnm{Deutsch}, \binits{M.}},
\bauthor{\bsnm{Marcus}, \binits{E.C.}}:
\bbtitle{The Handbook of Conflict Resolution: Theory and Practice}.
\bpublisher{John Wiley \& Sons},
\blocation{Hoboken, New Jersey}
(\byear{2014})
\end{bbook}
\endbibitem

\bibitem[\protect\citeauthoryear{Leung and Lam}{2019}]{leung2019buildingconsensus}
\begin{barticle}
\bauthor{\bsnm{Leung}, \binits{T.T.}},
\bauthor{\bsnm{Lam}, \binits{B.C.}}:
\batitle{Building consensus on user participation in social work: A conversation analysis}.
\bjtitle{Journal of Social Work}
\bvolume{19}(\bissue{1}),
\bfpage{20}--\blpage{40}
(\byear{2019})
\doiurl{10.1177/1468017318757357}
\end{barticle}
\endbibitem

\bibitem[\protect\citeauthoryear{Briggs et~al.}{2005}]{briggs2005consensusbuilding}
\begin{botherref}
\oauthor{\bsnm{Briggs}, \binits{R.O.}},
\oauthor{\bsnm{Kolfschoten}, \binits{G.L.}},
\oauthor{\bsnm{Vreede}, \binits{G.-J.d.}}:
Toward a theoretical model of consensus building.
AMCIS 2005 Proceedings,
12
(2005)
\end{botherref}
\endbibitem

\bibitem[\protect\citeauthoryear{Hawkins et~al.}{2017}]{hawkins2017childhoodattachment}
\begin{barticle}
\bauthor{\bsnm{Hawkins}, \binits{R.D.}},
\bauthor{\bsnm{Williams}, \binits{J.M.}},
\bauthor{\bsnm{{Scottish SPCA}}}:
\batitle{Childhood attachment to pets: Associations between pet attachment, attitudes to animals, compassion, and humane behaviour}.
\bjtitle{International journal of environmental research and public health}
\bvolume{14}(\bissue{5}),
\bfpage{490}
(\byear{2017})
\doiurl{10.3390/ijerph14050490}
\end{barticle}
\endbibitem

\bibitem[\protect\citeauthoryear{Lakestani et~al.}{2014}]{lakestani2014interpretationdog}
\begin{barticle}
\bauthor{\bsnm{Lakestani}, \binits{N.N.}},
\bauthor{\bsnm{Donaldson}, \binits{M.L.}},
\bauthor{\bsnm{Waran}, \binits{N.}}:
\batitle{Interpretation of dog behavior by children and young adults}.
\bjtitle{Anthrozo{\"o}s}
\bvolume{27}(\bissue{1}),
\bfpage{65}--\blpage{80}
(\byear{2014})
\doiurl{10.2752/175303714x13837396326413}
\end{barticle}
\endbibitem

\bibitem[\protect\citeauthoryear{Overall and Love}{2001}]{overall2001dogbites}
\begin{barticle}
\bauthor{\bsnm{Overall}, \binits{K.L.}},
\bauthor{\bsnm{Love}, \binits{M.}}:
\batitle{Dog bites to humans—demography, epidemiology, injury, and risk}.
\bjtitle{Journal of the American Veterinary Medical Association}
\bvolume{218}(\bissue{12}),
\bfpage{1923}--\blpage{1934}
(\byear{2001})
\doiurl{10.2460/javma.2001.218.1923}
\end{barticle}
\endbibitem

\bibitem[\protect\citeauthoryear{Muldoon et~al.}{2016}]{muldoon2016perspectiveswelfare}
\begin{barticle}
\bauthor{\bsnm{Muldoon}, \binits{J.C.}},
\bauthor{\bsnm{Williams}, \binits{J.M.}},
\bauthor{\bsnm{Lawrence}, \binits{A.}}:
\batitle{Exploring children’s perspectives on the welfare needs of pet animals}.
\bjtitle{Anthrozo{\"o}s}
\bvolume{29}(\bissue{3}),
\bfpage{357}--\blpage{375}
(\byear{2016})
\doiurl{10.1080/08927936.2016.1181359}
\end{barticle}
\endbibitem

\bibitem[\protect\citeauthoryear{Hawkins et~al.}{2020}]{hawkins2020children}
\begin{barticle}
\bauthor{\bsnm{Hawkins}, \binits{R.D.}},
\bauthor{\bsnm{{Scottish SPCA}}},
\bauthor{\bsnm{Williams}, \binits{J.M.}}:
\batitle{Children’s attitudes towards animal cruelty: Exploration of predictors and socio-demographic variations}.
\bjtitle{Psychology, Crime \& Law}
\bvolume{26}(\bissue{3}),
\bfpage{226}--\blpage{247}
(\byear{2020})
\doiurl{10.1080/1068316x.2019.1652747}
\end{barticle}
\endbibitem

\bibitem[\protect\citeauthoryear{Bjerke et~al.}{1998}]{bjerke1998attitudesnorwegian}
\begin{barticle}
\bauthor{\bsnm{Bjerke}, \binits{T.}},
\bauthor{\bsnm{{\O}deg{\aa}rdstuen}, \binits{T.S.}},
\bauthor{\bsnm{Kaltenborn}, \binits{B.P.}}:
\batitle{Attitudes toward animals among norwegian adolescents}.
\bjtitle{Anthrozo{\"o}s}
\bvolume{11}(\bissue{2}),
\bfpage{79}--\blpage{86}
(\byear{1998})
\doiurl{10.2752/089279398787000742}
\end{barticle}
\endbibitem

\end{thebibliography}

\end{document}